\documentclass{article}

\usepackage[eandd, final]{neurips_2026}
\usepackage{microtype}
\usepackage{graphicx}
\usepackage{subcaption}
\usepackage[utf8]{inputenc} 
\usepackage[T1]{fontenc}    
\usepackage{hyperref}       
\usepackage{url}            
\usepackage{booktabs}       
\usepackage{amsfonts}       
\usepackage{nicefrac}       
\usepackage{microtype}      
\usepackage{xcolor}         
\usepackage{amsmath}
\usepackage{amssymb}
\usepackage{mathtools}
\usepackage{amsthm}
\usepackage{listings}
\usepackage{xcolor}
\usepackage{tikz}
\usepackage{cleveref}
\usepackage{wrapfig}
\usepackage[textsize=tiny]{todonotes}

\usepackage{amsmath,amsfonts,bm}

\def\Figref#1{Figure~\ref{#1}}

\def\eqref#1{equation~\ref{#1}}

\def\1{\bm{1}}

\def\vx{{\bm{x}}}

\DeclareMathAlphabet{\mathsfit}{\encodingdefault}{\sfdefault}{m}{sl}
\SetMathAlphabet{\mathsfit}{bold}{\encodingdefault}{\sfdefault}{bx}{n}

\newcommand{\Ls}{\mathcal{L}}

\DeclareMathOperator*{\argmin}{arg\,min}

\newcommand{\hp}{\vx}

\newcommand{\hpmaxiter}{T}

\newcommand{\obj}{y}

\newcommand{\history}[1]{h_{#1}}

\newcommand{\modelparams}{\theta}

\newcommand{\decode}{\phi^{-1}}

\newcommand{\encode}{\phi}

\newcommand{\statnumblackboxes}{3{,}095}
\newcommand{\statnumtrajectories}{557{,}100}
\newcommand{\statnumoptimizers}{6}
\newcommand{\statnumsearchspaces}{102}

\newcommand{\statnumtokens}{2.5B}
\newcommand{\ourdataset}{\texttt{BBO-Pile}}

\title{An Open-Source Training Dataset for Foundation Models for Hyperparameter Optimization}

\author{%
 Aaron~Klein\textsuperscript{$\star$} \\
 ELLIS Institute T\"ubingen\\
 \And
 Herilalaina~Rakotoarison\textsuperscript{$\star$}\\
 University of Helsinki \\
 \AND
 Luca~Thale-Bombien \\
 Leipzig University, ScaDS.AI\\
 \And
  David~Salinas\\
  ELLIS Institute T\"ubingen, Prior Labs\\
  \texttt{} \\ 
}

\begin{document}

\maketitle
\renewcommand{\thefootnote}{\fnsymbol{footnote}}
\newcommand{\rebuttal}[1]{\textcolor{blue}{#1}}
\footnotetext[1]{Core contributors. Correspondence to  \texttt{aaron.klein@tue.ellis.eu}}

\begin{abstract}
Most hyperparameter optimization methods require extensive tuning of their own hyperparameters, often limiting their ability to generalize across different benchmark families.
Foundation models for hyperparameter optimization that learn optimization principles from a large collection of optimization trajectories offer a promising alternative, with the potential to outperform manually designed methods across diverse problem classes. 
However, prior work has either relied on non-public datasets or on purely synthetic data, limiting reproducibility and generalization to real-world problems. 
As a result, progress in this area has been constrained by the lack of large-scale, real-world, publicly available pre-training data. We introduce \ourdataset, the first open-source dataset comprising over 500K optimization trajectories evaluated across $\statnumblackboxes$ different hyperparameter optimization tasks and global optimization problems for different optimizers, which represents by far the largest public dataset for this task.
Using this dataset, we train a family of foundation models at multiple scales, ranging from 2M to 80M parameters and from 200M to 2B training tokens, and study their scaling behavior with respect to compute. 
Our results demonstrate that large-scale pre-training is a viable and effective approach to imitate hyperparameter optimization methods, paving the way for future research in this direction.
\end{abstract}

\section{Introduction}

Hyperparameter optimization (HPO) is a fundamental problem in machine learning~\citep{feurer-automlbook18a, franceschi-ftml25}, and is closely related to algorithm configuration~\citep{hutter-lion14}. It is typically framed as a black-box optimization problem because
we have \textit{no} access to structural information about the objective function itself; we can only query its output and do not have access to gradient information. 

Given an objective function $f: \mathbb{X}^D \xrightarrow{} \mathbb{R},$ the task is to find the optimal configuration $\vx_{\star} \in \argmin_{\vx \in \mathbb{X}^D} f(\vx)$ in the search space $\vx \in \mathbb{X}^D$ that minimizes the objective. 
The search space $\mathbb{X}^D$ itself can be highly diverse, ranging from low to high-dimensional, and may consist of discrete, categorical, or numerical parameters.

Despite decades of research~\citep{storn-jgo97,jones-jgo01a}, existing black-box optimization methods remain specialized, with each performing well only within a narrow class of problems. 
Current state-of-the-art approaches are often the product of a laborious, manual process that is frequently based on trial-and-error. 
Moreover, these methods are typically optimized for specific use cases and do not generalize well across different domains. 
For example, Figure~\ref{fig:comparison} in Appendix~\ref{app:comparison} shows that the Bayesian optimization variant CQR~\citep{salinas-icml23} performs competitively on hyperparameter optimization benchmarks but performs poorly on continuous global optimization problems.

\begin{figure}[t]
    \centering
    \includegraphics[width=0.9\linewidth]{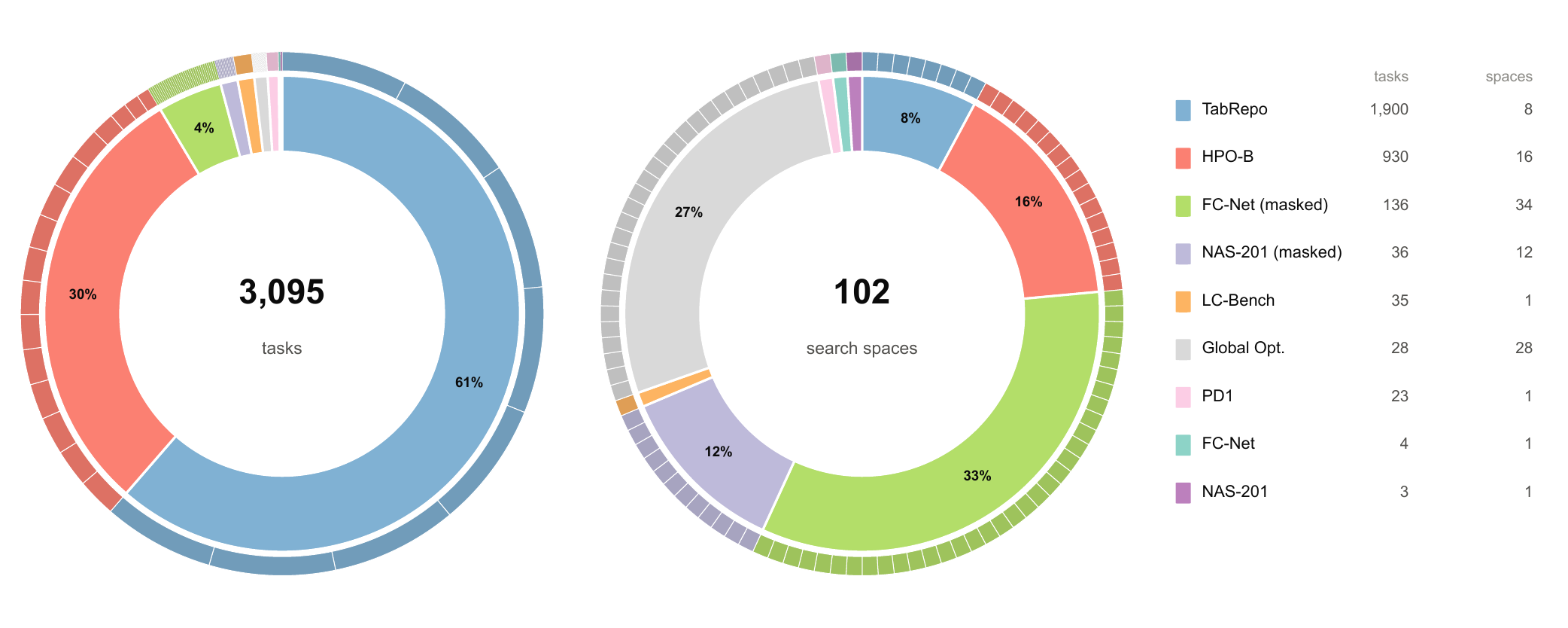}    
    \caption{Composition of our open-source dataset for pre-training foundation models for hyperparameter optimization. The dataset comprises $\statnumblackboxes$ tasks across $\statnumsearchspaces$ search spaces drawn from seven benchmarking families, including hyperparameter optimization, neural architecture search, and general black-box optimization benchmarks.}
    \label{fig:dataset_composition}
\end{figure}

Recently, foundation models that learn underlying principles purely from data have achieved remarkable success across domains such as computer vision~\citep{dosovitskiy-iclr21} and natural language processing~\citep{kaplan-arxiv19}, as well as in  structured settings like tabular data~\citep{hollmann-iclr23,hollmann-nature25} and time-series prediction~\citep{ansari-tmlr24}. 
Foundation models for hyperparameter optimization might be able to overcome the shortcomings of existing methods~\citep{song-icml24}.
For example, such models could, in principle, learn new optimization principles and encode different optimization strategies that are selected based on the properties of the search space.

OptFormer \citep{chen-neurips22}, one of the first works in this direction, treats optimization trajectories as sequences and trains auto-regressive encoder-decoder models on them. While this approach showed promising results, the pre-training data is \textit{not} publicly available, hindering reproducibility and the development of more advanced models.
Furthermore, compared to other domains such as vision~\citep{cherti2023} or text~\citep{kaplan-arxiv19}, it remains unclear how current approaches for training foundation models for black-box optimization scale with compute. Understanding this scaling behavior is essential for guiding the design of more powerful models as well as more expressive datasets.

Inspired by OptFormer, in this paper, we present the first fully open-source dataset, called \ourdataset, to train foundation models for hyperparameter optimization that paves the way for further research in this direction. More specifically, our contributions are as follows:

\begin{itemize}
    \item We provide the first large-scale open-source dataset for hyperparameter optimization consisting of $\statnumtrajectories$ optimization trajectories from $\statnumoptimizers$ different optimizers on $\statnumblackboxes$ HPO tasks and global optimization problems. The final dataset consists of $\sim \statnumtokens$ tokens and is available on our \href{https://huggingface.co/datasets/synetune/bbo-pile}{Hugging Face repository}. 
    
    \item We systematically analyze how well decoder-only transformers can imitate state-of-the-art hyperparameter optimization methods when parameter count and token budget are scaled by training a range of models from $2$M up to $80$M parameters and $200$M to $2$B tokens. Code to reproduce our experiments is available in our \href{https://github.com/syne-tune/bbo-pile}{GitHub repository}.
\end{itemize}

We discuss related work next in Section~\ref{sec:related_work}.
Section~\ref{sec:model} describes the training and inference process of our models and Section~\ref{sec:dataset} the generation process of our large-scale open-source dataset.
We present an empirical evaluation of our models across different scales in Section~\ref{sec:experiments}.
Section~\ref{sec:limitations} discusses limitations and we provide conclusions and overview of future work in Section~\ref{sec:discussion}.
\section{Related Work}\label{sec:related_work}

\paragraph{Black-Box Optimization:} 
One prominent approach for black-box optimization is \textit{Bayesian optimization} (BO) \citep{garnett_bayesoptbook_2023}, which is often applied to low-dimensional expensive-to-evaluate optimization problems. 
The central idea of BO is to model the unknown objective function $f(\vx)$ using a probabilistic surrogate model $p(f \mid D)$ trained on observed data $D$.
At each iteration, the surrogate model is used to construct an acquisition function that balances exploration and exploitation.
The next candidate is selected by optimizing the acquisition function, which is significantly cheaper to optimize than the true objective function.
Another prominent class of approaches is \textit{evolutionary algorithms}, which maintain and evolve a population of candidate solutions through repeated cycles of selection, mutation, and crossover \cite{talbi2009metaheuristics}. 
Mutations facilitate local exploration, while crossover operations recombine high-performing structures, thereby accelerating the search for strong candidates.

\paragraph{Foundation Models for Hyperparameter Optimization:} OptFormer~\citep{chen-neurips22} is one of the first families of foundation models for black-box optimization. It consists of encoder-decoder transformer architectures trained on offline collected optimization trajectories.
It provides models trained on optimization trajectories generated on two public benchmarking families (BBOB~\citep{hansen-arxiv16} and HPO-B~\citep{pineda-neurips21}) as well as one larger non-public dataset collected from the Google Vizier service.

\paragraph{Transfer Learning:} To accelerate the search process, transfer learning approaches leverage data from previous optimization runs to warm-start subsequent searches, for example, by reusing top-performing configurations from prior tasks as initial candidates~\citep{wistuba-ieee15} or by restricting the search space itself~\citep{perrone-neurips19}.
The flexibility of BO further allows modeling correlations across related tasks~\citep{springenberg-nips16,salinas-icml20} to start from a better prior of the objective function.
Despite their promise, most transfer learning methods for black-box optimization remain fundamentally limited.
They typically operate only within a \textit{fixed search space} and cannot generalize knowledge across different domains or problem settings.
In contrast, foundation models for black-box optimization aim to learn entire optimization strategies, enabling transfer not just within a single search space but also \textit{across domains}.

\paragraph{Black-Box Optimization with Large Language Models:} Recent work has begun exploring large language models (LLMs) for black-box optimization, mainly by leveraging their code-generation abilities rather than learning optimization directly.
For example, recent work evolved BO code \citep{li-arxiv25} and meta-heuristics \citep{vanstein-gecco25} using LLMs, or employed FunSearch~\citep{romera-nature24} to design new acquisition functions \citep{aglietti-arxiv25}.
While promising, these approaches only make incremental improvements to existing algorithms and remain \textit{confined} to narrow paradigms such as BO.
Other work~\citep{liu-iclr24} uses LLMs as models for BO directly, but frontier models are not built for optimization and struggle even with simple tasks like uniform sampling~\citep{schwanke-iclr26}, and fail to explore in a decision-making context~\citep{krishnamurthy-neurips24}.
\paragraph{Learning-to-Learn:} Foundation models for black-box optimization build on the long-standing idea of learning-to-learn~\citep{schmidhuber-diploma}, which seeks to discover new learning algorithms automatically~\citep{andrychowicz-neurips16,chen-icml17}.
Early approaches applied meta-learning or reinforcement learning to train recurrent networks that propose new candidate solutions for black-box optimization~\citep{chen-icml17}.
However, these models were limited to small architectures, short optimization horizons, and narrow problem types, preventing generalization beyond their training settings.
Moreover, their objectives often mimicked BO, biasing them toward rediscovering known strategies rather than inventing new ones.
\paragraph{Foundation Models for Structured Data:} Also related is the work on foundation models for structured datasets, such as tabular data~\citep{hollmann-iclr23,hollmann-nature25} or time series~\citep{ansari-tmlr24}.
While models such as TabPFN~\citep{hollmann-iclr23,hollmann-nature25} could, in principle, serve as surrogate models for BO~\citep{mueller-icml23}, they would also inherit the fundamental inefficiencies of BO, and not learn new optimization principles.
\section{Training on Optimizer Trajectories}\label{sec:model}

In what follows, we denote by $(\hp_t, \obj_t)_{t=1}^\hpmaxiter{}$ an optimization trajectory where $\hp_t \in \mathbb{X}^D$ is the input configuration and $\obj_t$ the observed objective at step $t$.

We first describe how optimizer trajectories are mapped to a sequence of tokens, how we train a decoder-only transformer model on such trajectories and the inference process to run optimization.

\subsection{Encoding and Tokenizing Optimizer Trajectories}
\label{sub:tokenization}

\begin{figure}
\lstset{
  basicstyle=\ttfamily\scriptsize,  
  breaklines=true,
  breakatwhitespace=false,
  breakindent=0pt,
  escapeinside={(*}{*)}
}
\begin{lstlisting}
(*\textcolor{blue}{<algorithm>:RS}*)
(*\textcolor{green}{<type>:<UNI>,<min\_value>:0.01,<max\_value>:1.0,<log-scale>\&}*)
(*\textcolor{green}{<type>:<UNI>,<min\_value>:1,<max\_value>:5,<linear-scale>\&}*)
(*\textcolor{green}{<type>:<CATEGORICAL>,<categories>:[0, 1]}*)
(*\textcolor{orange}{120,200,<1>*300}*)|(*\textcolor{red}{60,50,<0>*200}*)|
\end{lstlisting}
\caption{Illustration of the encoding of a trajectory for a search space
$\{"\text{a}": Log\mathcal{U}(0.01, 1.0), "\text{b}": \mathcal{U}(1, 5), "\text{c}": \{\text{"l1"}, \text{"l2"}\} \}$ optimized with random search for two trials.
The first line encodes the
\textcolor{blue}{optimizer}
used, the second line encodes the
\textcolor{green}{search space}
and the third line encodes the optimizer trajectory containing a \textcolor{orange}{first trial}
 and a
 \textcolor{red}{second one}. The encoding of trials is discussed in Section~\ref{sub:tokenization} and illustrated in Figure~\ref{fig:encoding_hp}.
\label{fig:encoding_meta}
}
\end{figure}

We largely follow the approach of Chen et al.~\citep{chen-neurips22} to encode and tokenize an optimization trajectory.  
First, we encode metadata including optimizer name and the search space as shown in \Figref{fig:encoding_meta}. 
However, in contrast to Chen et al.~\citep{chen-neurips22}, we do not include the names of the optimization task or values of categorical hyperparameters, in order to prevent the model from overfitting to this information.

We then encode each configuration and objective value $(\hp_t, \obj_t)$ with a string $s=\encode{(\hp_t, \obj_t)}$ as shown in \Figref{fig:encoding_hp}. 
To encode numerical configurations and objective values, we apply min-max scaling to $[0, 1]$ with values observed in the optimizer trajectory before discretizing values to be in $[0, Q-1]$ where $Q=1000$ in our experiments.  
For categorical parameters, we simply encode them with their index with $<i>$ to distinguish from numerical values.

To simplify the sampling process, we order configurations by placing numerical parameters first and categorical parameters last.
We use the special characters $\star$ and $\vert$ to indicate the end of the configuration and the end of the observed metric, respectively.
This allows us to map optimizer trajectories to strings. We then use Byte-Pair Encoding \cite{sennrich2016neural} to train our tokenizer.

\subsection{Training}

We train decoder-only transformer models based on the Qwen3 architecture~\cite{qwen3-arxiv25}, which uses Rotary Position Embeddings~\cite{su-2024}, Grouped Query Attention, and applies Root Mean Square Normalization to the query and key vectors to improve training stability. 
Given an encoded optimization trajectory, we optimize the standard causal language modeling objective: $\Ls(\theta) = -\sum_{i=1}^{n} \log p_\theta(s_i \mid s_{<i})$,

where $s_i$ denotes the $i$-th token in the sequence and $\theta$ are the model parameters. 

\subsection{Inference}\label{sub:inference}

\begin{figure*}

\centering







\begin{tikzpicture}[
    scale=0.76, transform shape,
    node distance=3.5cm,
    box/.style={rectangle, draw, minimum width=2cm, minimum height=0.8cm, align=center},
    arrow/.style={->, >=stealth, thick}
]

    \node (searchspace) at (0, 1.0) {\large \textbf{Search Space:} 
    \{\textcolor{blue}{"\texttt{dropout}": $Log\mathcal{U}(0.01, 1.0)$}, \textcolor{green}{"\texttt{alpha}": $\mathcal{U}(1.0, 5.0)$}, \textcolor{orange}{"\texttt{reg}": $\{\text{"l1"}, \text{"l2"}\}$ } \}
    };

    \node[align=center] (decoded) at (-3, 0) {
        \large
        $\hp{} = \{${\textcolor{blue}{"\texttt{dropout}": 0.12}},
        {\textcolor{green}{"\texttt{alpha}": 3.0}, {\textcolor{orange}{"\texttt{reg}": "l2"}}}$\}$ \quad
        \textcolor{red}{$\obj{}=0.3$}
    };

    \node[right=2cm of decoded] (s){
        \large s = \texttt{"}{\textcolor{blue}{\texttt{120}}}\texttt{,}{\textcolor{green}{\texttt{200}}}\texttt{,}{\textcolor{orange}{\texttt{<1>}}}\texttt{*}{\textcolor{red}{\texttt{300}}}\texttt{|"}
    };

    \draw[arrow] ([yshift=1.2mm]decoded.east) -- ([yshift=1.5mm]s.west) node[midway, above] {\footnotesize $\encode{}$};

    \draw[arrow] ([yshift=-1.2mm]s.west) -- ([yshift=-1.5mm]decoded.east) node[midway, below] {\footnotesize $\decode{}$};

\end{tikzpicture}
\caption{Illustration of the encoding and decoding of hyperparameter and objective values.}
\label{fig:encoding_hp} 
\end{figure*}

\paragraph{Sampling:} 
We sample a completion string from the trained model $s \sim p_\modelparams{}(\cdot|\history{})$ where $\history{}$ denotes the current encoded history consisting of the encoding of the search space and previous observations (see Figure~\ref{fig:encoding_meta} for an example). We perform constrained decoding using a grammar which ensures that all values are valid and can be decoded. 
To accelerate inference we deploy our model using vLLM, and provide a runtime comparison of our model in Appendix~\ref{app:runtime}.

\paragraph{Decoding:} To decode the string $s$ we use a decoding function $(\hp{}, \obj{}) = \decode{}(s)$. We map all integer values of continuous hyperparameters to their initial ranges and all categorical hyperparameters to their associated value and obtain the output value by decoding the output token with the same decoder used for continuous values.

\section{Dataset}\label{sec:dataset}

We now describe the construction and data collection of \ourdataset. 
Trajectories are collected by running a set of $\statnumoptimizers$ optimizers, five state-of-the-art algorithms plus random search, across a diverse collection of black-box benchmark families.
In total, our dataset contains \statnumtrajectories{} completed optimization runs across \statnumblackboxes{} black-box tasks (see Figure~\ref{fig:dataset_composition}).
To our knowledge, this is the largest publicly available dataset of black-box optimization trajectories spanning such a diverse set of benchmarks and optimizers (see Table~\ref{tab:hpo_benchmarks} in Appendix~\ref{app:comparison_datasets} for a comparison).

\subsection{Benchmark Families}

Because most black-box optimization problems are expensive to evaluate, generating a dataset of this scale would be prohibitively costly. 
Instead, we rely on offline datasets from the literature, which enable either table lookup or the construction of surrogate benchmarks to predict objective values for arbitrary configurations in the search space~\citep{eggensperger-aaai15}. 
We use a simple $k$-nearest-neighbor regressor for all benchmarks to minimize the introduction of additional modeling bias.
Note that, while surrogate benchmarks are standard practice in AutoML and have been shown to preserve the relative ranking of optimizers~\citep{eggensperger-aaai15,zela-iclr22}, their fidelity is limited by the density of the underlying offline evaluations, and configurations far from evaluated points may be assigned inaccurate objective values.

We include four public hyperparameter optimization benchmarks: HPO-B~\citep{pineda-neurips21}, LC-Bench~\citep{zimmer-ieee21}, PD1~\citep{wang2024pre}, and TabRepo~\citep{salinas2023tabrepo}\footnote{We remove $28$ tasks from TabRepo since all configurations in the search space achieve identical performance.}. 
In addition, we include two neural architecture search benchmarks, FC-Net~\citep{klein-arxiv19a} and NAS-Bench-201~\citep{dong-iclr20}, as well as 28 synthetic global optimization problems.

To further expand the diversity of our dataset, we apply hyperparameter masking to the FC-Net and NAS-Bench-201 benchmark families, to create two new families 'masked FC-Net' and 'masked NAS-Bench-201'. Specifically, we mask up to two hyperparameters  within their respective search spaces. The masked hyperparameters are fixed to constant values. We fix this constant as the marginal best value, which is, the value that yields the best mean performance when marginalizing over all other hyperparameters. This masking procedure introduces an additional $172$ distinct black-box tasks.

Overall, the dataset comprises \statnumblackboxes{} black-box tasks spanning \statnumsearchspaces{} distinct search spaces, which include numerical, integer, and categorical input variables. 
We note, however, that the dataset is dominated by TabRepo and HPO-B, which together account for $91.4\%$ of all tasks, and that 46 of the 102 search spaces are masked variants of FC-Net and NAS-Bench-201.
Per-family statistics are reported in Table~\ref{tab:validation_splits}.
Figure~\ref{fig:dataset_composition} summarizes the composition of the dataset. A detailed description of each benchmark family is provided in Appendix~\ref{app:benchmarks}.

\subsection{Methods and Evaluation Protocol} 

We consider different black-box optimization families, such as BO, evolutionary algorithms and random search~\citep{bergstra-jmlr12a} to construct the optimization trajectories.
For BO, we include the following methods that have shown strong performance on these benchmarks: BORE~\citep{tiao-icm21}, CQR~\citep{salinas-icml23}, HEBO~\citep{cowen2020empirical}, and TPE~\citep{bergstra-nips11a}. 
We include the evolutionary method regularized evolution (REA)~\citep{real-aaai19} which often performs competitively on these benchmarks. 
We do not include methods such as CMA-ES~\citep{hansen-eda06} or Differential Evolution~\citep{storn-jgo97}, as they typically require larger evaluation budgets than the 100 evaluations considered here to become competitive.
We use the implementations from the Syne Tune~\citep{salinas-automl22} library.
A more detailed description of each method is presented in Appendix~\ref{app:methods}.

Each optimizer runs for the same budget of $T=100$ evaluations on every task.
For each (optimizer, task) pair we run 30 repetitions with a different seed.
This leads to  $\text{\#total-runs} = \#\text{optimizers} \times \#\text{tasks} \times \#\text{seeds}$, which gives \statnumtrajectories{} unique optimization trajectories.
We ran all optimizers on CPUs, resulting in an estimated total of 50,595 CPU hours.

\subsection{Data Augmentation}\label{subsec:augmentation}

We employ simple data augmentation strategies to mitigate overfitting and expand our token count. With these augmentations, the dataset comprises a total of $\statnumtokens$ tokens.

\textbf{Permutation:} Following Chen et al.~\citep{chen-neurips22}, we augment the dataset by permuting the order of hyperparameter configurations. To maintain consistency, we preserve the convention of listing numerical parameters before categorical ones. For a search space with $N$ numerical hyperparameters and $C$ categorical hyperparameters, there are up to $N! \times C!$ possible permutations. The specific number of permutations sampled for each benchmark family is detailed in Table~\ref{tab:validation_splits}. 

\textbf{Trajectory length:} Due to the quantization process described above, the distribution of objective values shifts throughout the optimization. 
For example, during inference at $T=2$ iterations, the performance of the superior trial is mapped to $0$, while the other is mapped to $Q-1$. 
However, during training, the model will only observe full trajectories, which normally occur only at the end of optimization. 
To mitigate this discrepancy, we sample shorter trajectories of varying lengths $T \in \{5,10,20,50,100\}$ prior to quantization.

\subsection{Train/Validation Splits}
We use a fixed train/validation split to evaluate the generalization capabilities of our model. 
To generate the validation split, we consider the following two evaluation settings:
\begin{enumerate}
\item \textbf{Generalization to unseen tasks within seen search spaces.}
To measure this setting, we hold out 60 tasks across all benchmark families to form the validation set. All trajectories originating from a single task are assigned to the same split to avoid information leakage.
\item \textbf{Generalization to unseen search spaces.}
To evaluate this setting, we evaluate search spaces not seen during training on TabRepo, HPO-B, and the synthetic global optimization benchmarks. We consider those benchmarks since they contain multiple search spaces allowing us to reserve some for validation. 
\end{enumerate}

Table~\ref{tab:validation_splits} shows the number of tasks per benchmark family that are used for validation.
These two settings capture complementary generalization challenges: adapting to new instances within familiar search spaces and transferring knowledge to entirely novel search spaces.
Overall, the validation set comprises approximately $10\%$ of the total number of tasks.

\begin{table}[t]
\scriptsize
\centering
\caption{Validation split composition for the two generalization settings.}
\label{tab:validation_splits}
\begin{tabular}{lccccc}
\toprule
\textbf{Benchmark Family} 
& \textbf{\# Tasks} 
& \textbf{\# Search Spaces}
& \textbf{\# Held-out Tasks} 
& \textbf{\# Held-out Search Spaces}
& \textbf{\# Permutations} \\
\midrule
FC-Net                & 4 & 1 & 1   & -- & 20\\
Masked FC-Net         & 136 & 34 & --  & -- & 20\\
NAS-Bench-201         & 3 & 1 & 1   & -- & 20\\
Masked NAS-Bench-201  & 36 & 12 & --  & -- & 4\\
LC-Bench              & 35 & 1 & 5   & -- & 20\\
PD1                   & 23 & 1 & 3   & -- & 4\\
HPO-B                 & 930 & 16 & 15  & 1 (44 tasks) & 0\\
TabRepo               & 1900 & 8 & 35  & 1 (241 tasks) & 0\\
Global Optimization   & 28 & 28 & -- & 5 (1 task each) & 0\\
\midrule
\textbf{Total}  & 3095 & 102   & 60  & 7 (249 tasks) & --\\
\bottomrule
\end{tabular}
\end{table}

\section{Experiments}\label{sec:experiments}

We now comprehensively evaluate our dataset by training a range of models across different token and parameter scales.
Section~\ref{sub:compute} discusses the scaling behavior of our model under varying compute budgets. 
We evaluate how well our models imitate the optimization trajectories of the original optimizers on a range of validation tasks in Section~\ref{sub:eval}.

Code for training and deploying our model, as well as for generating the training data, is available on our \href{https://github.com/syne-tune/bbo-pile}{GitHub Repo}. The dataset and all model checkpoints are available on our \href{https://huggingface.co/datasets/synetune/bbo-pile}{Hugging Face repository}. 

We rely on the LitGPT library~\citep{litgpt-2023} for model implementation and 
\texttt{PyTorch-Lightning}\footnote{\url{https://github.com/Lightning-AI/pytorch-lightning}} for efficient training. 
All training runs are performed on NVIDIA H100 GPUs.
We estimate a total compute of 1872 GPUh to train all models in our grid.
We use AdamW~\citep{loshchilov-iclr19} with $\beta_1=0.9$, $\beta_2=0.95$, $\text{weight\_decay}=0.1$, gradient clipping with $\text{max\_norm}=1.0$, and a cosine learning rate schedule with $10\%$ linear warm-up. 
All models are trained with \texttt{bf16-mixed}, without gradient accumulation, and a context length of 4096 tokens.

\subsection{Scaling Compute}\label{sub:compute}

\begin{figure*}[t]
    \centering
    \begin{subfigure}[h]{0.45\textwidth}
        \centering
        \includegraphics[width=0.9\linewidth]{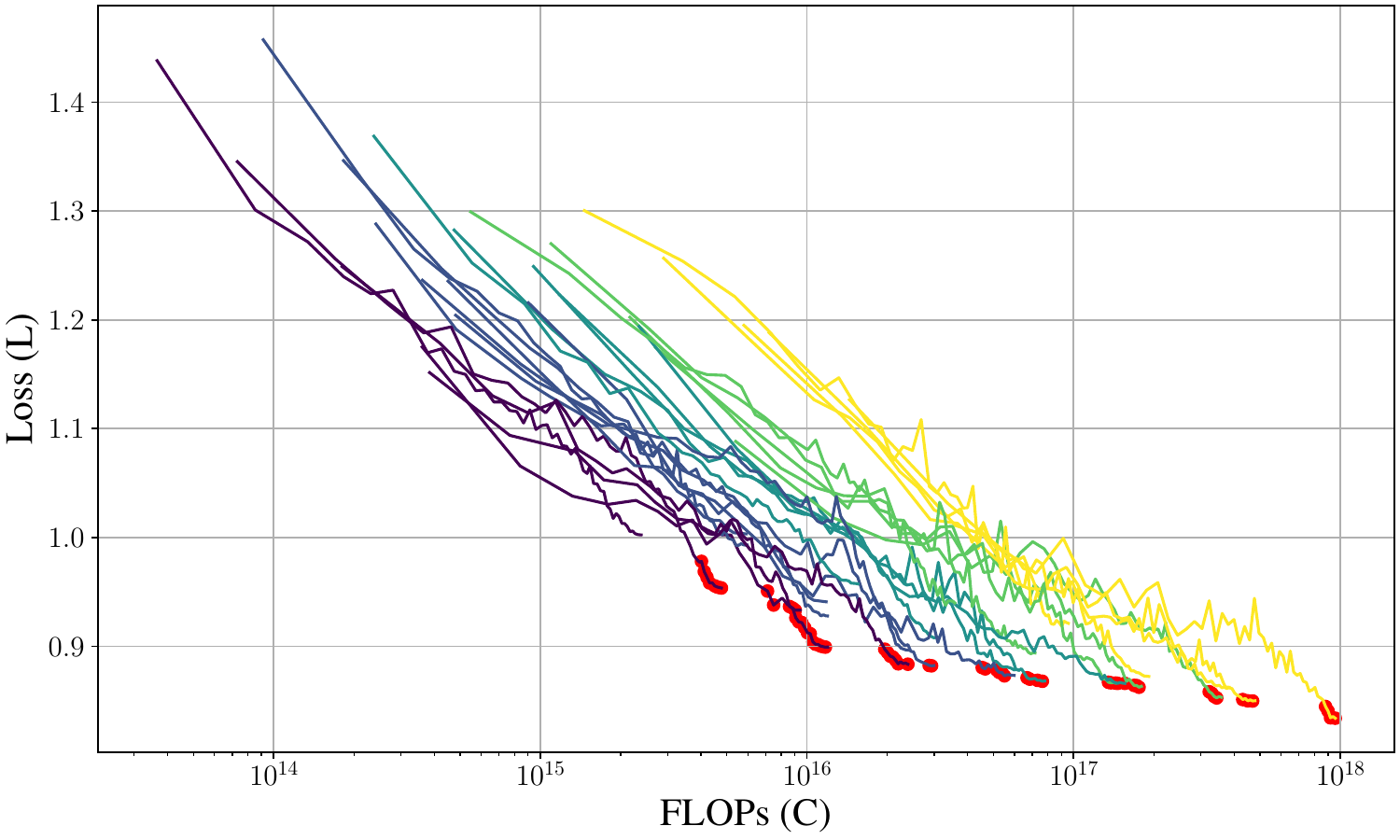}
        \caption{Learning Curves}   
    \end{subfigure}  
     \begin{subfigure}[h]{0.45\textwidth}
        \centering
        \includegraphics[width=0.9\linewidth]{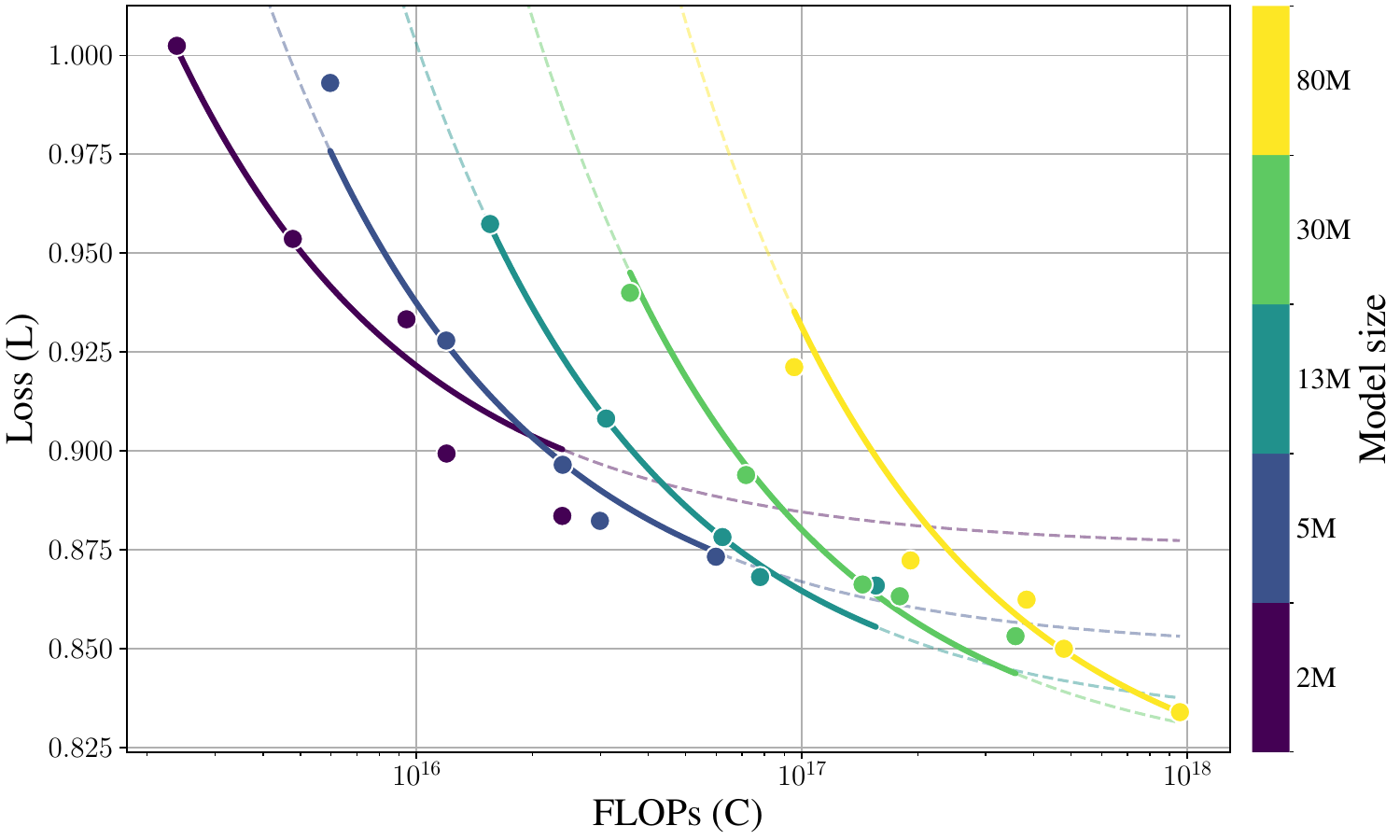}
        \caption{$N$/$D$ Scaling Law}   
    \end{subfigure}  
    \caption{\textbf{Left:} Validation loss curves of each parameter count N / token budget D pair across FLOPs $C \approx 6 \times N \times D$. We select the model with the best learning rate and batch size according to our grid search. Color indicates parameter count and red dots mark Pareto optimality after initial convergence phase.
    \textbf{Right:} Final validation loss of each $(N, D)$ pair (dots) and the fitted $L(N, D)$ for each model size (lines; dashed lines extrapolate beyond the trained token budgets).}
    \label{fig:scaling}
\end{figure*}

We train a range of models with different parameter counts ($N$), ranging from
$2$M to $80$M parameters (see Table~\ref{tab:model_architectures} in
Appendix~\ref{app:model_grid}). Each model is trained under different token
budgets ($D$) spanning from $200$M to $2$B tokens. For each $(N, D)$ pair, we
select the optimal learning rate ($LR$) and global batch size ($GBS$) by
performing a simple grid search over $LR \in \{5e^{-3}, 1e^{-2}, 2e^{-2}\}$ and
$GBS \in \{4, 8, 16\}$ (see Figure~\ref{fig:hp_grid} in
Appendix~\ref{app:model_grid}).

In Figure~\ref{fig:scaling} (left), we show the validation loss of the best
configuration for each $(N, D)$ pair. We observe higher stochasticity compared
to LLMs due to the nature of optimization search trajectories. To study the
scaling behavior~\cite{kaplan-arxiv19}, we fit a relationship between validation
loss and compute, using Pareto-dominant points from the validation curves.
Figure~\ref{fig:scaling_fit} in Appendix~\ref{app:model_grid} shows
that this relationship is well-captured by a power law, which also extrapolates
accurately to our largest configuration (80M parameters trained on 2B tokens).
Note, however, that the fitted exponent ($0.0157$) is shallower than those
typically reported for LLM pre-training~\cite{hoffmann-arxiv22}, which indicates
loss improvements from increasing compute, through model size or token budget,
are modest. The small exponent might be due to the redundancies in the
data, induced by our data augmentation strategy (Section~\ref{subsec:augmentation}) that
comes with the relatively high number of repetitions. We also note that we only
consider about 2.6 orders of magnitude of compute, which is substantially
smaller than what is typically used for LLMs (e.g., 8 orders of magnitude).

We additionally perform a Chinchilla-style decomposition separating
the parameter ($N$) and data ($D$) exponents. Following Hoffmann et
al.~\citep{hoffmann-arxiv22}, we fit $L(N, D) = E + A N^{-\alpha} + B
D^{-\beta}$ to the final loss of every $(N, D)$ run and obtain $L = 0.79 +
21.4\,N^{-0.38} + 9.0 \times 10^{4}\,D^{-0.71}$ (Figure~\ref{fig:scaling},
right). Since the data exponent exceeds the parameter exponent ($\beta >
\alpha$), the fit suggests that, within the range we study, increasing the token
budget reduces the loss more than increasing model capacity. However, these
estimates are obtained over a narrow compute range and may not extrapolate to
the scales typical of LLM pre-training.

\subsection{Imitation of Optimizers}\label{sub:eval}

\begin{figure*}[t]
    \centering
    \begin{subfigure}[h]{0.24\textwidth}
        \centering
        \includegraphics[width=.99\linewidth]{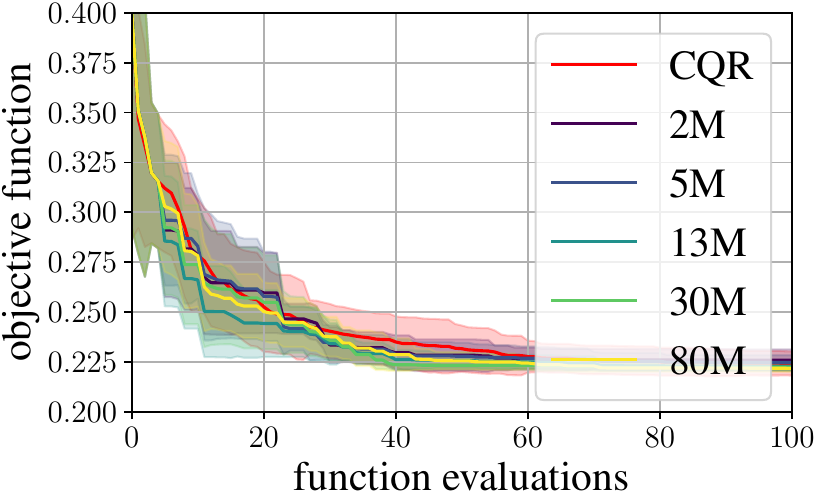}
        \caption{\tiny{FC-Net Protein}}   
    \end{subfigure}  
    \begin{subfigure}[h]{0.24\textwidth}
        \centering
        \includegraphics[width=.99\linewidth]{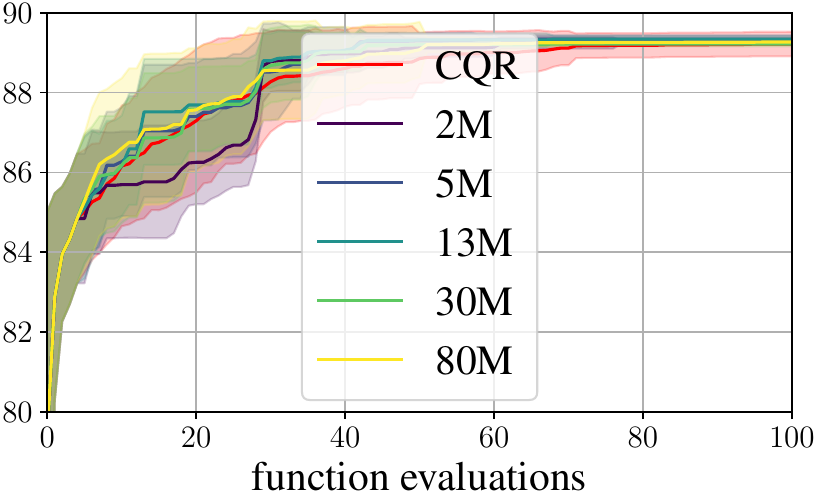}
        \caption{\tiny{LC-Bench F-MNIST}}   
    \end{subfigure}  
    \begin{subfigure}[h]{0.24\textwidth}
        \centering
        \includegraphics[width=.99\linewidth]{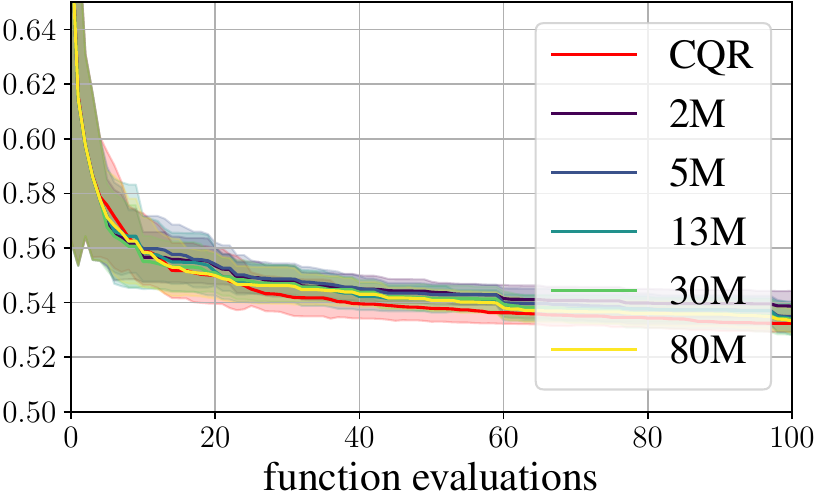}
        \caption{\tiny{NAS-Bench-201 ImageNet16-120}}   
    \end{subfigure}  
    \begin{subfigure}[h]{0.24\textwidth}
        \centering
        \includegraphics[width=.99\linewidth]{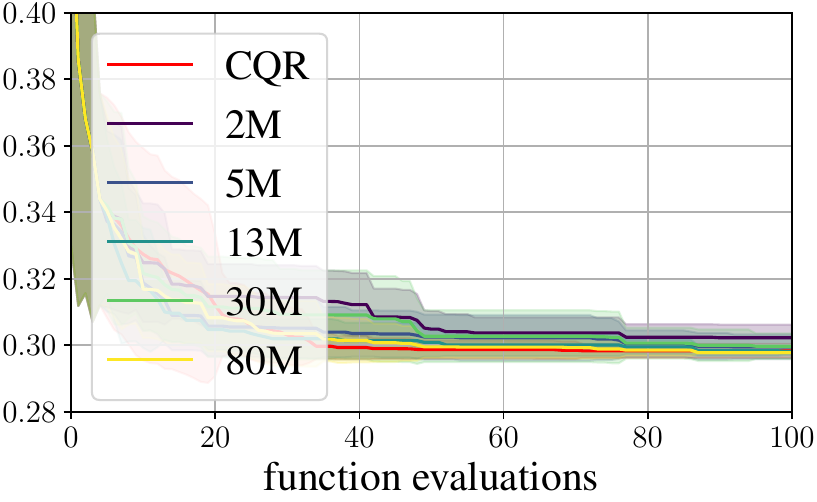}
        \caption{\tiny{TabRepo CatBoost F-MNIST}}   
    \end{subfigure}  

    \begin{subfigure}[h]{0.24\textwidth}
        \centering
        \includegraphics[width=.99\linewidth]{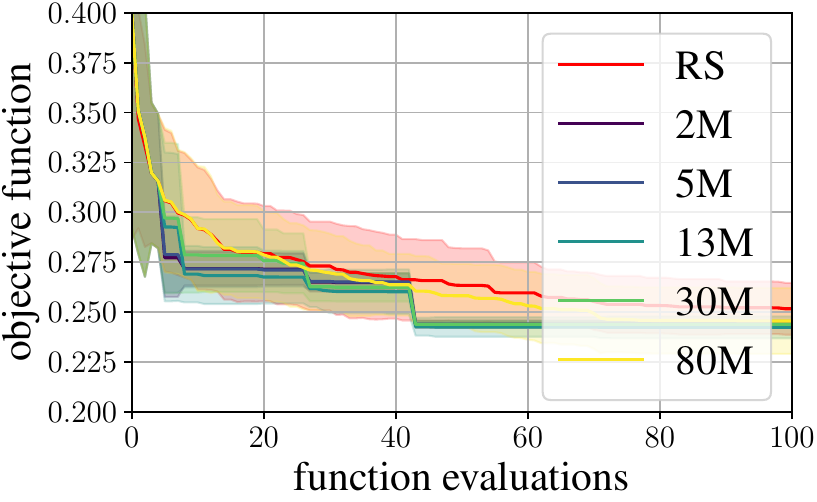}
        \caption{\tiny{FC-Net Protein}}   
    \end{subfigure}  
    \begin{subfigure}[h]{0.24\textwidth}
        \centering
        \includegraphics[width=.99\linewidth]{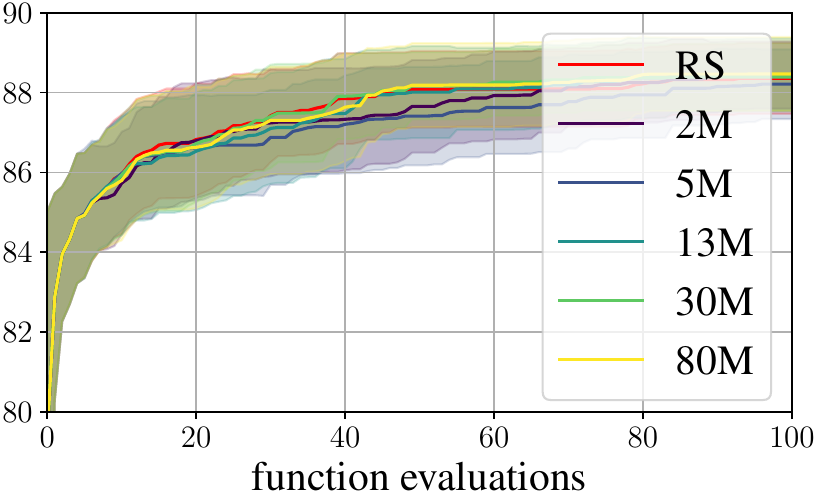}
        \caption{\tiny{LC-Bench F-MNIST}}   
    \end{subfigure}  
    \begin{subfigure}[h]{0.24\textwidth}
        \centering
        \includegraphics[width=.99\linewidth]{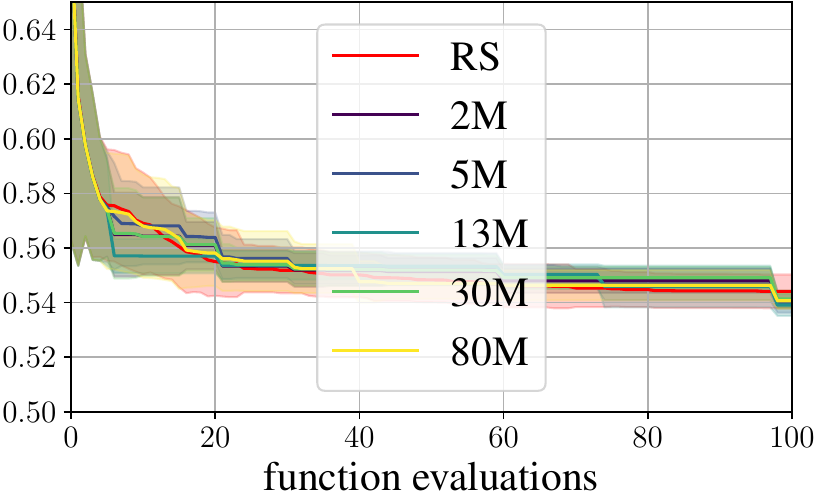}
        \caption{\tiny{NAS-Bench-201 ImageNet16-120}}
    \end{subfigure}  
    \begin{subfigure}[h]{0.24\textwidth}
        \centering
        \includegraphics[width=.99\linewidth]{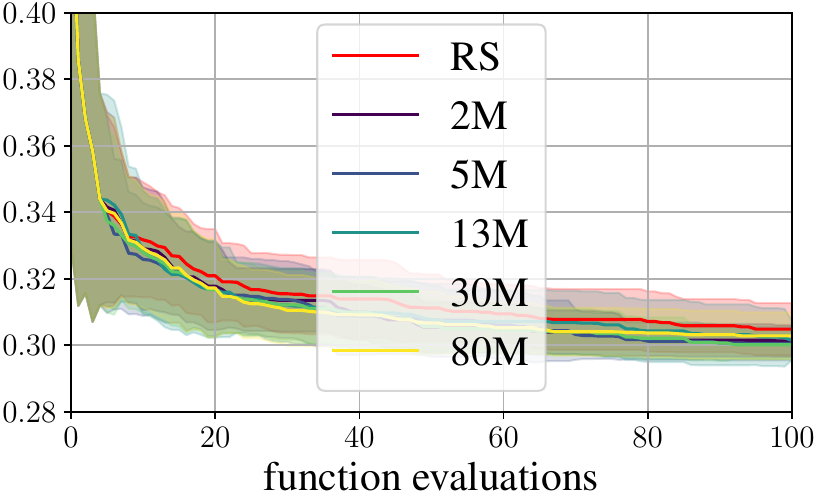}
        \caption{\tiny{TabRepo CatBoost F-MNIST}}
    \end{subfigure}  
    
    \caption{Comparison of the original CQR / RS method with CQR / RS simulated by our models at \textit{different parameter} scales. 
Figures~(a–c) / (e-g) show results on tasks with search spaces seen during training, while Figure~(d) / (h) reports performance on a task with an unseen search space (TabRepo).}
    \label{fig:imitation_parameters}
\end{figure*}

\begin{figure*}[t]
    \centering
    \begin{subfigure}[h]{0.24\textwidth}
        \centering
        \includegraphics[width=.99\linewidth]{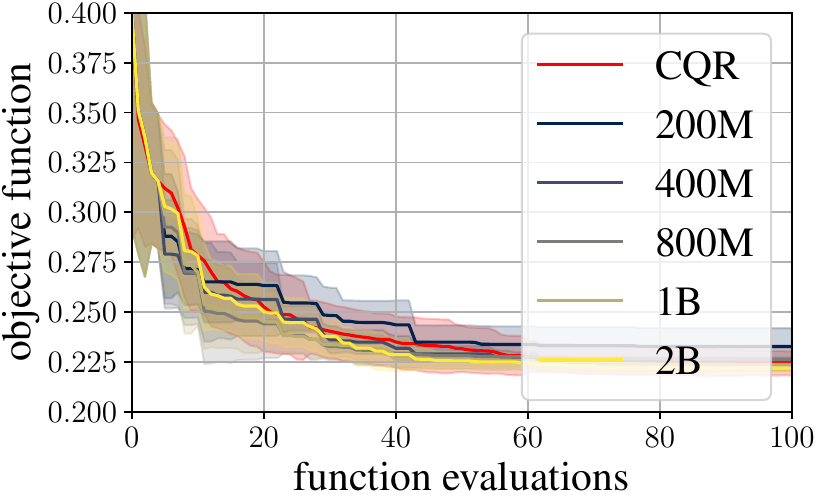}
        \caption{\tiny{FC-Net Protein}}
    \end{subfigure}  
    \begin{subfigure}[h]{0.24\textwidth}
        \centering
        \includegraphics[width=.99\linewidth]{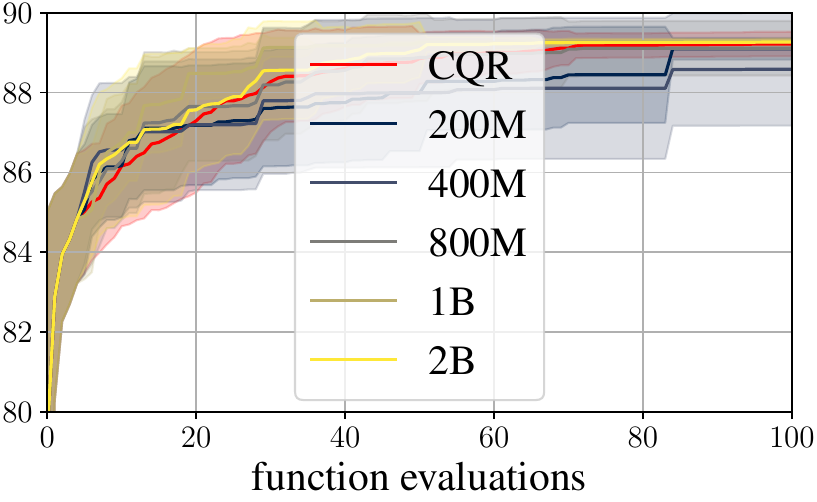}
        \caption{\tiny{LC-Bench F-MNIST}}   
    \end{subfigure}  
    \begin{subfigure}[h]{0.24\textwidth}
        \centering
        \includegraphics[width=.99\linewidth]{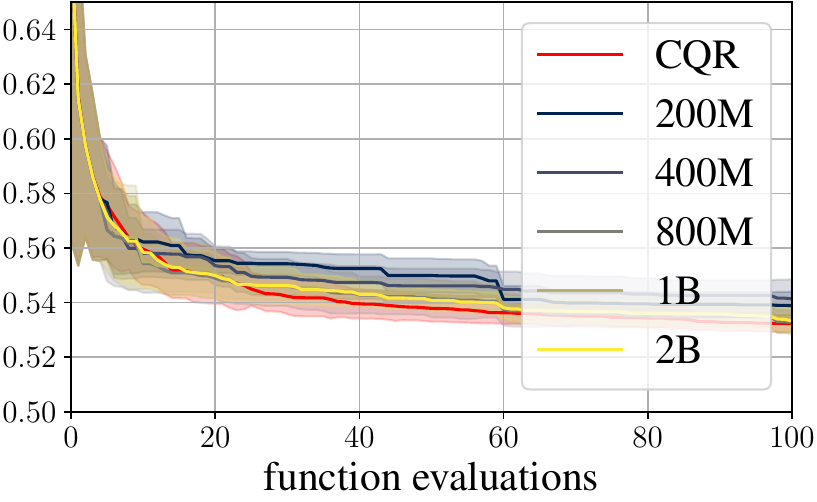}
        \caption{\tiny{NAS-Bench-201 ImageNet16-120}}   
    \end{subfigure}  
    \begin{subfigure}[h]{0.24\textwidth}
        \centering
        \includegraphics[width=.99\linewidth]{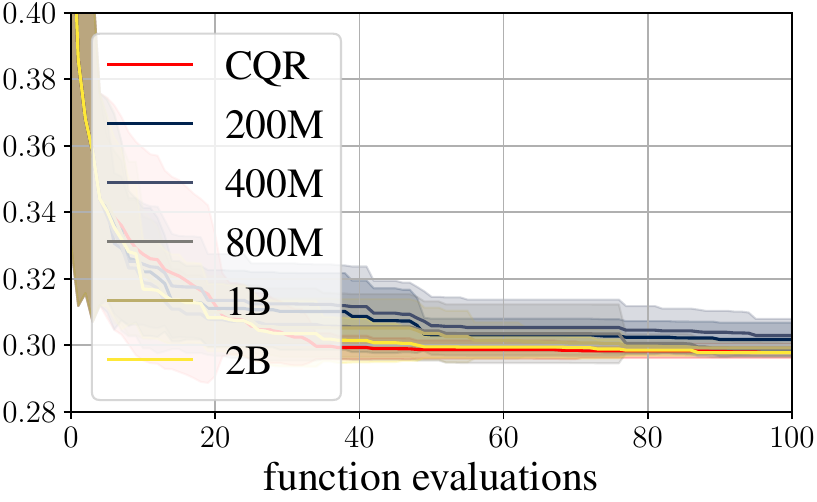}
        \caption{\tiny{TabRepo CatBoost F-MNIST}}   
    \end{subfigure}  

   \begin{subfigure}[h]{0.24\textwidth}
       \centering
       \includegraphics[width=.99\linewidth]{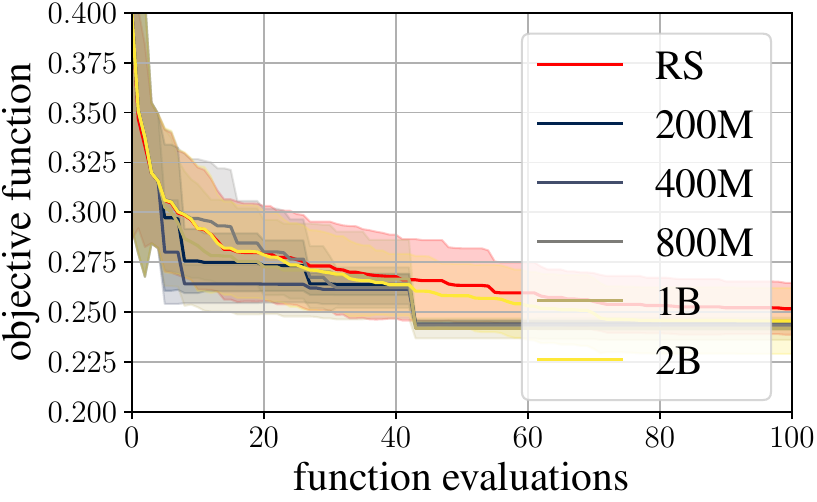}
       \caption{\tiny{FC-Net Protein}}   
   \end{subfigure}  
   \begin{subfigure}[h]{0.24\textwidth}
       \centering
       \includegraphics[width=.99\linewidth]{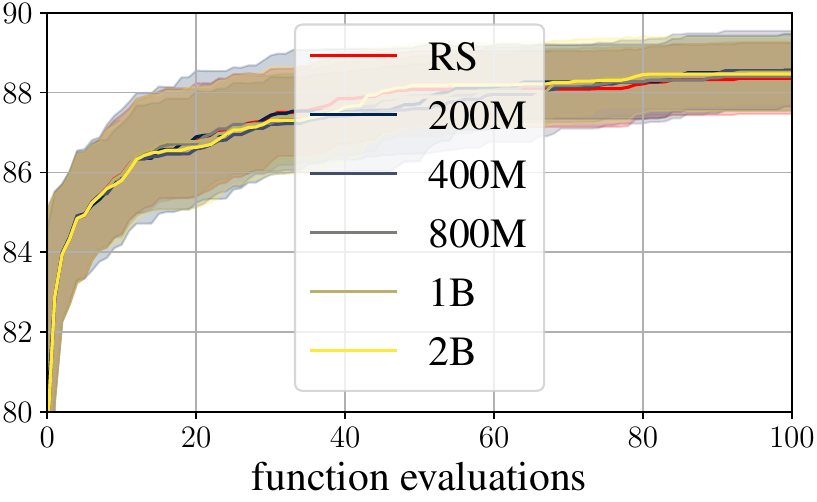}
       \caption{\tiny{LC-Bench F-MNIST}}   
   \end{subfigure}  
   \begin{subfigure}[h]{0.24\textwidth}
       \centering
       \includegraphics[width=.99\linewidth]{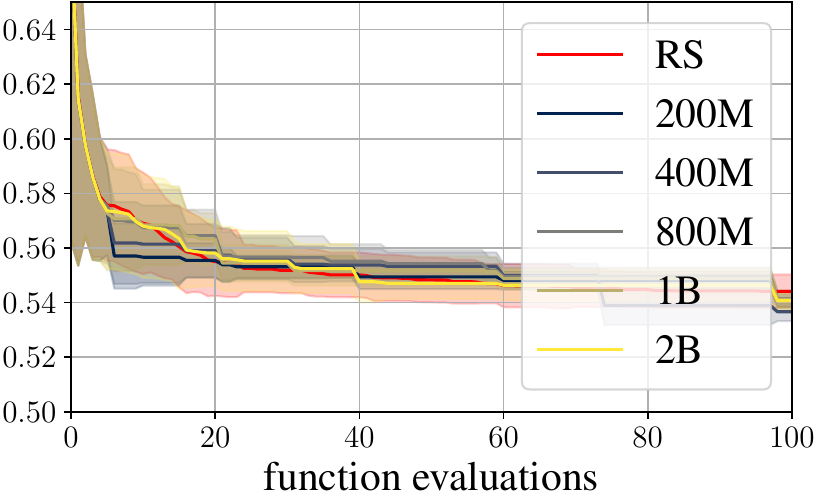}
       \caption{\tiny{NAS-Bench-201 ImageNet16-120}}   
   \end{subfigure}  
   \begin{subfigure}[h]{0.24\textwidth}
       \centering
       \includegraphics[width=.99\linewidth]{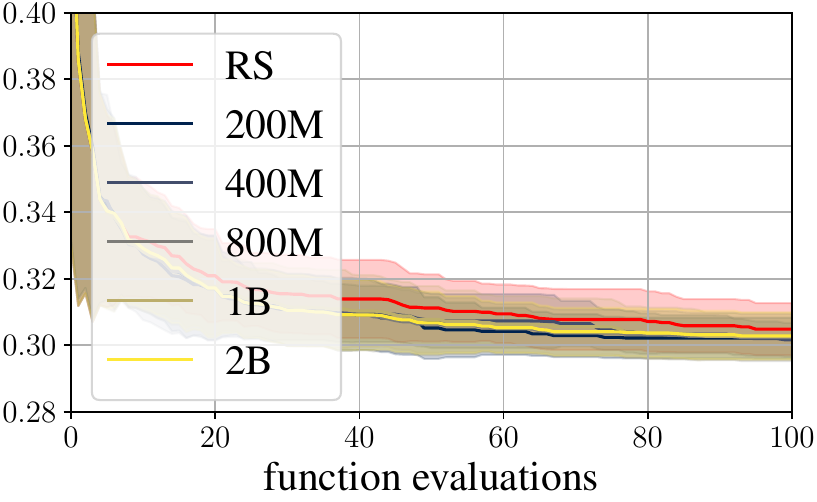}
       \caption{\tiny{TabRepo CatBoost F-MNIST}}   
   \end{subfigure}  
   
    \caption{Comparison of the original CQR / RS method with CQR / RS simulated by our models at different \textit{token budgets}. 
    Figures~(a–c) / (e-g) show results on tasks with search spaces seen during training, while Figure~(d) / (h) reports performance on a task with an unseen search space (TabRepo).}
    \label{fig:imitation_tokens}
\end{figure*}

\begin{figure*}[t]
    \centering
   \begin{subfigure}[h]{0.24\textwidth}
       \centering
       \includegraphics[width=.99\linewidth]{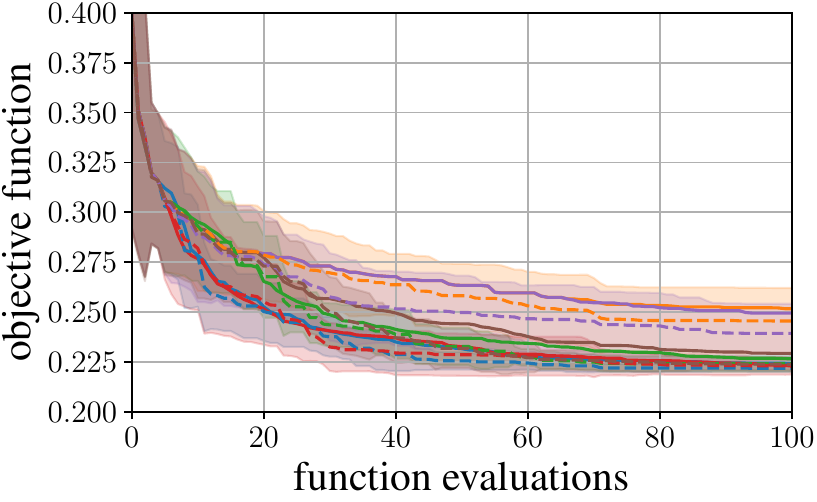}
       \caption{\tiny{FC-Net Protein}}   
   \end{subfigure}  
   \begin{subfigure}[h]{0.24\textwidth}
       \centering
       \includegraphics[width=.99\linewidth]{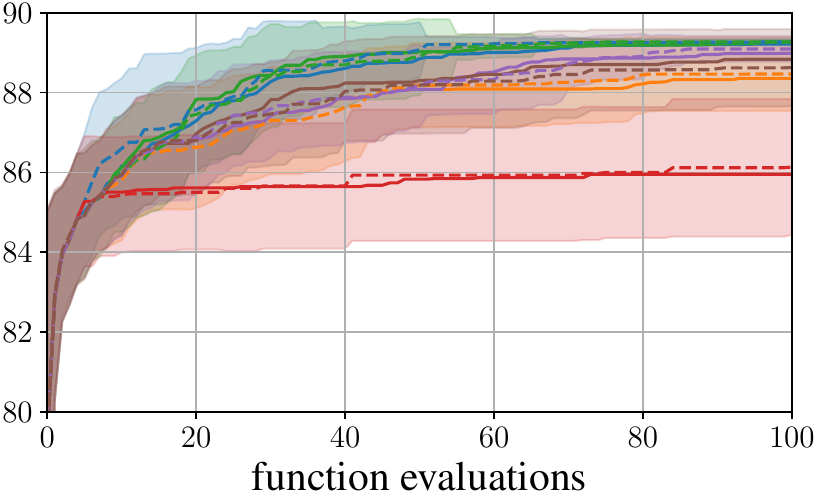}
       \caption{\tiny{LC-Bench F-MNIST}}   
   \end{subfigure}  
   \begin{subfigure}[h]{0.24\textwidth}
       \centering
       \includegraphics[width=.99\linewidth]{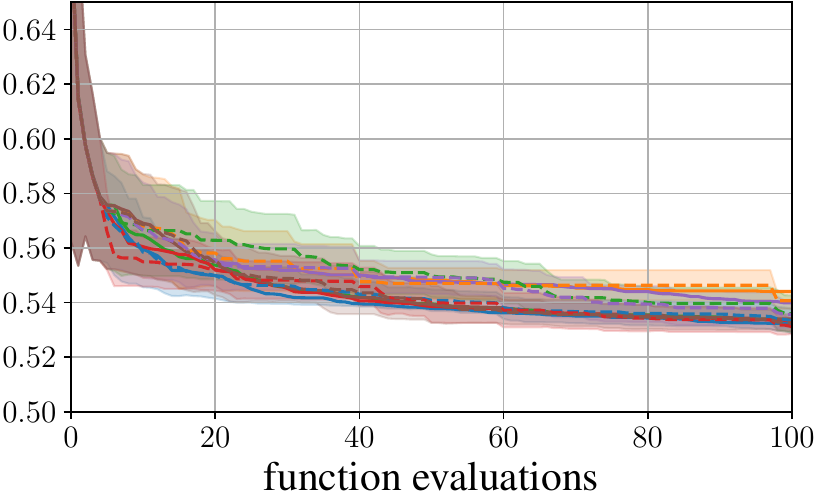}
       \caption{\tiny{NAS-Bench-201 ImageNet16-120}}   
   \end{subfigure}  
   \begin{subfigure}[h]{0.24\textwidth}
       \centering
       \includegraphics[width=.99\linewidth]{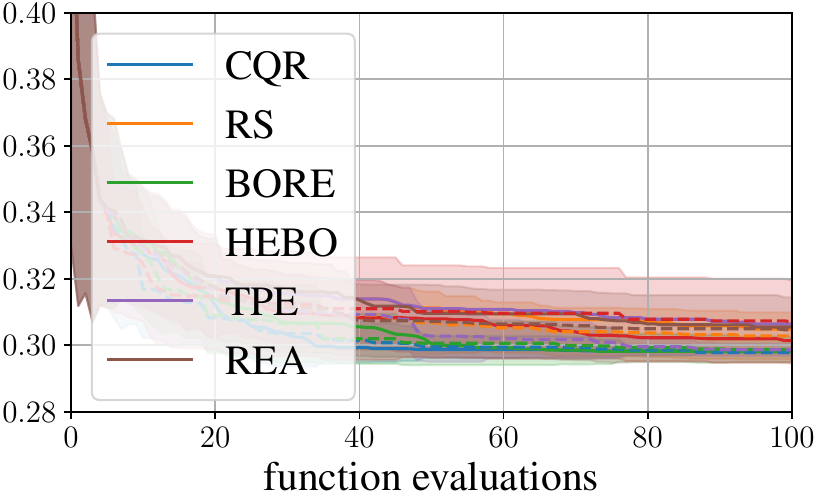}
       \caption{\tiny{TabRepo CatBoost F-MNIST}}   
   \end{subfigure}  
   
    \caption{ 
    Our 80M model (dashed lines) vs.\ original optimizers (solid lines) on tasks with search spaces seen during training (a–c) and an unseen search space, TabRepo (d).}
    \label{fig:imitation_all}
\end{figure*}

\begin{figure*}[t]
    \centering
   \begin{subfigure}[h]{0.24\textwidth}
       \centering
       \includegraphics[width=.99\linewidth]{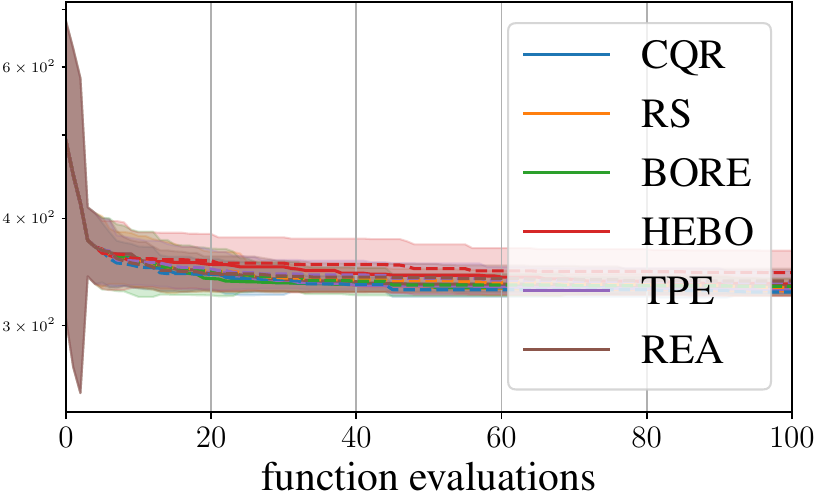}
       \caption{\tiny{DeepAR Electricity}}   
   \end{subfigure}  
   \begin{subfigure}[h]{0.24\textwidth}
       \centering
       \includegraphics[width=.99\linewidth]{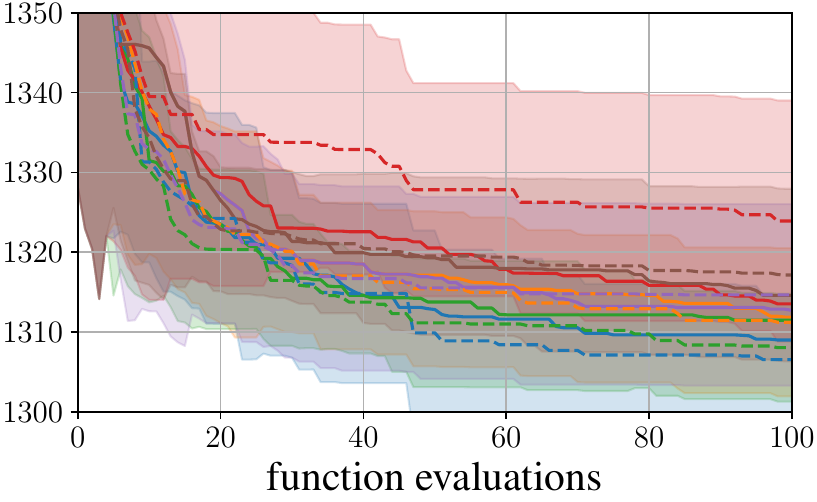}
       \caption{\tiny{DeepAR M4-Quarterly}}   
   \end{subfigure}  
   \begin{subfigure}[h]{0.24\textwidth}
       \centering
       \includegraphics[width=.99\linewidth]{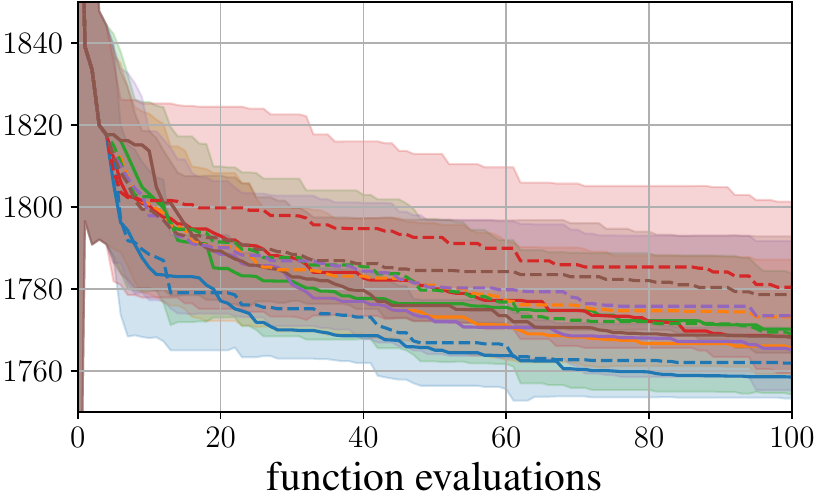}
       \caption{\tiny{DeepAR M4-Yearly}}   
   \end{subfigure}  
   \begin{subfigure}[h]{0.24\textwidth}
       \centering
       \includegraphics[width=.99\linewidth]{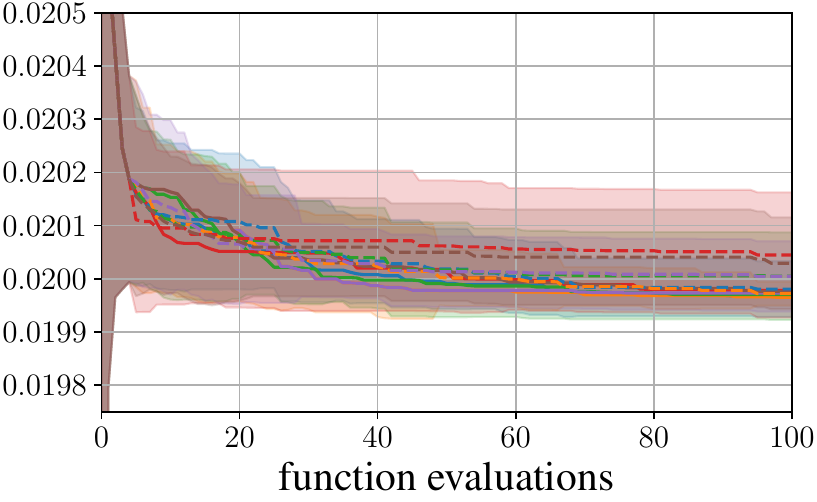}
       \caption{\tiny{DeepAR Traffic}}   
   \end{subfigure}  
   
    \caption{ 
    Comparison of optimizers on a completely unseen benchmark family (DeepAR).}
    \label{fig:imitation_deepar}
\end{figure*}

We validate models trained on \ourdataset{} by assessing how closely the optimization trajectories generated by our models match those of the original optimizers. 
As discussed in Section~\ref{sec:model}, we only change the meta-information in the prompt to switch between different optimization methods.

\textbf{Imitation across scale:} First, we analyze how well our models reproduce the optimization trajectories of the original methods as we scale parameter count and token budget. 
For each (N, D) pair, we select the model from the grid search with the lowest final validation loss.
We focus on CQR, as it performs well across most hyperparameter optimization benchmarks (see Appendix~\ref{app:comparison}) and exhibits a non-trivial sampling distribution. 
Additionally, we include RS, which, despite its simplicity, differs substantially from CQR, allowing us to assess whether the models can distinguish between distinct optimization policies.

Figure~\ref{fig:imitation_parameters} shows the mean and standard deviation over 30 runs of the original CQR/RS methods (red) and of our models at different parameter scales on three unseen tasks from search spaces included in the training data (FC-Net Protein, LC-Bench Fashion-MNIST, NAS-Bench-201 ImageNet16-120), as well as on one task from an unseen search space (TabRepo CatBoost-Fashion-MNIST). 
We present results on other tasks in Appendix~\ref{app:imitation}.
We observe that even smaller models are able to capture the behavior of CQR and RS, but performance is matched more closely with larger parameter counts.
For example, on LC-Bench Fashion-MNIST, a model with $2M$ parameters is insufficient to fully match the performance of CQR, particularly during the early iterations.

Figure~\ref{fig:imitation_tokens} shows the same pattern when scaling the token budget for the $80$M model. 
A budget below $800$M tokens seems to be insufficient to fully capture the sampling distribution of CQR on some benchmarks, e.g., TabRepo CatBoost F-MNIST (see Figure~\ref{fig:imitation_tokens} (d)).
However, models trained for 1B or 2B tokens closely match the original performance also on this benchmark.
Aggregate ranks and regrets over all optimizers, reported in Appendix~\ref{app:eval_scale}, confirm both trends.

\textbf{Comparison of different optimization methods}: Figure~\ref{fig:imitation_all} compares the performance of different optimizers imitated by our 80M model trained on 2B tokens (dashed lines) with the performance of the original methods (solid lines).
To characterize the behavior and coverage of the original optimizers, Appendix~\ref{app:coverage} reports the fraction of their evaluations that fall into each decile of the objective values on FC-Net and NAS-Bench-201.
Overall, the imitated optimizers achieve performance that is often close to that of the original methods, both on unseen tasks within known search spaces and on tasks from previously unseen search spaces in TabRepo. However, the agreement is weaker for unseen search spaces in global optimization problems (see Figure~\ref{fig:unseen_search_spaces} (i–l) in Appendix~\ref{app:unseen_search_spaces}).
We observe that these tasks exhibit substantially higher variation in their loss landscapes, which likely contributes to the performance gap. 
Future work could address this limitation by extending the data augmentation strategy to improve generalization in such settings.

 \textbf{Generalization to an unseen benchmark family}: We evaluate our approach on an unseen benchmark family consisting of hyperparameter optimization tasks for DeepAR~\citep{salinas2020deepar} on time-series datasets~\citep{salinas-icml20} (see Figure~\ref{fig:imitation_deepar}). 
Additional results for more tasks are provided in Appendix~\ref{app:imitation_test}.
Although the trajectories match the original methods less closely than in previous experiments, clear differences between optimization strategies remain observable, and our model achieves a final performance comparable to that of the original methods.

\begin{figure}
    \centering
    \begin{subfigure}[h]{\textwidth}
       \centering
       \includegraphics[width=\linewidth]{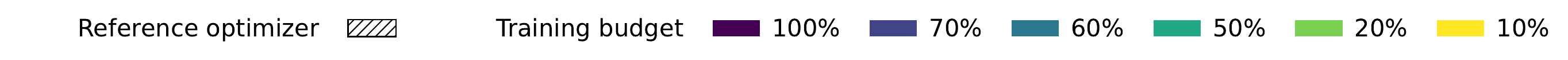}
       \includegraphics[trim=0 35 0 10,clip,width=\linewidth]{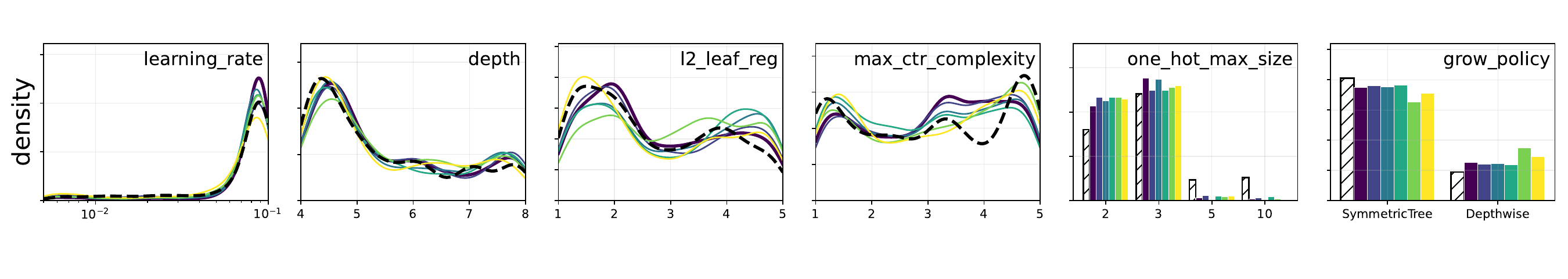}
       \caption{Imitation of BORE on TabRepo search space (CatBoost, Fashion-MNIST)}\label{fig:sampling_dist:bore_tabrepo} 
   \end{subfigure}
   \begin{subfigure}[h]{\textwidth}
       \centering
        \includegraphics[width=\linewidth]{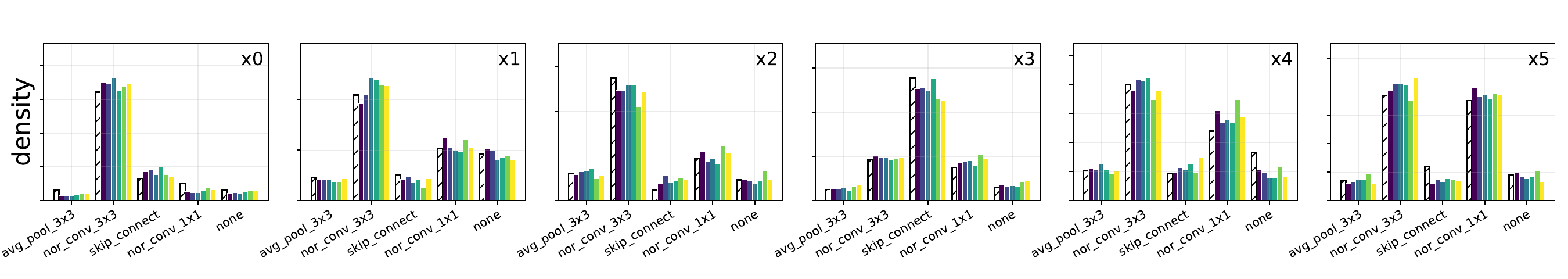}
       \caption{Imitation of CQR on NAS-Bench-201 search space (ImageNet 16-120)}\label{fig:sampling_dist:cqr_nas201}
   \end{subfigure}

   \caption{Comparison of estimated per-hyperparameter densities between our model and each reference optimizer. Our model is shown at different training stages (100\% = $2$B tokens).}\label{fig:sampling_dist}
\end{figure}

\textbf{Per-hyperparameter density comparison.} We additionally compare the marginal distribution over hyperparameters induced by our model relative to a reference optimizer. Given a set of observations ($T=40$) generated offline, we condition both the reference optimizer and our model on this same history, draw 500 samples from each method, and estimate the marginal density of every hyperparameter. Figure~\ref{fig:sampling_dist} qualitatively compares these densities for two reference optimizers on two benchmarks (TabRepo with BORE, NAS-Bench-201 with CQR). We observe that, as training progresses, the estimated densities of our model increasingly align with those of the corresponding reference optimizer. See Appendix~\ref{app:sampling_distributions} for detailed comparisons of an extended set of optimizers and search spaces.

\textbf{Inference cost.} Using vLLM, our 80M model requires roughly 0.5 seconds per proposed configuration on an H100 GPU, about twice as fast as the original CQR implementation on CPU, while the runtime per trial remains constant over the optimization horizon (see Appendix~\ref{app:runtime}). Since this cost is independent of the imitated method, our models can potentially amortize the overhead of more expensive optimizers, for example Bayesian optimization with Gaussian processes or costly acquisition function optimization.
\section{Limitations}\label{sec:limitations}

The current dataset has several limitations. 
First, it primarily focuses on the AutoML domain, which provides the largest collection of publicly available benchmarks. However, these benchmarks often exhibit specific characteristics, such as single modality or high observation noise, that may limit generalization to other domains, such as chemical design or computational fluid dynamics.

Second, we considered mostly real-world tasks, but leveraging synthetic optimization tasks (for example, sampled from a generative model~\citep{klein-nips19}) will likely be useful to boost further performance.
Synthetic data has been proven effective for tabular and time-series foundation models \cite{hollmann-iclr23,hollmann-nature25,ansari-tmlr24} as it allows one to further improve coverage and diversity. 

Third, our current models are trained only on trajectories with up to 100 trials. 
While this is typically sufficient for expensive black-box problems such as hyperparameter optimization or neural architecture search, it constitutes a limitation for cheap-to-evaluate problems, such as those commonly used in benchmarking suites like BBOB~\citep{hansen-arxiv16}.

Fourth, our scaling analysis covers only about 2.6 orders of magnitude of compute and models of up to 80M parameters, and the fitted compute exponent is substantially shallower than those reported for LLM pre-training, so our scaling estimates may not extrapolate to larger scales. We attribute this partly to redundancy in the data introduced by our data augmentation strategy and the high number of repetitions per optimizer and task (Section~\ref{subsec:augmentation}), which future versions of the dataset should reduce by adding more diverse tasks and optimizers.

Finally, the coverage of optimization algorithms remains limited. Incorporating additional algorithm classes, such as CMA-ES~\citep{hansen-eda06} or Differential Evolution~\citep{storn-jgo97}, would expose the models to a broader range of optimization principles. 
However, this would require extending the current trajectory format to represent population states.

\section{Discussion and Future Work}\label{sec:discussion}

We present our open-source dataset \ourdataset{} for training foundation models for hyperparameter optimization and analyze model performance under varying compute budgets. 
We believe this work opens several promising directions for future research. 
The current models are bounded by the performance of the original optimizers. Exceeding them would require post-training, for example to learn new optimization principles or to automatically select the most suitable optimization strategy for a given search space.
Moreover, since the model can generate full optimization trajectories, future work could explore reasoning-based or test-time scaling approaches to overcome the myopic limitations of sequential optimization methods.

\section*{Acknowledgements}

Aaron Klein and David Salinas acknowledge support by the EC under the grant No. 101195233 (OpenEuroLLM).
The authors gratefully acknowledge the computing time made available to them on the high-performance computer at the NHR Center of TU Dresden. This center is jointly supported by the Federal Ministry of Research, Technology and Space of Germany and the state governments participating in the NHR. Herilalaina Rakotoarison acknowledges compute resources funded by the state of Baden-Württemberg through bwHPC and the German Research Foundation (DFG) under grant numbers 455622343 (bwForCluster NEMO 2) and 539134284 through EFRE (FEIH2698644). Luca Thale-Bombien was further supported by the BMFTR through a scholarship of DAAD project 57616814 (SECAI, School of Embedded and Composite AI) as part of the program Konrad Zuse Schools of Excellence in Artificial Intelligence.

\begin{center}
\includegraphics[width=0.3\textwidth]{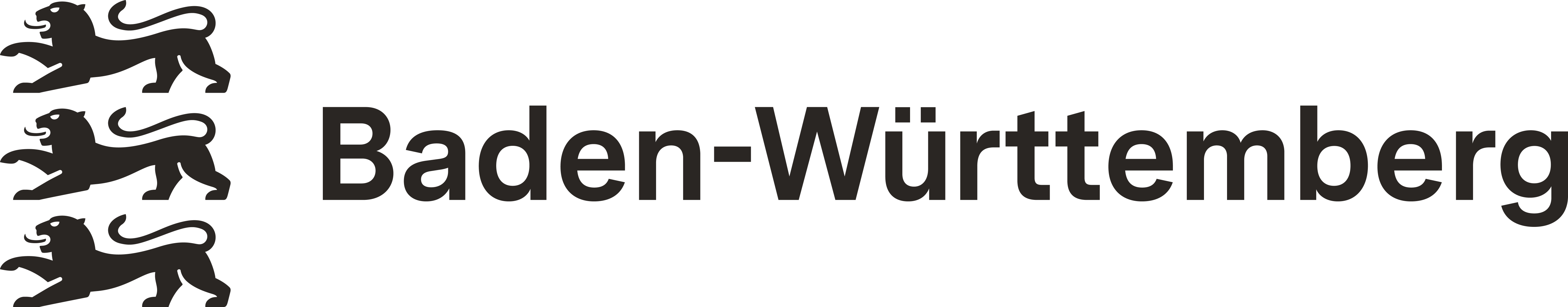} ~~~ \includegraphics[width=0.3\textwidth]{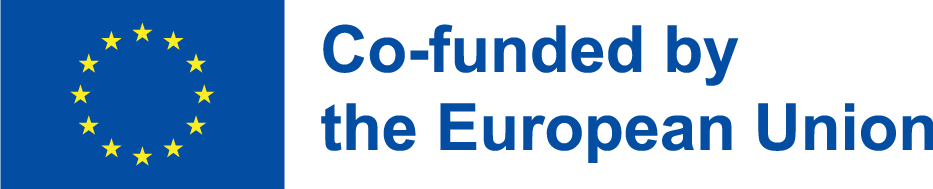} ~~~ \includegraphics[width=0.15\textwidth]{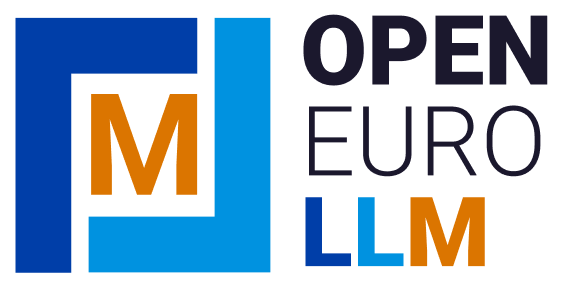} 
\end{center}

\bibliography{local,lib}
\bibliographystyle{abbrv}

\clearpage
\appendix
\section{Benchmark Families}\label{app:benchmarks}

We consider the following benchmark families from the literature:
\begin{itemize}
    \item \textbf{FC-Net}~\citep{klein-arxiv19a}: Considers the optimization of the hyperparameters and architecture details of a fully connected neural network trained on regression problems.
    \item \textbf{NAS-Bench-201}~\citep{dong-iclr20}: Neural architecture search benchmark to optimize the architecture of convolutional neural networks for image classification problems.
    \item \textbf{PD1}~\citep{wang2024pre}: The PD1 Neural Net Tuning Dataset contains Nesterov-momentum training runs across 23 neural-network tuning tasks.
    \item \textbf{LC-Bench}~\citep{zimmer-ieee21}: Contains training data for different architectures and hyperparameters evaluated on OpenML datasets.
    \item \textbf{HPO-B}~\citep{pineda-neurips21}: Large-scale benchmark, constructed from the OpenML repository. We use HPO-B-v2 with 16 different search spaces and
6,347,916 evaluations on 101 datasets.
    \item \textbf{Global Optimization Problems}: This benchmark defines 28 noise-free, real-valued, single-objective functions designed to capture typical difficulties in continuous optimization and emphasize challenging, interpretable problem landscapes.
    \item \textbf{TabRepo}~\citep{salinas2023tabrepo}:  Tabular model evaluations containing the predictions and metrics of 1310 models evaluated on 241 classification and regression tasks.
\end{itemize}

\section{Comparison to other Hyperparameter and Black-Box Optimization Datasets}\label{app:comparison_datasets}
Table~\ref{tab:hpo_benchmarks} shows a comparison of \ourdataset{} to other existing datasets.
\begin{table}[h]
    \footnotesize
    \centering
    \caption{Overview of black-box optimization trajectory datasets. \label{tab:dataset_stats}}
    \label{tab:hpo_benchmarks}
    \begin{tabular}{l c c}
        \hline
        Dataset & \textbf{\# Search Spaces} & \textbf{\# Tasks}  \\
        \hline
        BBOB \cite{hansen-arxiv16}$^\dagger$ & 1 & 24 \\
        RBV2 \cite{binder2020collecting} & 7 & 796 \\

        HPO-B-v2 \cite{pineda-neurips21} & 16 & 101 \\
        YAHPO Gym \cite{pfisterer-automl22} & 14 & $\sim$700 \\        
        OptFormer \cite{chen-neurips22} & Unknown & $\sim$750,000 \\
        BBO-Pile & \statnumsearchspaces{} & \statnumblackboxes{} \\
        \hline
    \end{tabular}
    \par\smallskip
    {\footnotesize $^\dagger$BBOB defines a single search space (continuous box $[-5,5]^d$) across all dimensions. We therefore count it as one search space rather than one per dimensionality evaluated.}
\end{table}

\section{Description of Hyperparameter Optimizers}\label{app:methods}

We consider the following hyperparameter optimization algorithms to construct our dataset:

\begin{itemize}
    \item \textbf{BORE}~\citep{tiao-icm21}: Bayesian Optimization by Density-Ratio Estimation reframes the hyperparameter optimization process as a classification problem. A probabilistic classifier is trained to score configurations by how likely they are to be high-performing.
    \item \textbf{CQR}~\citep{salinas-icml23}: Conformalized Quantile Regression serves as a surrogate model for Bayesian Optimization that captures heteroskedastic objective noise by estimating conditional quantiles instead of assuming fixed Gaussian observation noise. To ensure reliability under finite data, it applies split conformal prediction to adjust these intervals with an empirical offset $\gamma_{j}$ providing coverage guarantees for the predicted performance. These calibrated quantiles then facilitate efficient Thompson sampling, where the next configuration is selected by picking the candidate with the lowest randomly sampled quantile value.
    \item \textbf{HEBO}~\citep{cowen2020empirical}: Heteroscedastic and Evolutionary Bayesian Optimization is designed to overcome the limitations of standard Gaussian Processes, such as the assumptions of homoscedasticity and stationarity. It employs non-linear input and output warping to handle non-constant noise and skewed objective distributions. HEBO uses an evolutionary algorithm for candidate selection, to optimize a multi-objective ensemble of acquisition functions, identifying a Pareto-front of configurations that balance different exploration-exploitation trade-offs.
    \item \textbf{REA}~\citep{real-aaai19}: Regularized Evolution is a population-based search algorithm that modifies traditional tournament selection by introducing \textit{aging evolution}. Rather than discarding the worst-performing individuals, REA removes the oldest configurations from the population to prevent premature convergence through better search-space exploration.
    \item \textbf{TPE}~\citep{bergstra-nips11a}: Tree-structured Parzen Estimator is a Bayesian optimization algorithm that models the relationship between hyperparameters and performance by density estimation rather than direct regression. Instead of modeling $p(y|x)$, TPE uses kernel density estimators to model two distributions: $l(x) = p(x|y < y^*)$ for \textit{good} configurations and $g(x) = p(x|y \ge y^*)$ for \textit{bad} ones, split by a quantile threshold $y^*$. The next configuration is selected by maximizing the ratio $l(x)/g(x)$, which is mathematically monotonically related to the Expected Improvement (EI).
    \item \textbf{RS}~\citep{bergstra-jmlr12a}: Defines a uniform distribution over the search space from which samples are drawn repeatedly.
\end{itemize}

\section{Comparison of Optimizers}\label{app:comparison}

\begin{figure*}[t!]
    \centering
    \begin{subfigure}[h]{0.49\textwidth}
        \centering
        \includegraphics[width=0.49\textwidth]{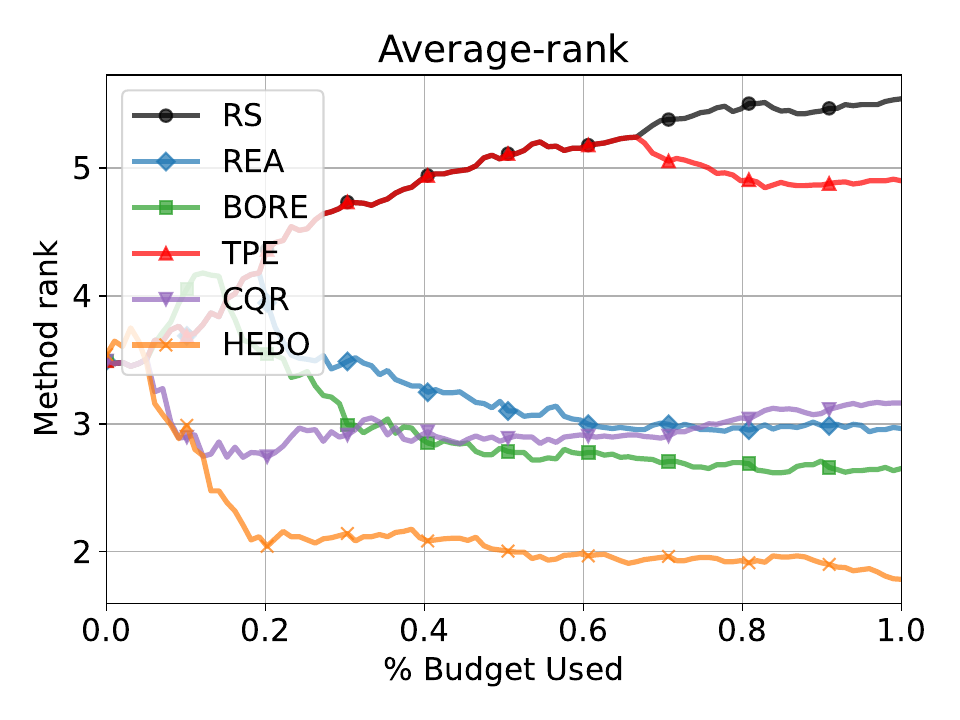}
        \includegraphics[width=0.49\textwidth]{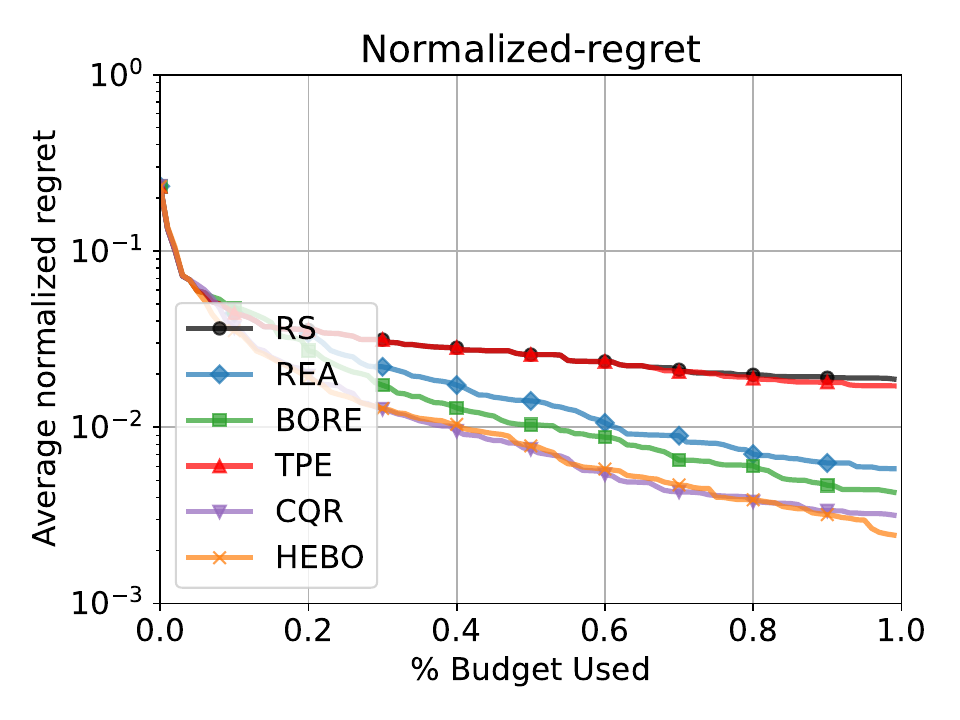}
        \caption{FC-Net}
    \end{subfigure} 
    \begin{subfigure}[h]{0.49\textwidth}
    \centering
        \includegraphics[width=0.49\textwidth]{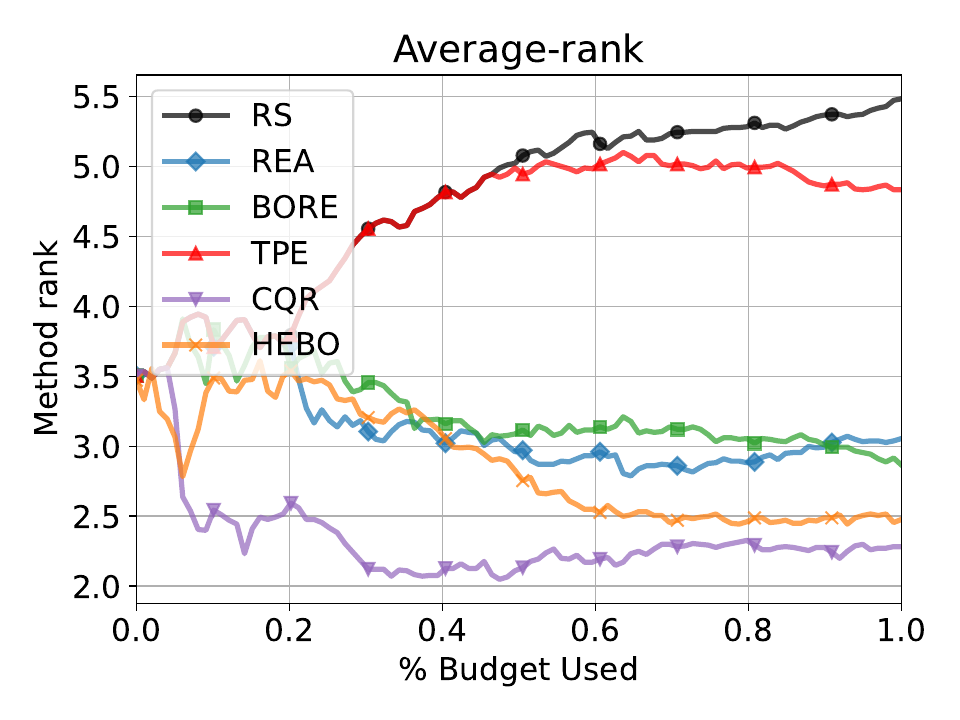}
        \includegraphics[width=0.49\textwidth]{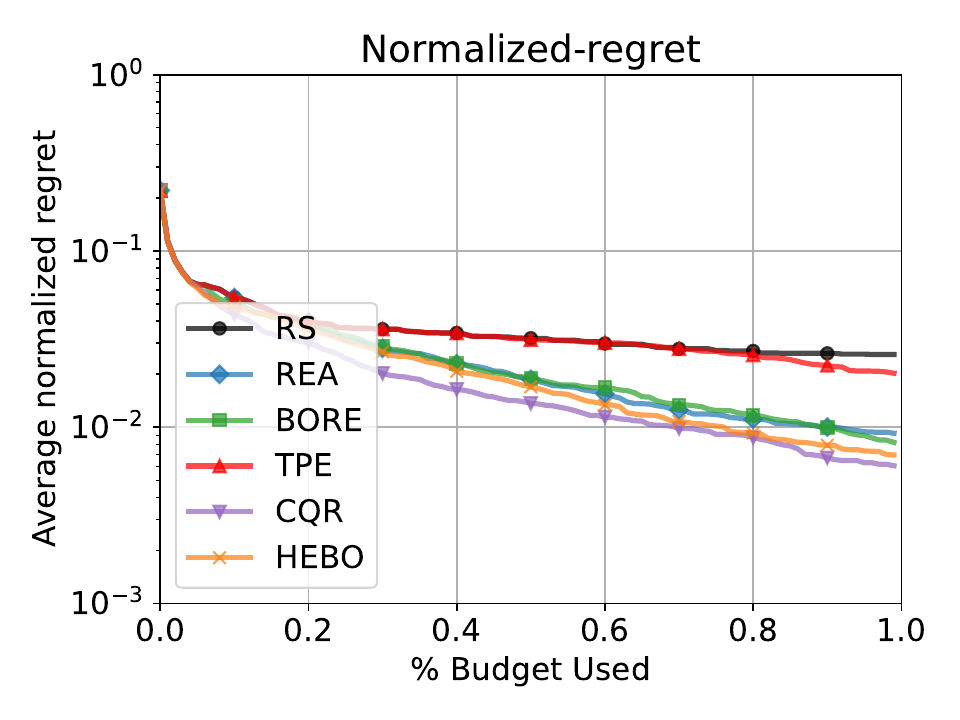}
        \caption{NAS-Bench-201}
    \end{subfigure}
    \\
    \begin{subfigure}[h]{0.49\textwidth}
    \centering
        \includegraphics[width=0.49\textwidth]{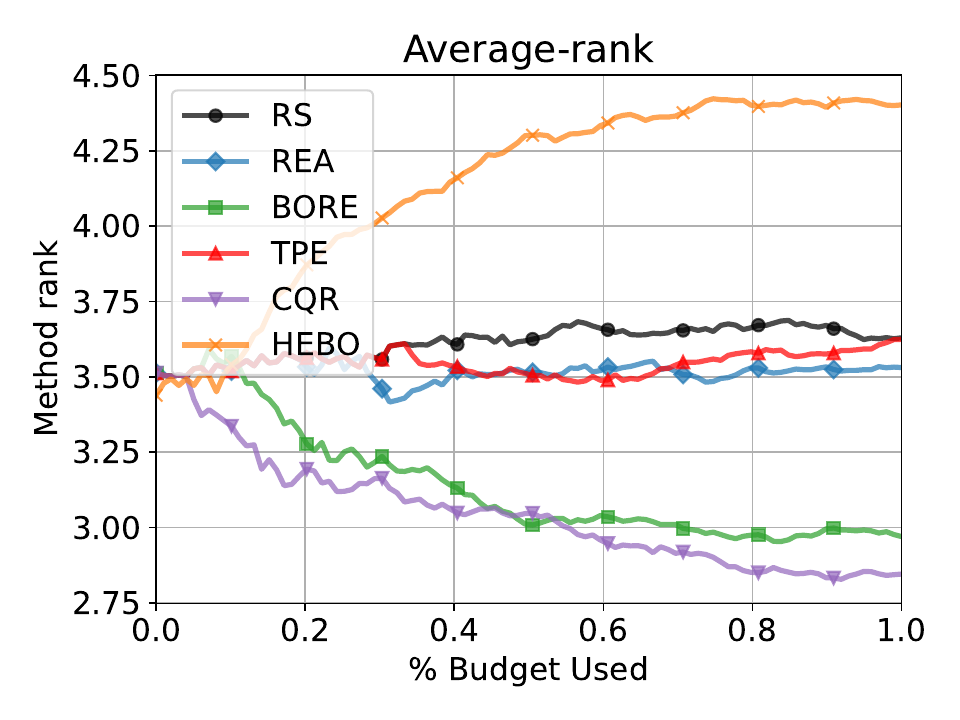}
        \includegraphics[width=0.49\textwidth]{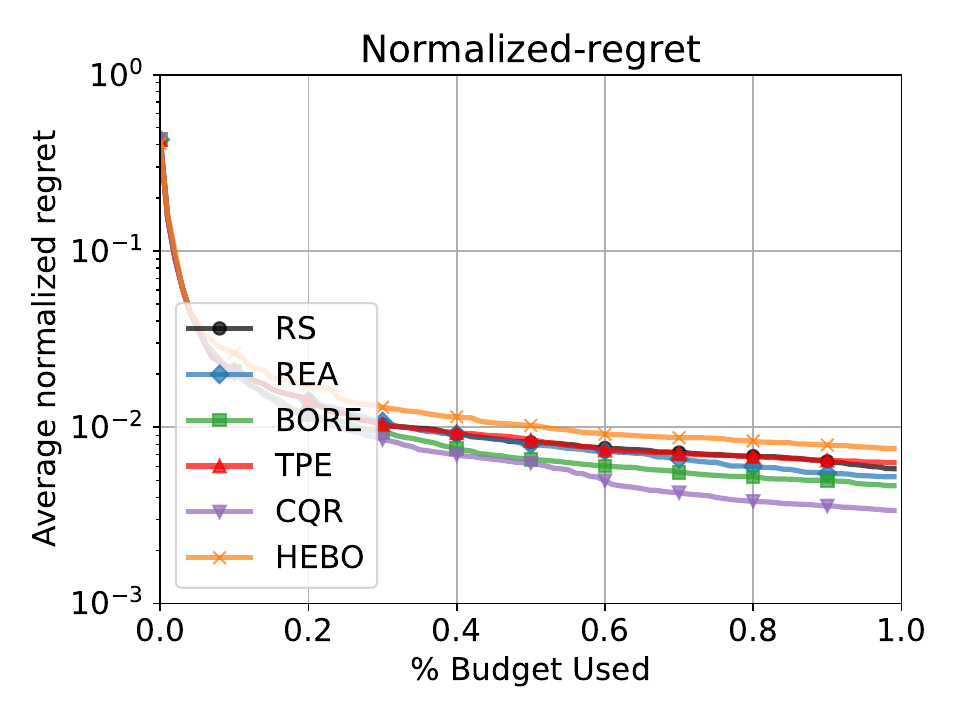}
        \caption{PD1}   
    \end{subfigure}
    \begin{subfigure}[h]{0.49\textwidth}
    \centering
        \includegraphics[width=0.49\textwidth]{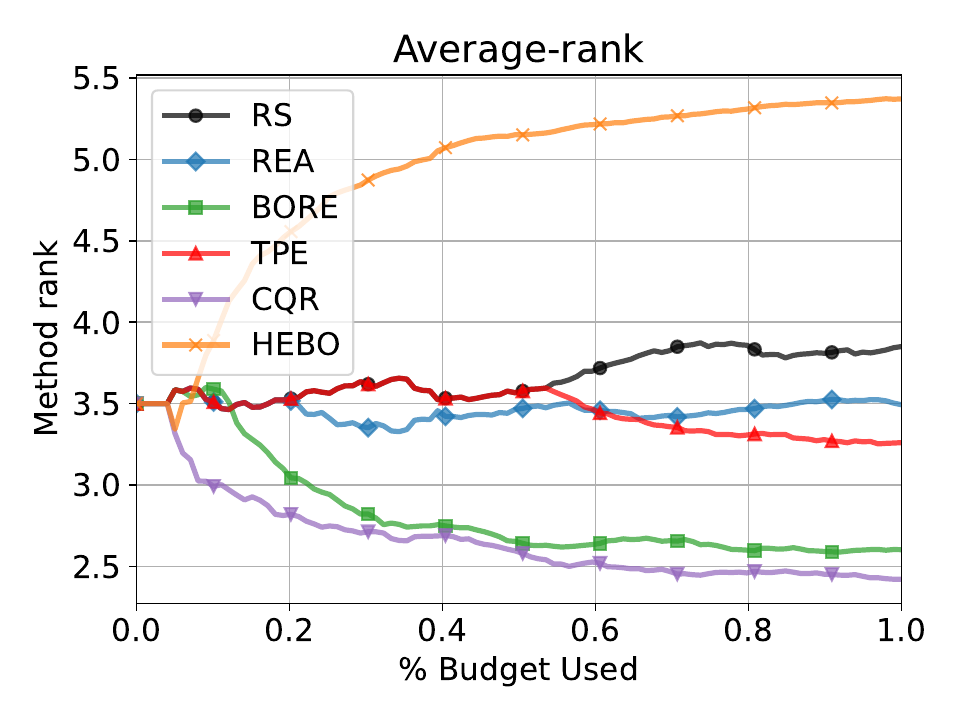}
        \includegraphics[width=0.49\textwidth]{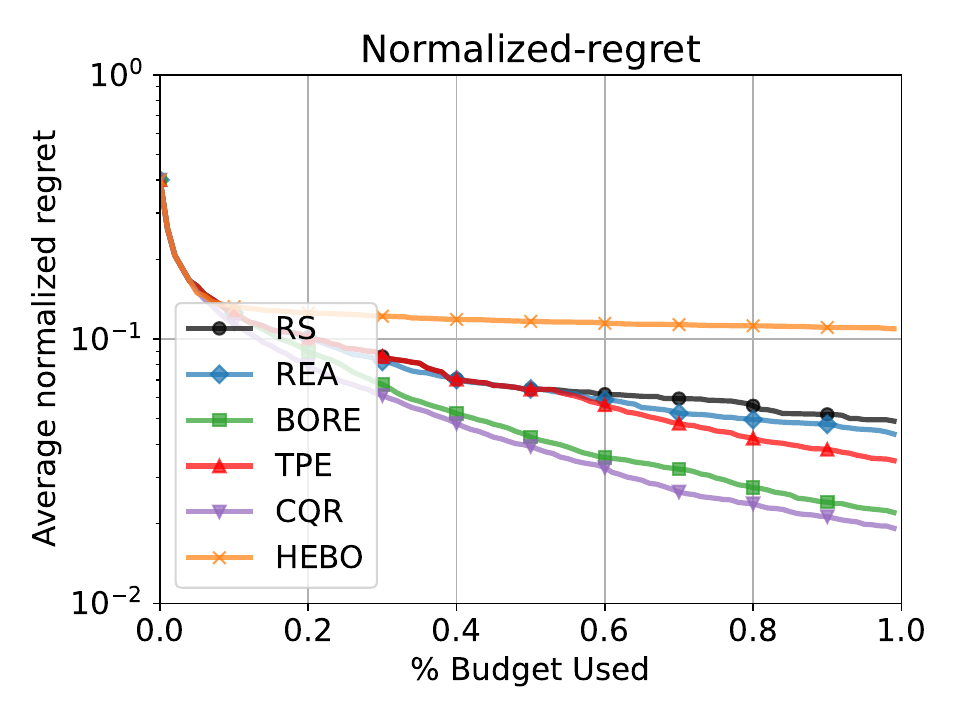}
        \caption{LC-Bench}
    \end{subfigure}
    \\
    \begin{subfigure}[h]{0.49\textwidth}
    \centering
        \includegraphics[width=0.49\textwidth]{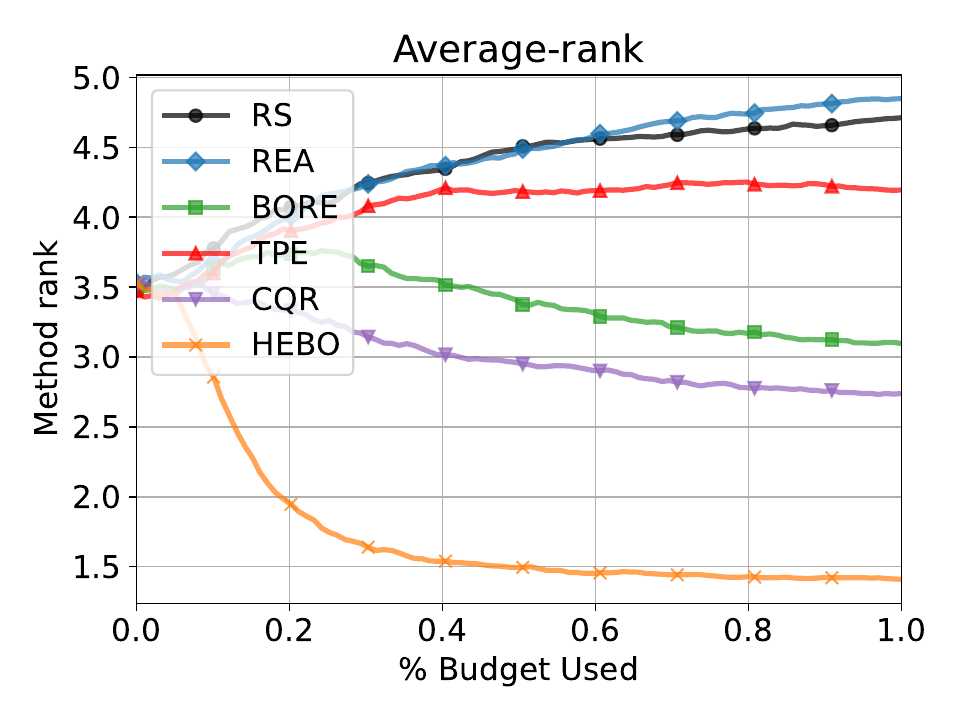}
        \includegraphics[width=0.49\textwidth]{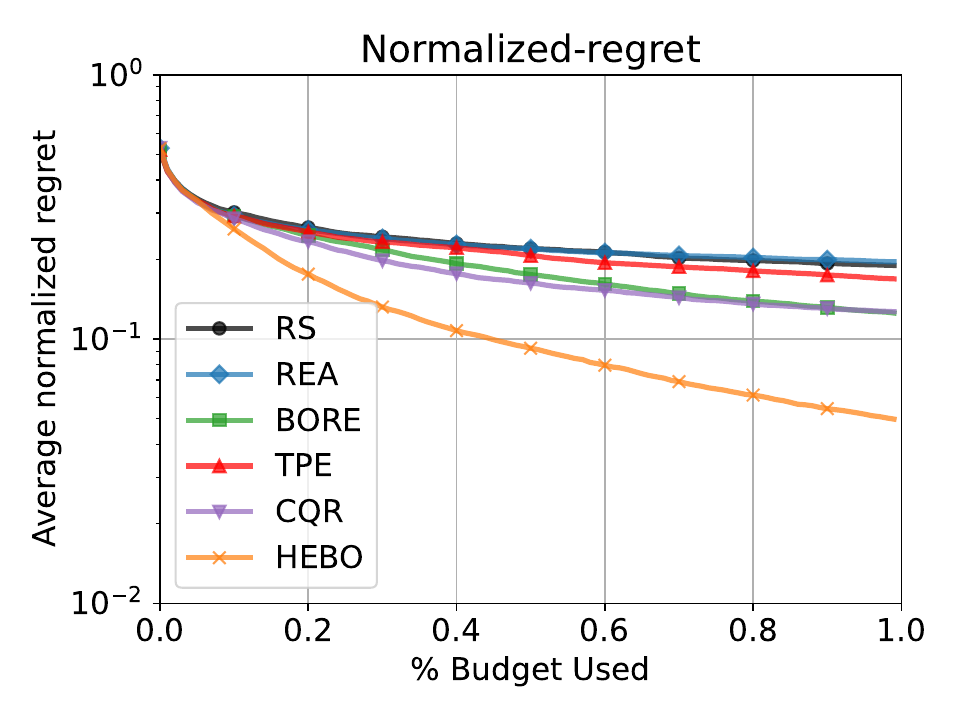}
        \caption{Global Optimization}   
    \end{subfigure}   
    \begin{subfigure}[h]{0.49\textwidth}
    \centering
        \includegraphics[width=0.49\textwidth]{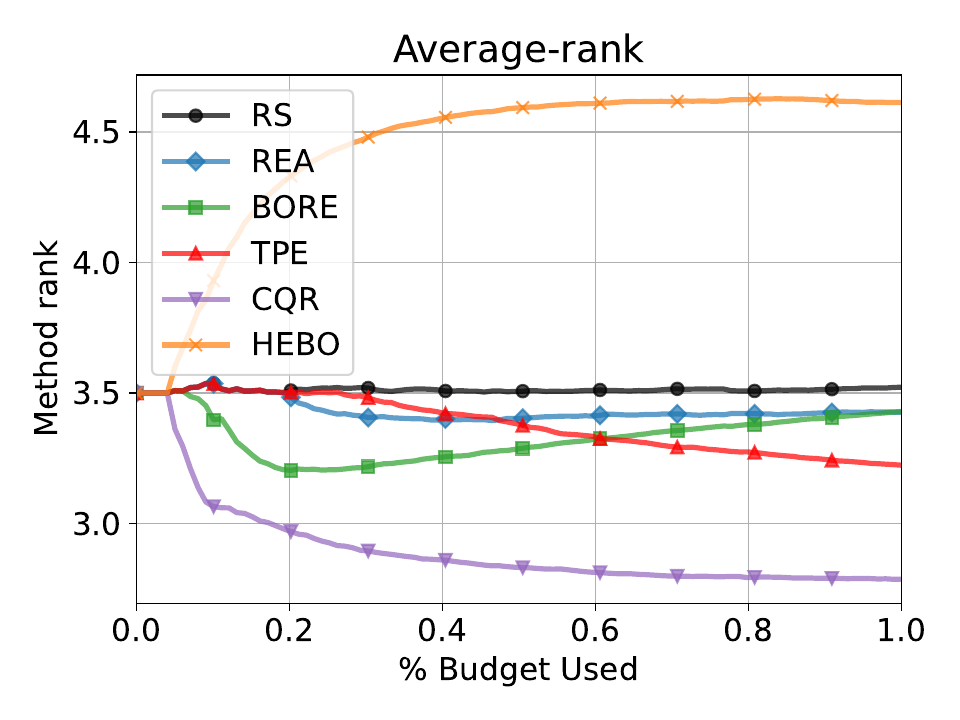}
        \includegraphics[width=0.49\textwidth]{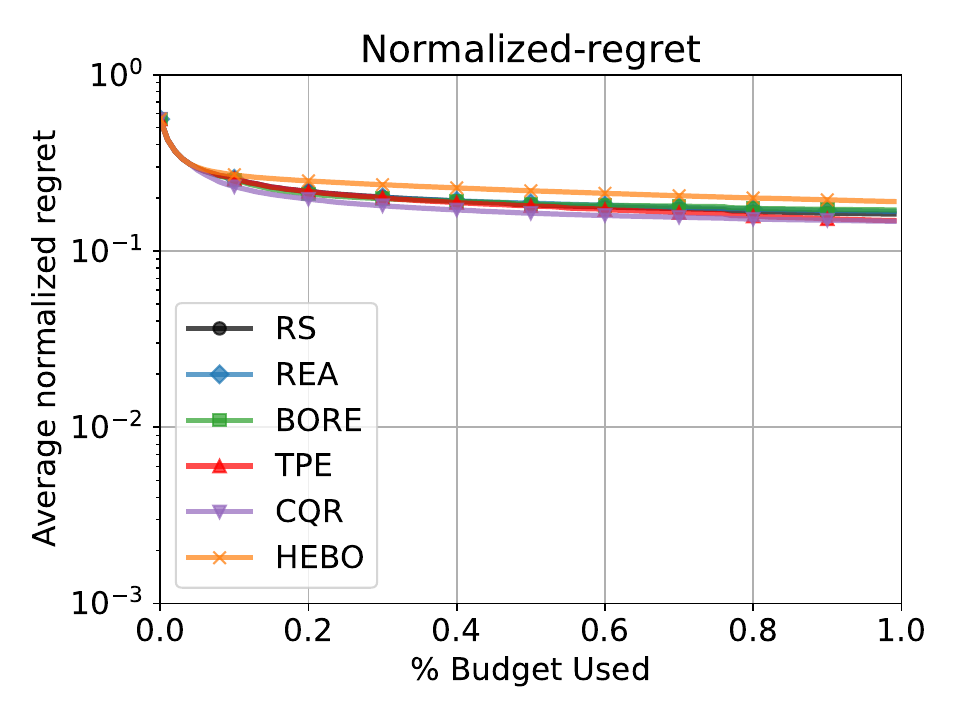}
        \caption{HPO-B}   
    \end{subfigure}   
    \\
    \begin{subfigure}[h]{0.49\textwidth}
    \centering
        \includegraphics[width=0.49\textwidth]{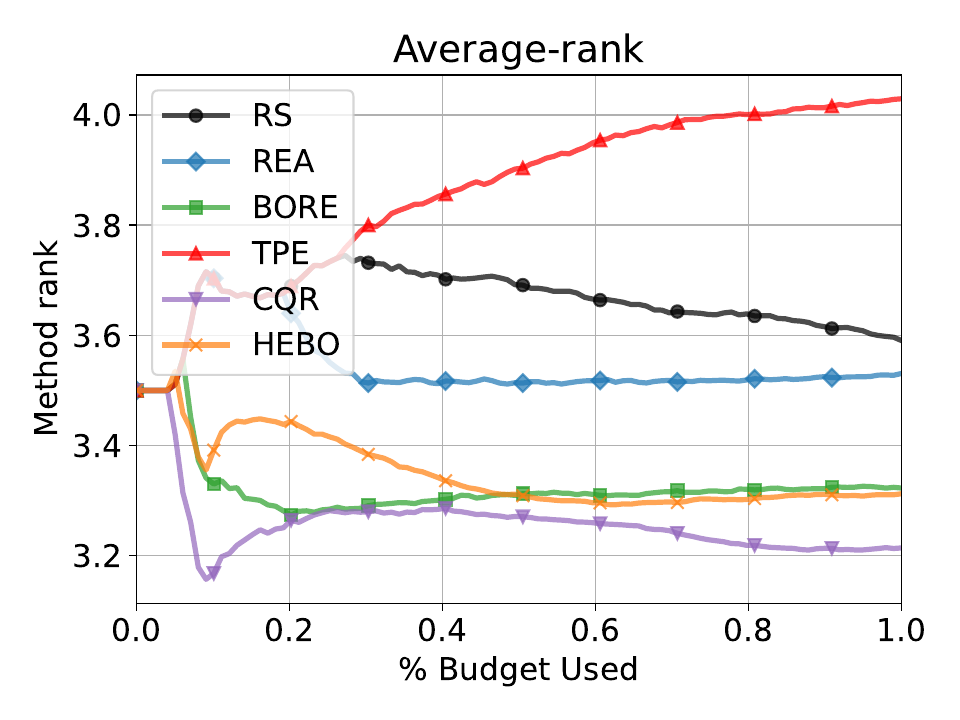}
        \includegraphics[width=0.49\textwidth]{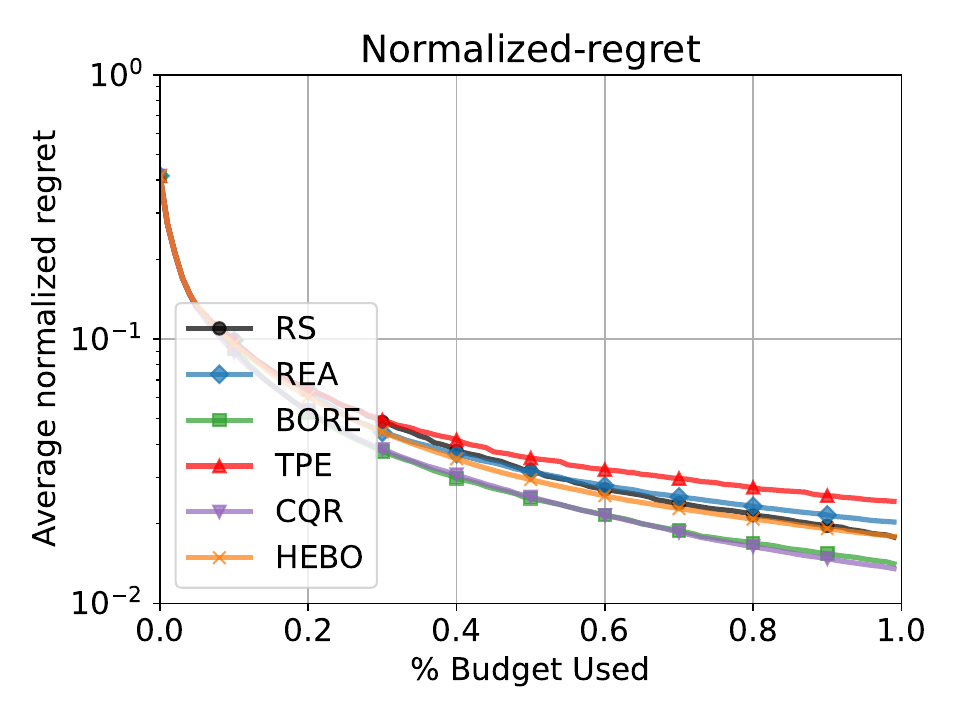}
        \caption{TabRepo}   
    \end{subfigure}   
    \caption{Ranks (left) and normalized regret (right) of all optimization methods averaged across all tasks for each benchmarking family.}
    \label{fig:comparison}
\end{figure*}

Figure~\ref{fig:comparison} compares all methods across different benchmark families. 
We report both the average rank and the normalized regret aggregated over all tasks within each benchmark family. 
For all methods, we use the default hyperparameters suggested by Syne Tune~\citep{salinas-automl22}. 
While some methods, such as CQR or HEBO, outperform others on specific benchmarks, no single method consistently dominates across all benchmark families.

\section{Distribution of Evaluations Across Objective-Value Deciles}\label{app:coverage}

We quantify the coverage of the original search space by our dataset by estimating the deciles of all loss values in the original space and then, for a given optimization trajectory, counting how many evaluations fall in each decile.
Table~\ref{tab:coverage} shows results on two representative benchmarks, FC-Net (protein) and NAS-Bench-201 (ImageNet16-120).
For example, on FC-Net, CQR concentrates 76.7\% of evaluations in the best decile with tight error bars across 30 seeds, while random search is flat at about 10\% per decile (uniform, as expected for random search).

\begin{table}[h]
    \centering
    \caption{Fraction of evaluations (\%) falling into each decile of the objective values of the search space, from Q1 (best) to Q10 (worst), as mean $\pm$ standard deviation over 30 seeds.}
    \label{tab:coverage}
    \scriptsize
    \begin{tabular}{lcccccc}
        \toprule
        Decile & RS & TPE & BORE & CQR & HEBO & REA \\
        \midrule
        \multicolumn{7}{l}{\textit{FC-Net (protein)}} \\
        Q1 (best)  & 9.2 $\pm$ 2.7  & 19.0 $\pm$ 3.8 & 63.8 $\pm$ 6.8 & 76.7 $\pm$ 9.5 & 83.8 $\pm$ 4.9 & 66.0 $\pm$ 6.5 \\
        Q2         & 10.1 $\pm$ 3.4 & 11.3 $\pm$ 3.8 & 10.0 $\pm$ 3.1 & 6.8 $\pm$ 3.4  & 4.4 $\pm$ 3.0  & 9.9 $\pm$ 4.1 \\
        Q3         & 9.7 $\pm$ 2.7  & 11.1 $\pm$ 3.1 & 5.6 $\pm$ 2.7  & 3.3 $\pm$ 2.4  & 2.1 $\pm$ 2.0  & 5.0 $\pm$ 2.2 \\
        Q4         & 10.1 $\pm$ 3.0 & 8.8 $\pm$ 2.8  & 4.5 $\pm$ 1.6  & 2.8 $\pm$ 2.0  & 2.2 $\pm$ 1.8  & 3.4 $\pm$ 2.1 \\
        Q5         & 10.1 $\pm$ 3.6 & 8.6 $\pm$ 3.5  & 2.3 $\pm$ 1.9  & 2.0 $\pm$ 1.6  & 1.6 $\pm$ 1.3  & 2.4 $\pm$ 1.5 \\
        Q6         & 10.2 $\pm$ 2.1 & 8.5 $\pm$ 2.7  & 1.9 $\pm$ 1.1  & 1.5 $\pm$ 1.5  & 1.2 $\pm$ 1.1  & 2.2 $\pm$ 1.6 \\
        Q7         & 10.2 $\pm$ 3.4 & 8.6 $\pm$ 3.2  & 1.9 $\pm$ 1.5  & 1.7 $\pm$ 1.7  & 1.0 $\pm$ 1.0  & 2.7 $\pm$ 1.9 \\
        Q8         & 9.0 $\pm$ 2.5  & 7.1 $\pm$ 1.9  & 2.9 $\pm$ 1.9  & 1.2 $\pm$ 1.0  & 1.2 $\pm$ 1.1  & 1.8 $\pm$ 1.2 \\
        Q9         & 10.5 $\pm$ 2.9 & 8.4 $\pm$ 2.4  & 4.3 $\pm$ 2.2  & 2.6 $\pm$ 2.3  & 1.4 $\pm$ 1.1  & 4.1 $\pm$ 1.8 \\
        Q10 (worst)& 10.9 $\pm$ 3.0 & 8.5 $\pm$ 2.4  & 2.7 $\pm$ 1.5  & 1.5 $\pm$ 1.5  & 1.1 $\pm$ 1.0  & 2.3 $\pm$ 1.5 \\
        \midrule
        \multicolumn{7}{l}{\textit{NAS-Bench-201 (ImageNet16-120)}} \\
        Q1 (best)  & 10.0 $\pm$ 2.3 & 43.2 $\pm$ 6.4 & 72.0 $\pm$ 5.3 & 80.2 $\pm$ 4.0 & 48.7 $\pm$ 4.8 & 65.7 $\pm$ 5.8 \\
        Q2         & 10.0 $\pm$ 2.3 & 10.7 $\pm$ 3.1 & 10.3 $\pm$ 3.5 & 7.8 $\pm$ 2.5  & 13.4 $\pm$ 3.4 & 13.6 $\pm$ 3.0 \\
        Q3         & 9.9 $\pm$ 2.3  & 7.1 $\pm$ 2.3  & 4.8 $\pm$ 1.7  & 3.1 $\pm$ 1.4  & 8.6 $\pm$ 2.1  & 5.1 $\pm$ 2.9 \\
        Q4         & 9.8 $\pm$ 2.2  & 6.3 $\pm$ 2.1  & 2.6 $\pm$ 1.4  & 2.2 $\pm$ 1.1  & 6.3 $\pm$ 2.7  & 2.9 $\pm$ 1.4 \\
        Q5         & 10.2 $\pm$ 2.1 & 6.5 $\pm$ 1.8  & 2.4 $\pm$ 0.9  & 1.6 $\pm$ 1.0  & 5.6 $\pm$ 3.0  & 4.0 $\pm$ 1.5 \\
        Q6         & 9.7 $\pm$ 2.1  & 5.7 $\pm$ 1.8  & 2.1 $\pm$ 1.0  & 1.7 $\pm$ 0.9  & 4.3 $\pm$ 2.0  & 2.5 $\pm$ 1.3 \\
        Q7         & 10.6 $\pm$ 2.3 & 5.6 $\pm$ 1.9  & 1.8 $\pm$ 1.0  & 1.0 $\pm$ 0.7  & 3.9 $\pm$ 1.6  & 1.9 $\pm$ 1.3 \\
        Q8         & 9.5 $\pm$ 1.5  & 5.1 $\pm$ 1.3  & 1.5 $\pm$ 0.8  & 0.9 $\pm$ 0.7  & 3.9 $\pm$ 1.2  & 2.0 $\pm$ 1.4 \\
        Q9         & 10.2 $\pm$ 1.9 & 5.1 $\pm$ 1.7  & 1.0 $\pm$ 0.7  & 0.6 $\pm$ 0.4  & 3.1 $\pm$ 2.0  & 1.3 $\pm$ 0.8 \\
        Q10 (worst)& 10.1 $\pm$ 2.3 & 4.9 $\pm$ 1.2  & 1.4 $\pm$ 0.7  & 0.9 $\pm$ 0.4  & 2.1 $\pm$ 1.4  & 1.0 $\pm$ 0.7 \\
        \bottomrule
    \end{tabular}
\end{table}

\section{Grid Search}\label{app:model_grid}

\begin{figure}[h]
    \centering
    \begin{subfigure}[h]{0.99\textwidth}
        \centering
        \includegraphics[width=0.19\textwidth]{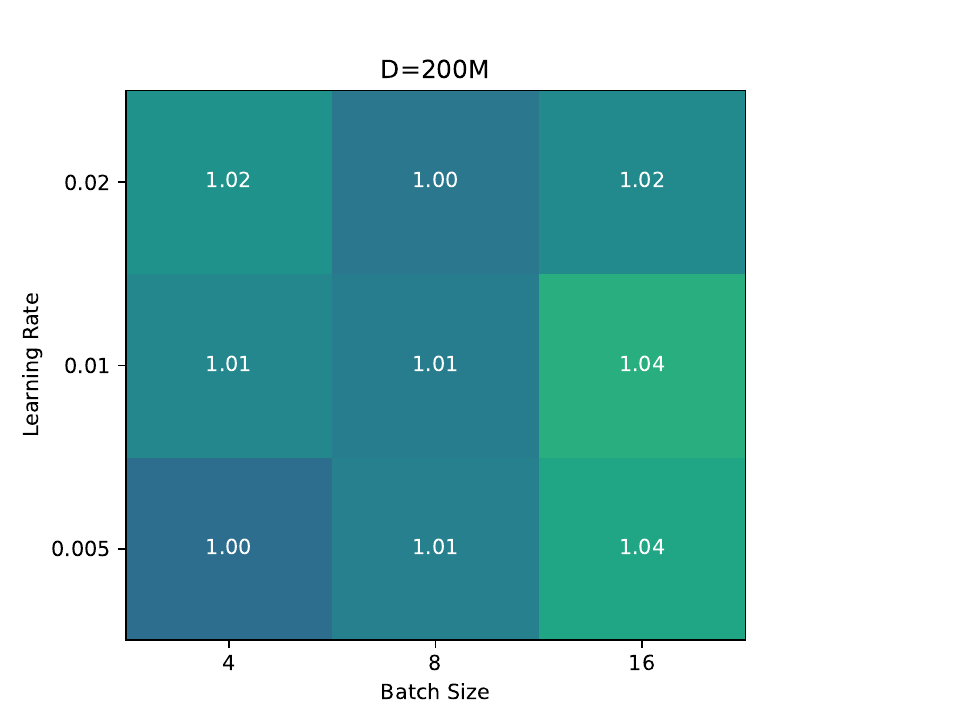}
        \includegraphics[width=0.19\textwidth]{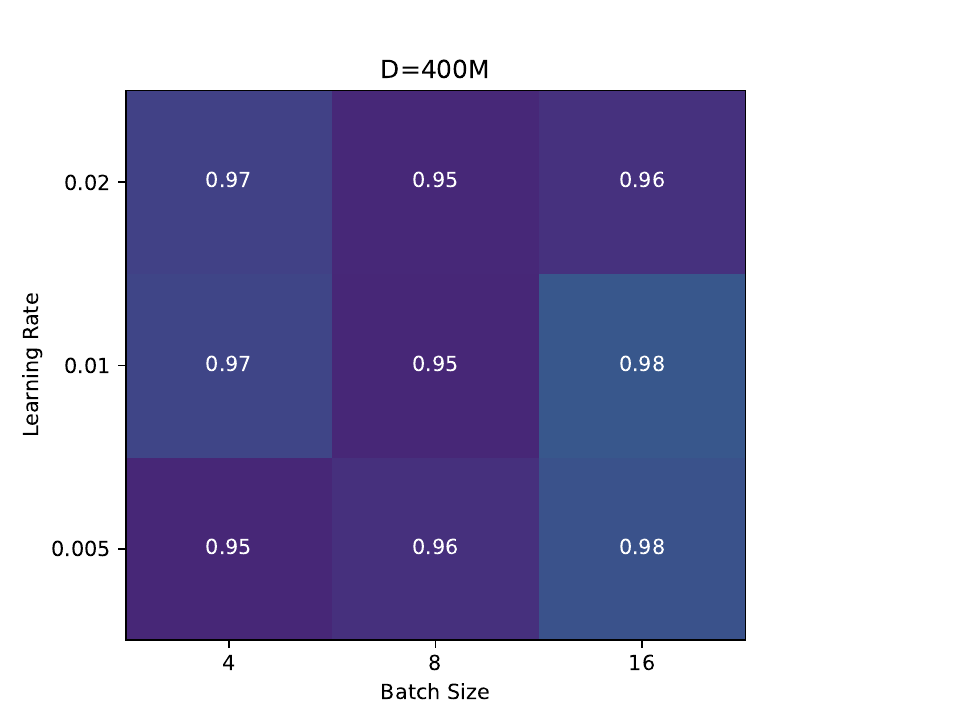}        
        \includegraphics[width=0.19\textwidth]{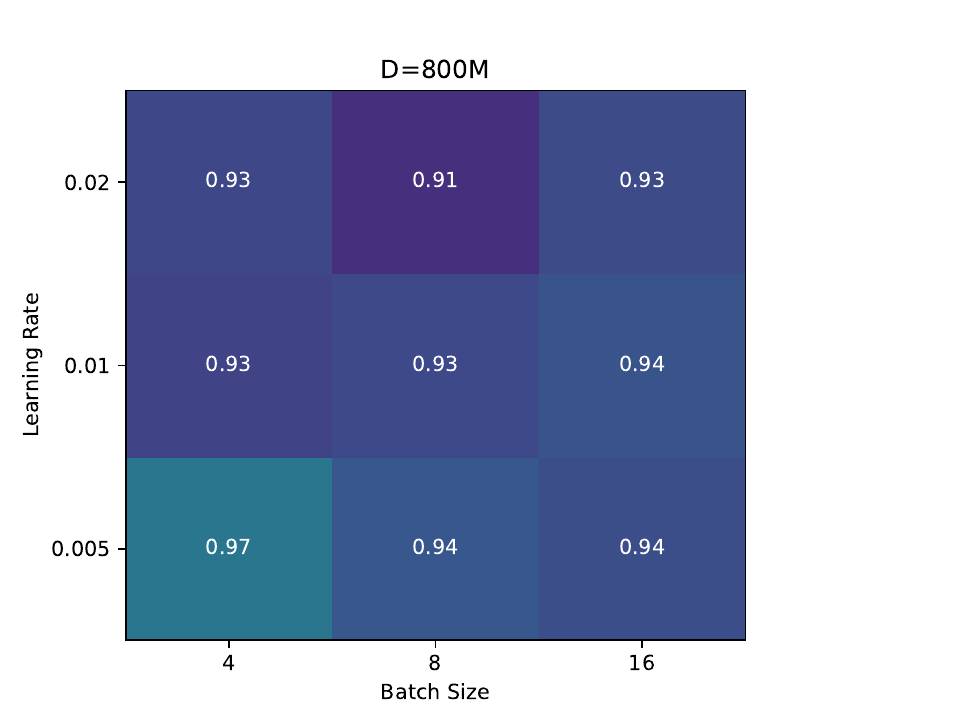}        
        \includegraphics[width=0.19\textwidth]{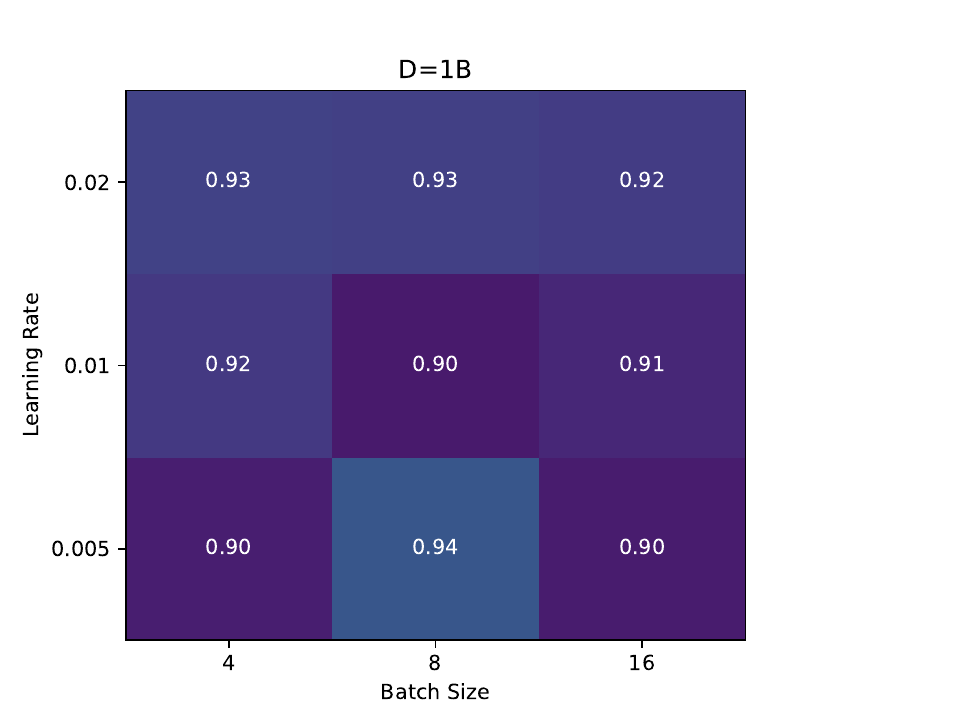}        
        \includegraphics[width=0.19\textwidth]{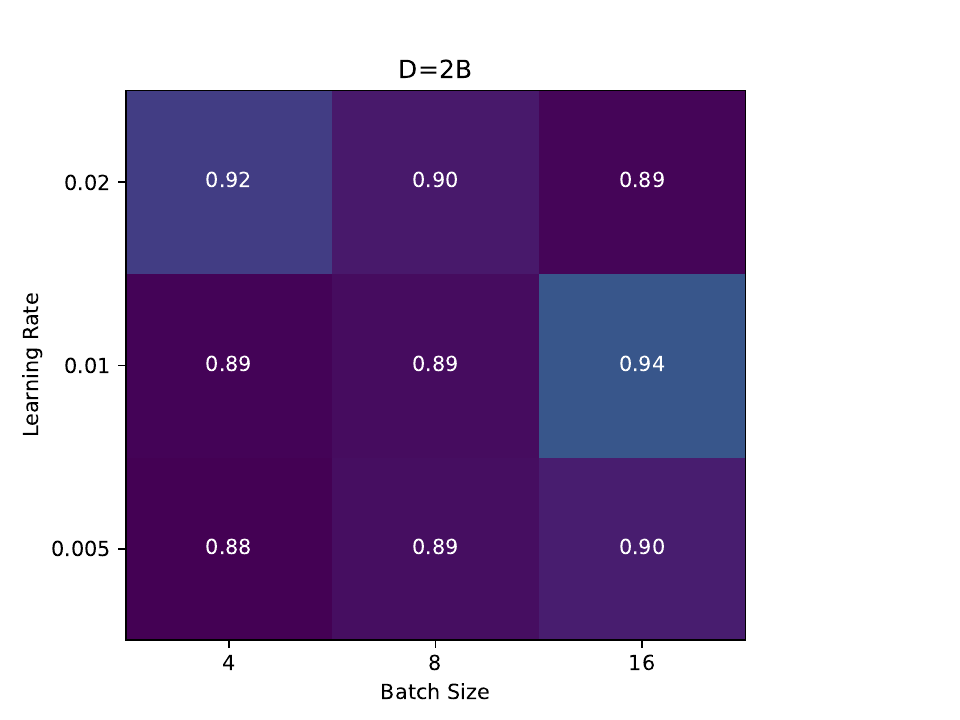}                
        \caption{2M parameters}   
    \end{subfigure}   
    \begin{subfigure}[h]{0.99\textwidth}
        \centering
        \includegraphics[width=0.19\textwidth]{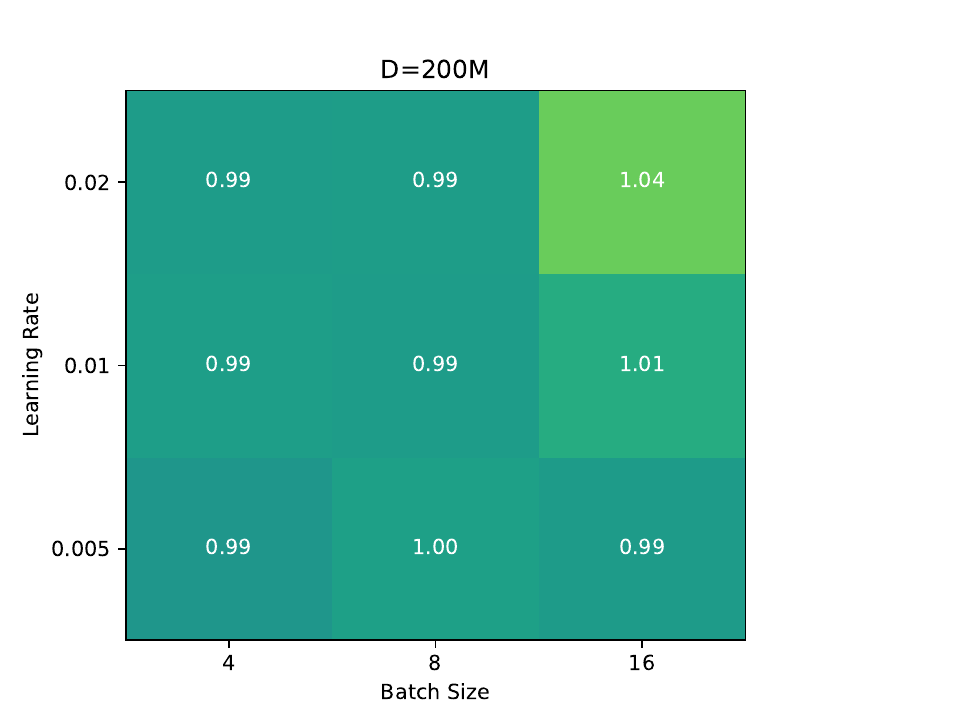}               
        \includegraphics[width=0.19\textwidth]{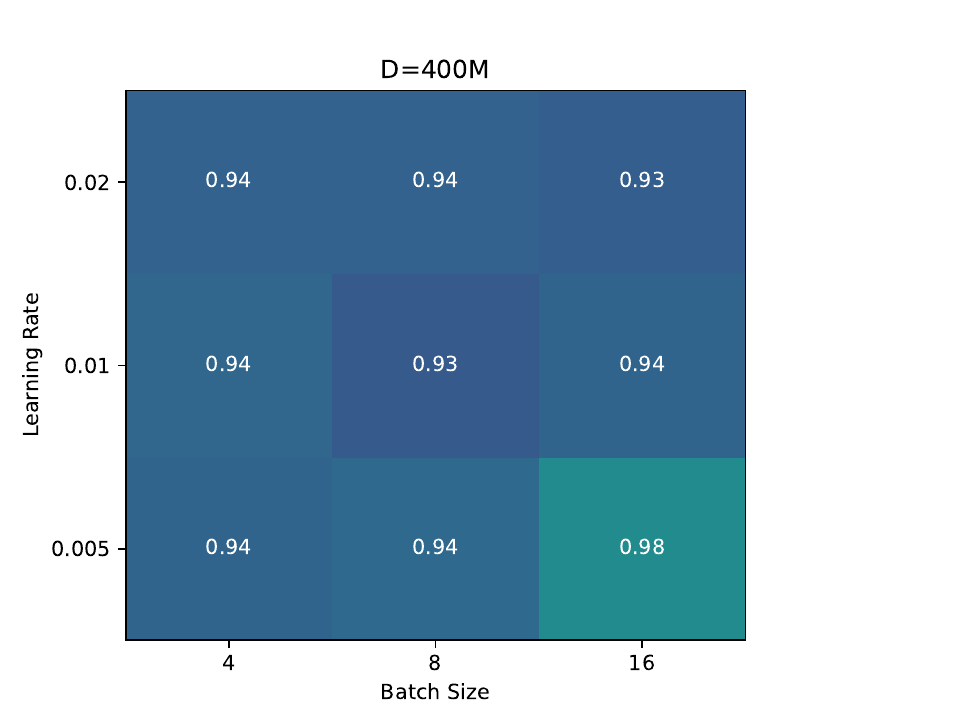}        
        \includegraphics[width=0.19\textwidth]{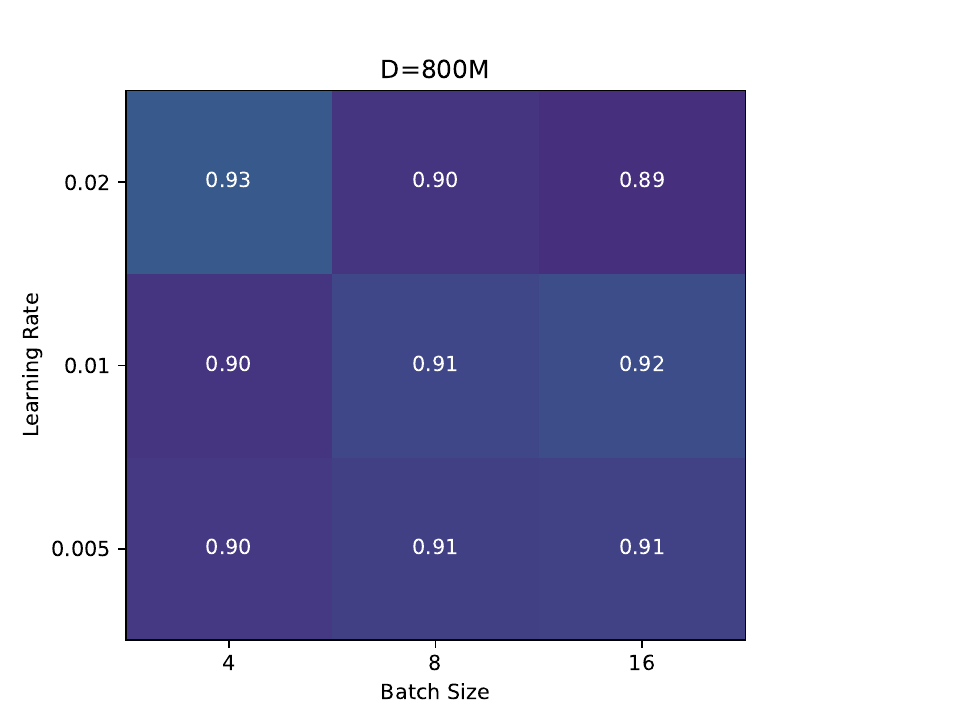}       
        \includegraphics[width=0.19\textwidth]{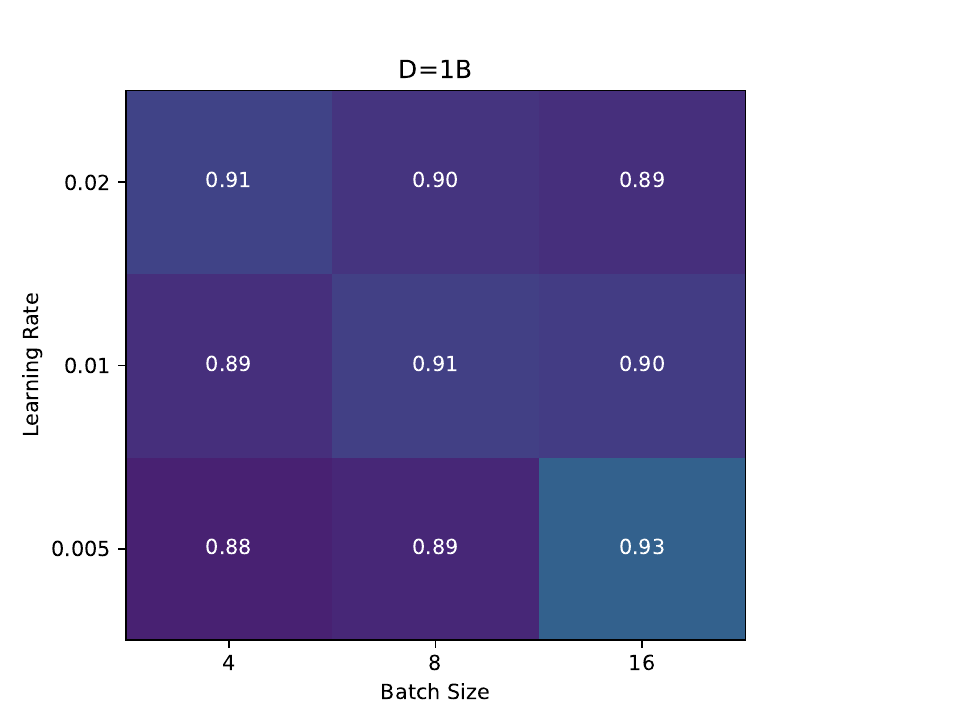}       
        \includegraphics[width=0.19\textwidth]{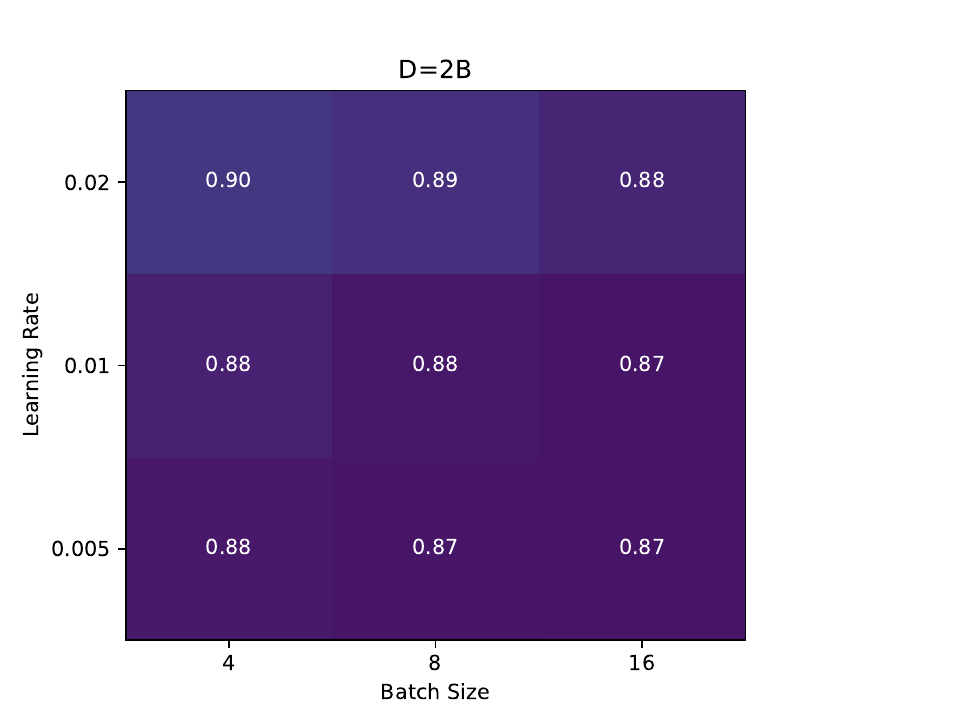}        

        \caption{5M parameters}   
    \end{subfigure}   
    \begin{subfigure}[h]{0.99\textwidth}
        \includegraphics[width=0.19\textwidth]{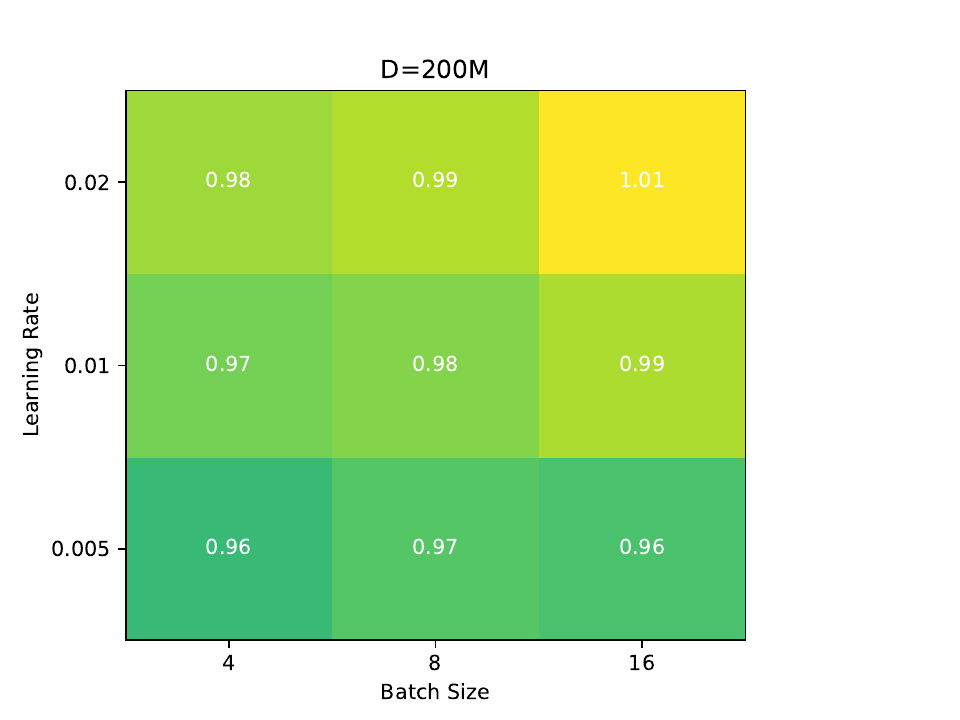}
        \includegraphics[width=0.19\textwidth]{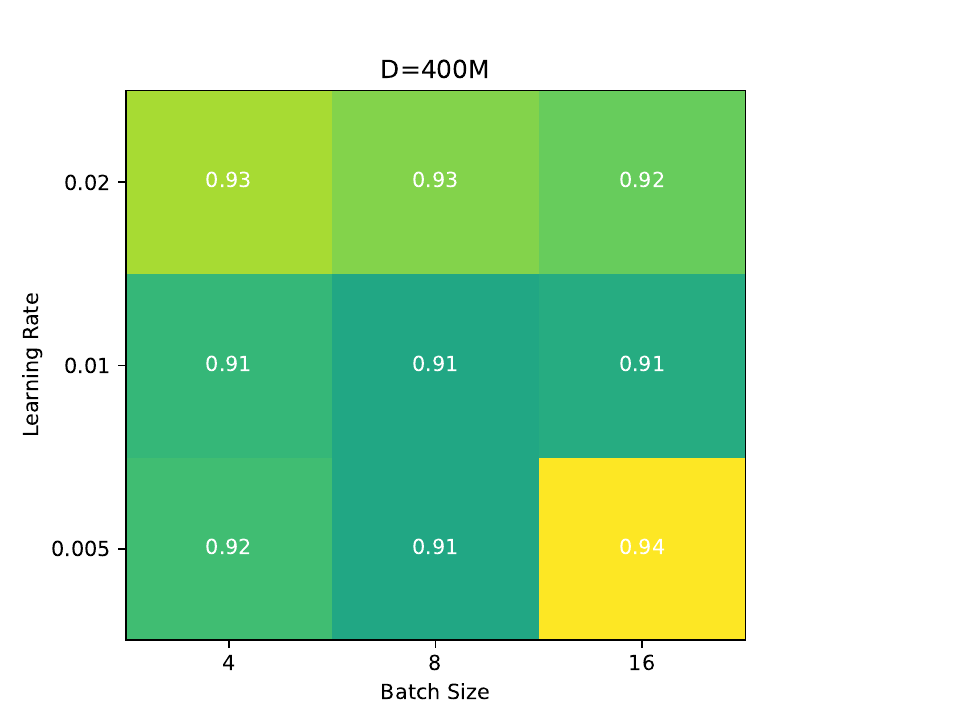}
        \includegraphics[width=0.19\textwidth]{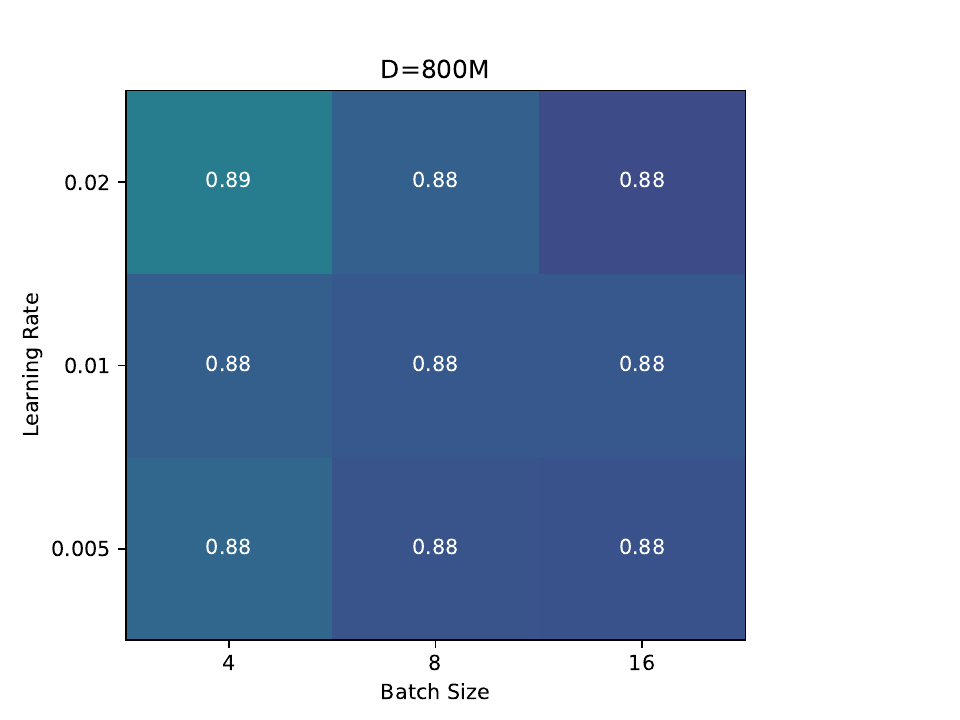}
        \includegraphics[width=0.19\textwidth]{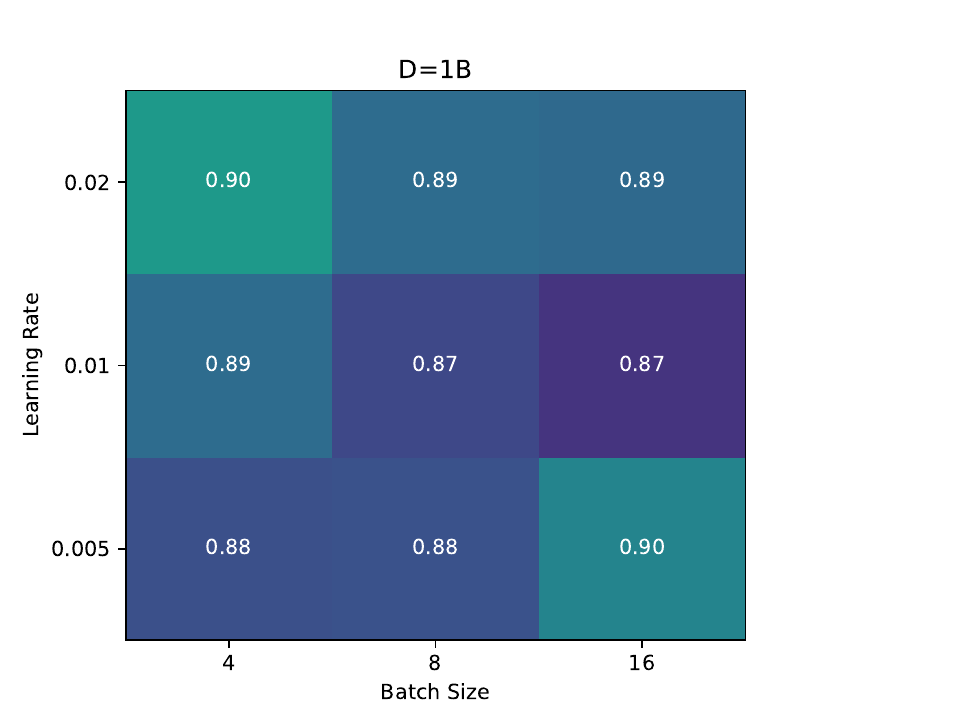}
        \includegraphics[width=0.19\textwidth]{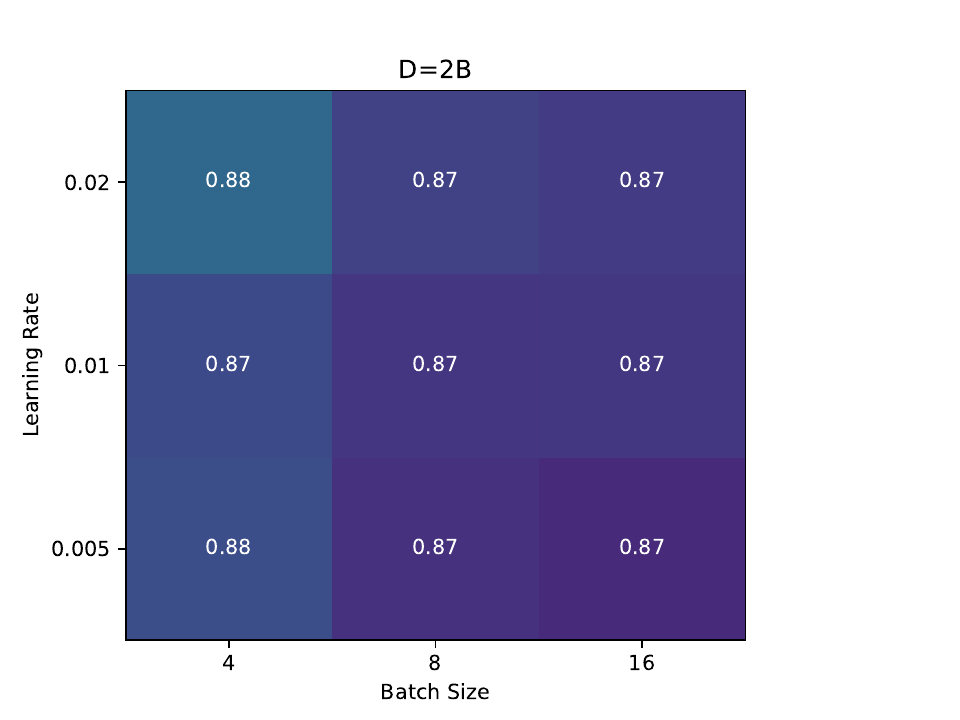}

        \caption{13M parameters}   
    \end{subfigure}      
     \begin{subfigure}[h]{0.99\textwidth}
        \includegraphics[width=0.19\textwidth]{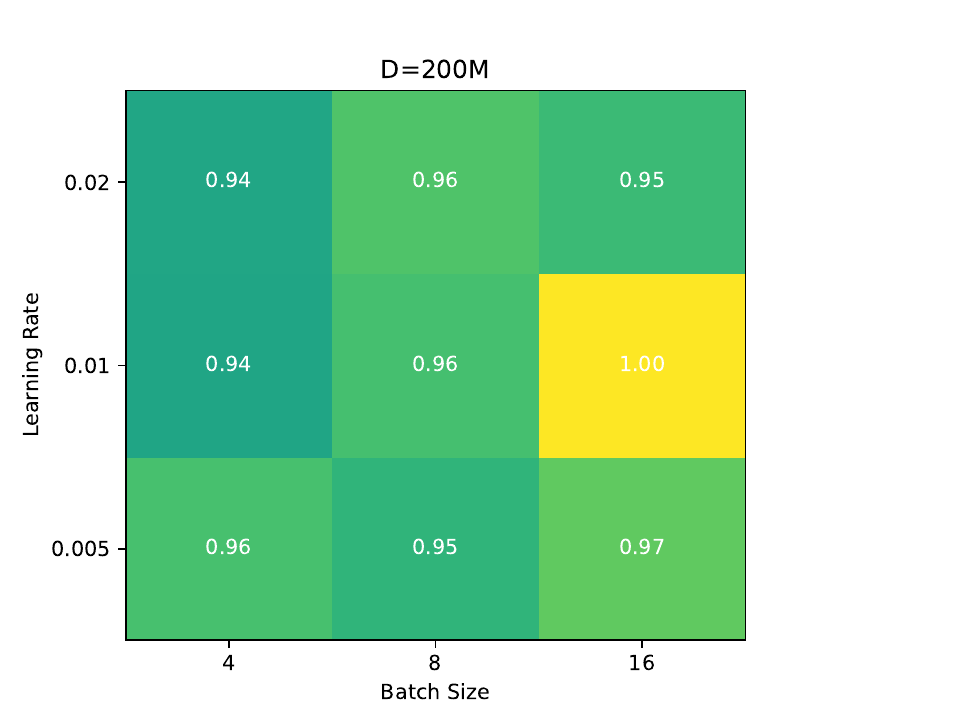}
        \includegraphics[width=0.19\textwidth]{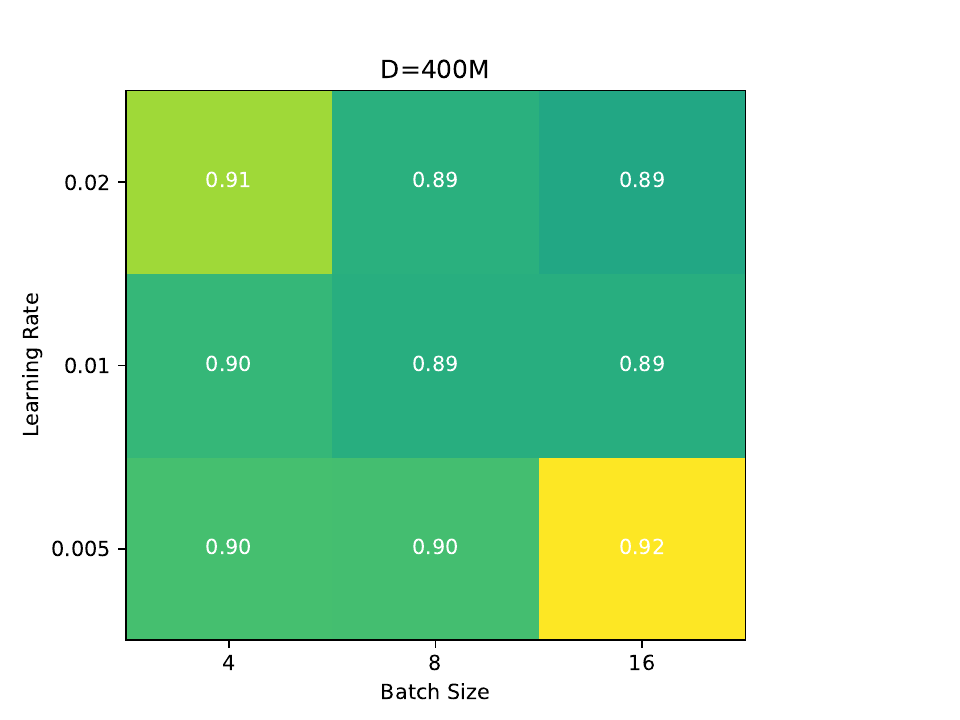}
        \includegraphics[width=0.19\textwidth]{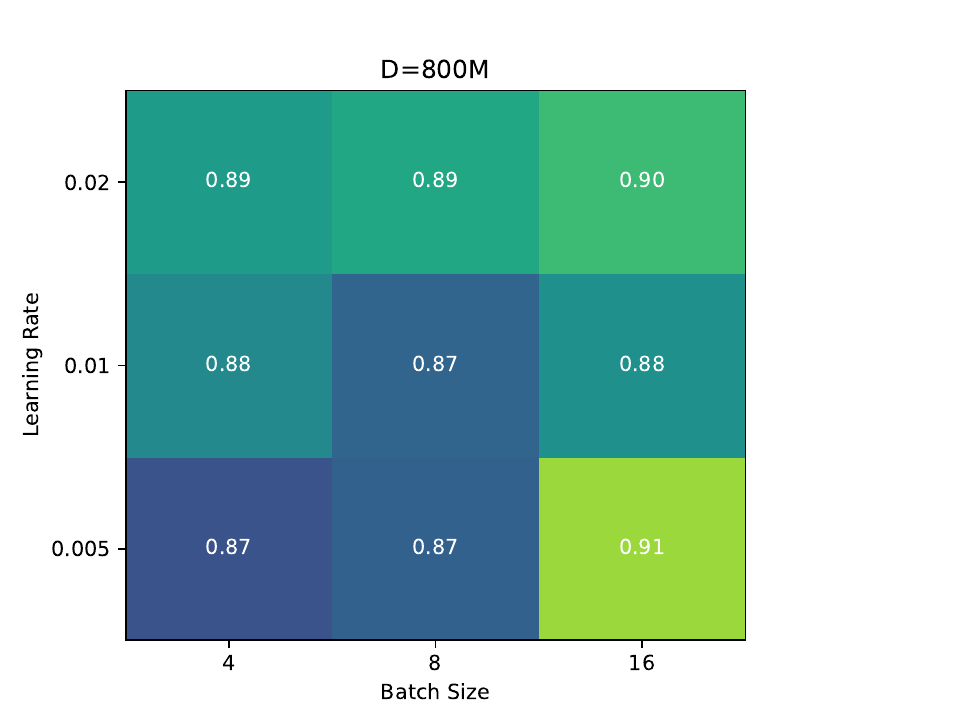}
        \includegraphics[width=0.19\textwidth]{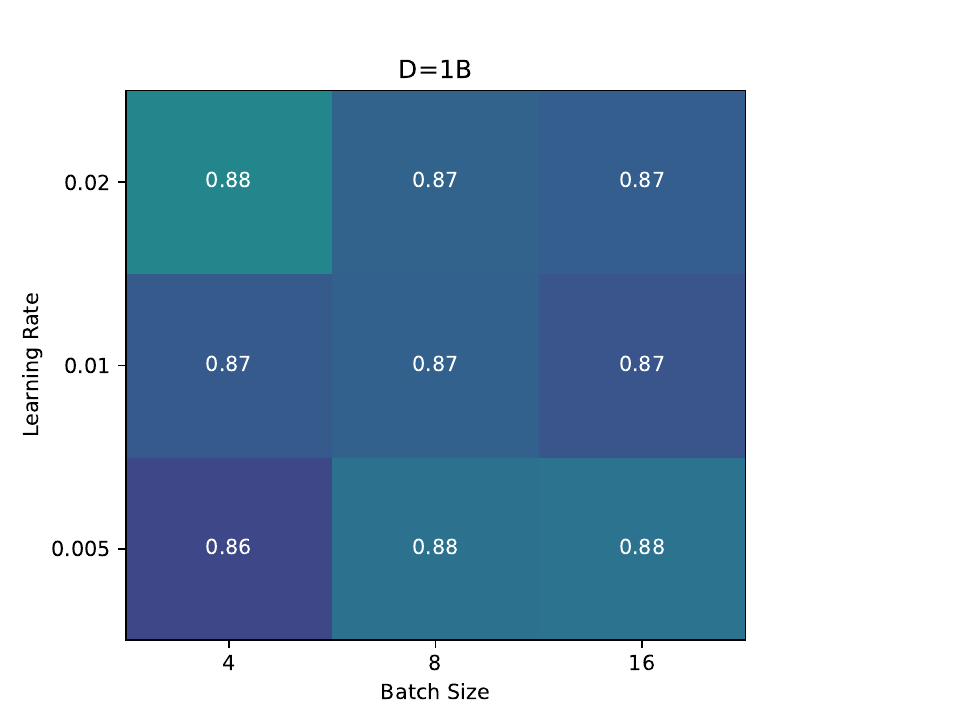}
        \includegraphics[width=0.19\textwidth]{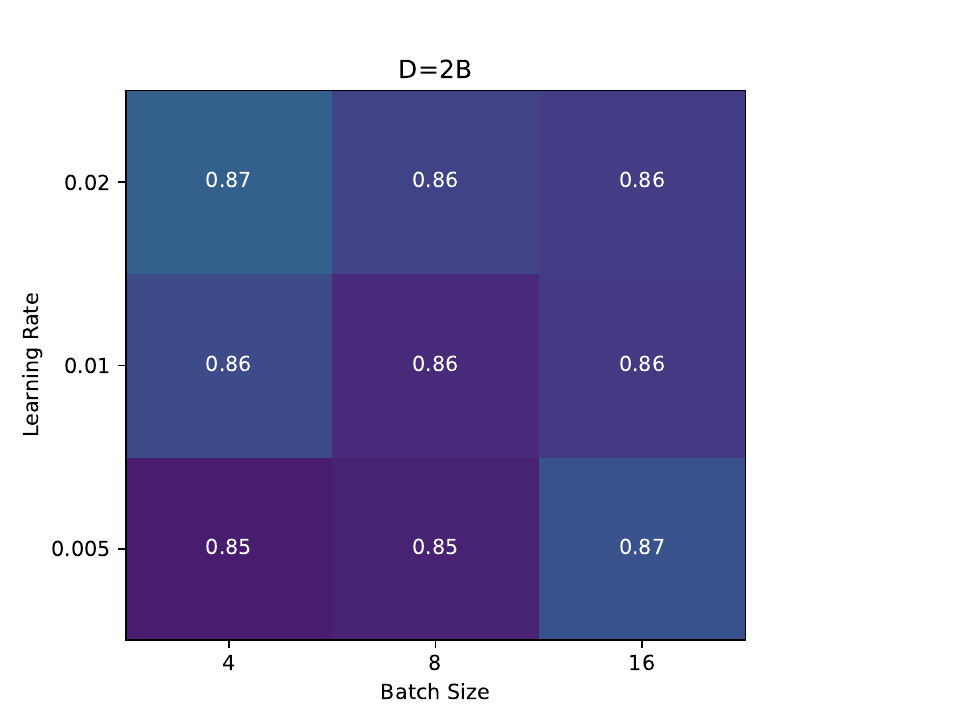}

        \caption{30M parameters}   
    \end{subfigure}         
    \begin{subfigure}[h]{0.99\textwidth}
        \includegraphics[width=0.19\textwidth]{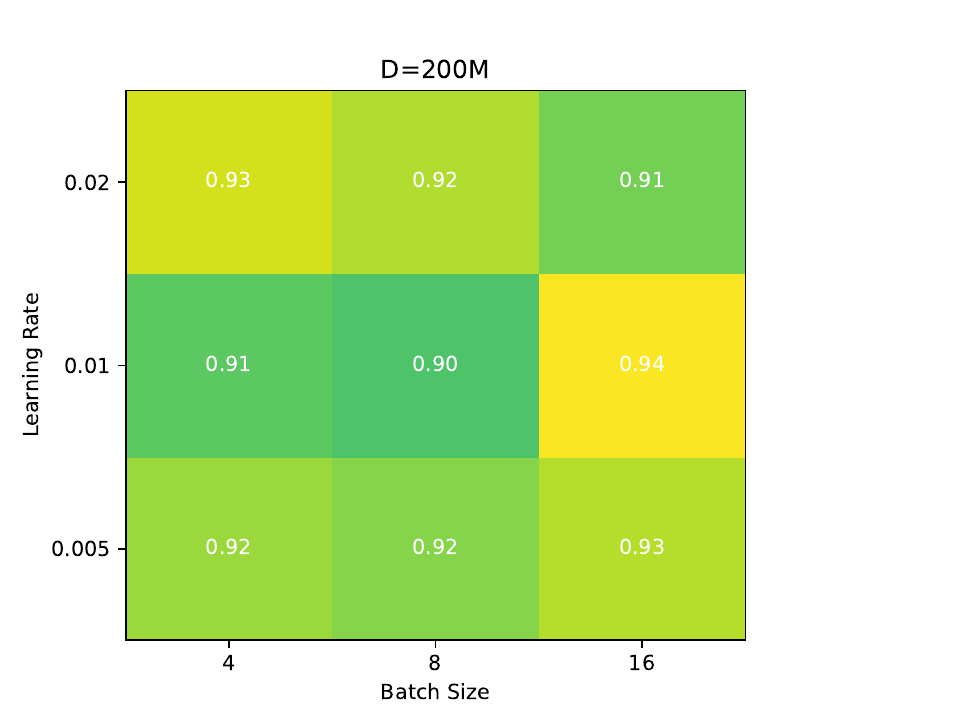}
        \includegraphics[width=0.19\textwidth]{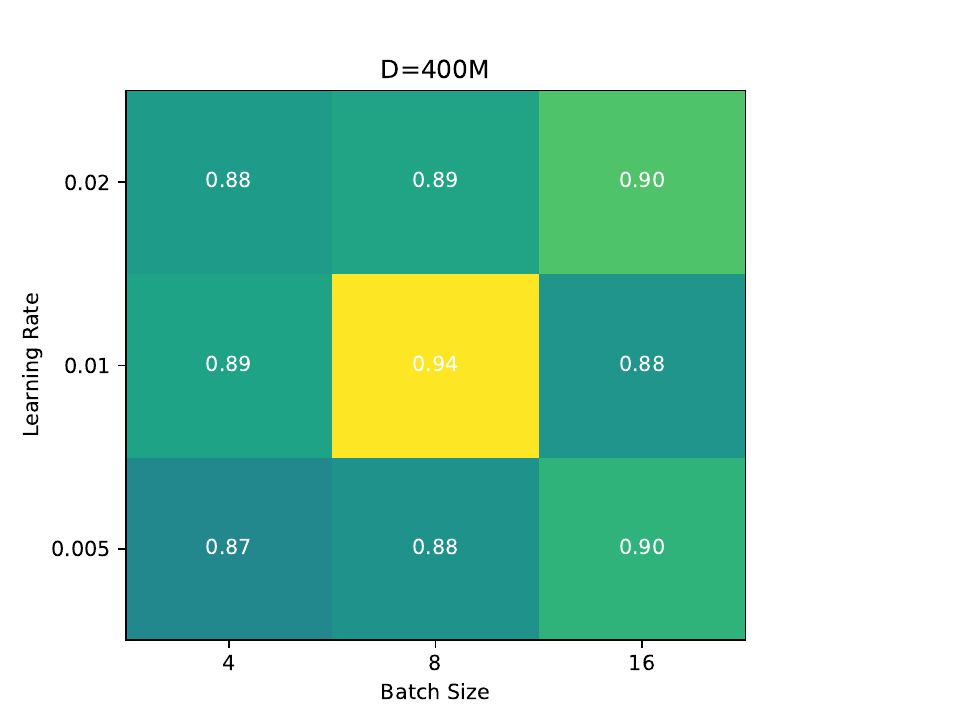}
        \includegraphics[width=0.19\textwidth]{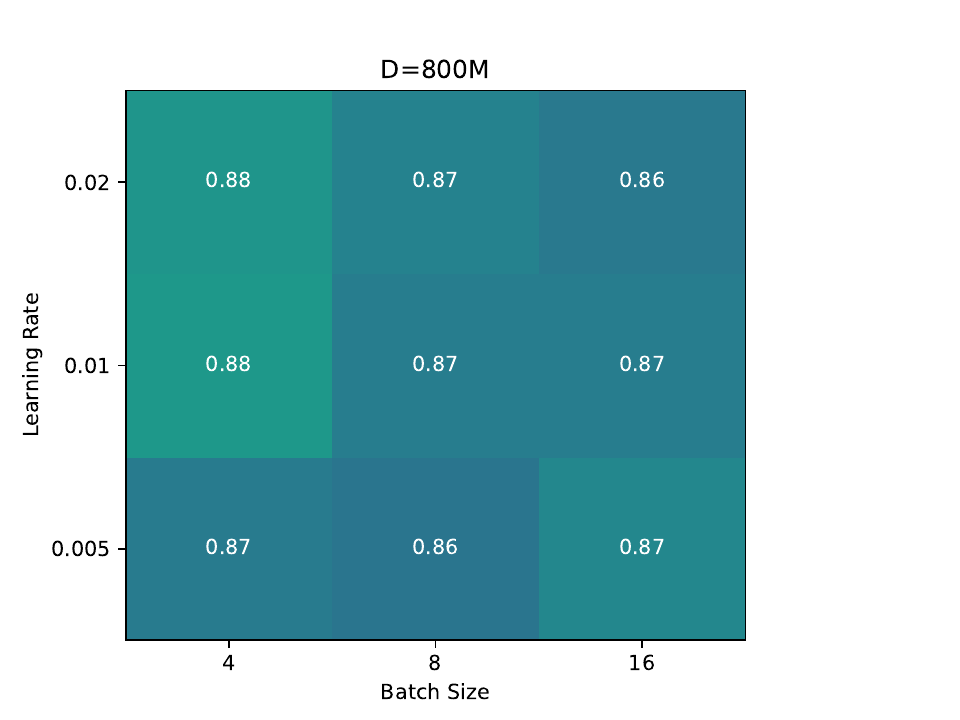}
        \includegraphics[width=0.19\textwidth]{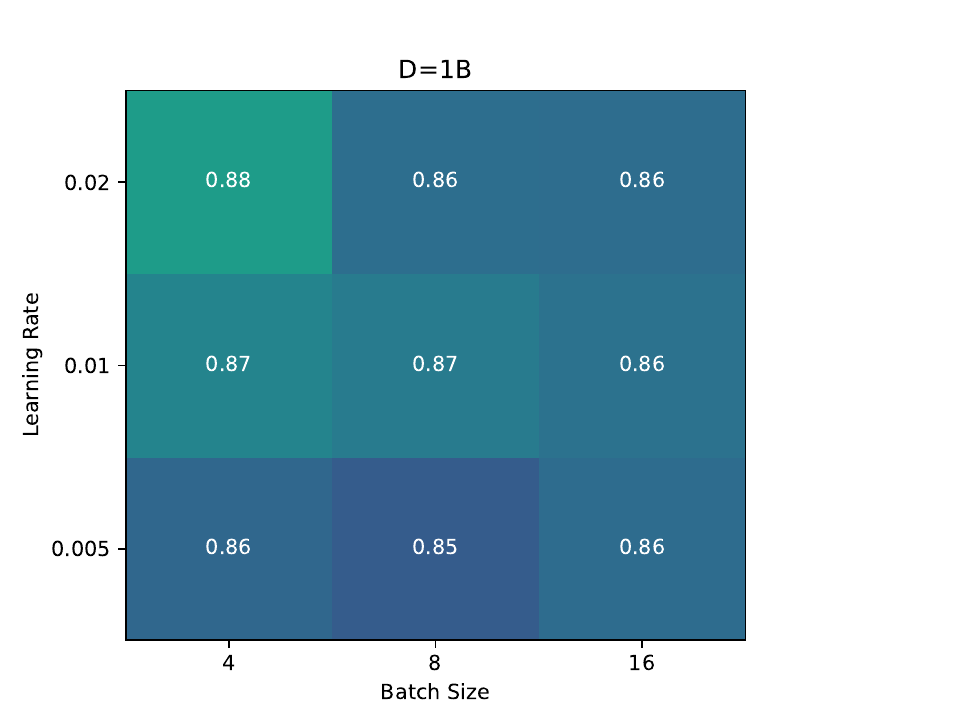}
        \includegraphics[width=0.19\textwidth]{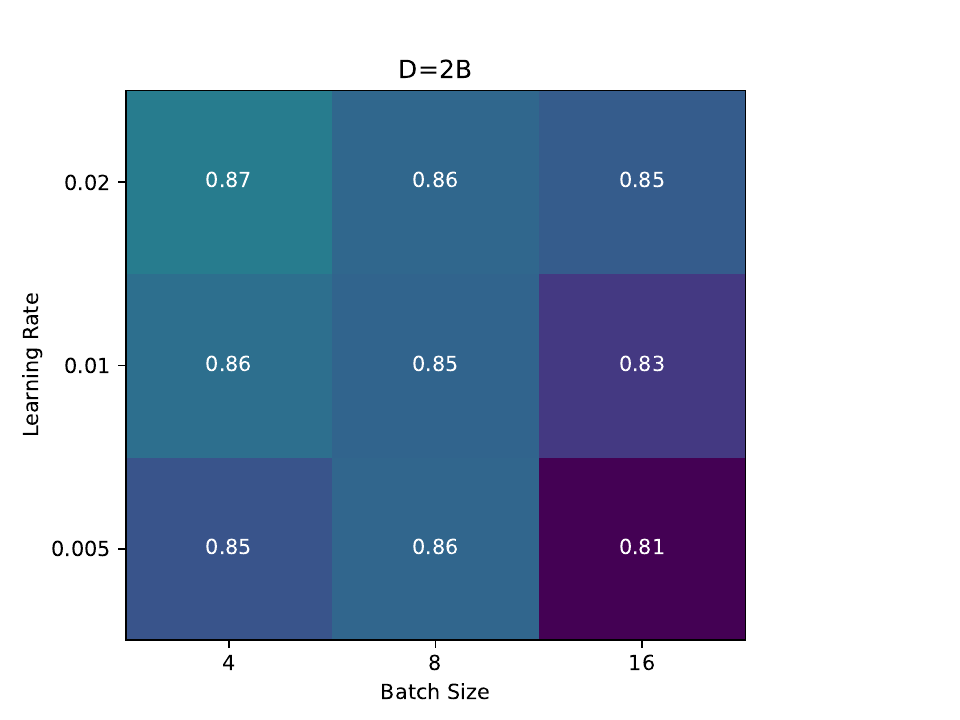}

        \caption{80M parameters}   
    \end{subfigure}      

    \caption{Hyperparameter grid of each model and token budget. Color indicates the validation loss of the final checkpoint.}
    \label{fig:hp_grid}
\end{figure}

\begin{table*}[t]
\centering
\caption{Grid of model architectures used for scaling experiments.}
\label{tab:model_architectures}
\begin{tabular}{lcccccc}
\toprule
\textbf{Model Size} 
& \textbf{\# Layers} 
& \textbf{\# Heads} 
& \textbf{\# Groups} 
& \textbf{Model Dim} 
& \textbf{Head Dim} 
& \textbf{FFN Size} 
\\
\midrule
2M   & 7  & 8  & 4 & 128 & 64  & 384  \\
5M   & 14 & 8  & 4 & 128 & 64  & 384  \\
13M  & 14 & 16 & 8 & 128 & 128 & 384  \\
30M  & 14 & 16 & 8 & 256 & 128 & 768  \\
80M  & 14 & 16 & 8 & 512 & 128 & 1536 \\
\bottomrule
\end{tabular}
\end{table*}

As discussed in Section~\ref{sub:compute}, we perform a grid search to identify the optimal learning rate and batch size for each combination of parameter count and token budget. 
Table~\ref{tab:model_architectures} shows the architecture configuration of each model size.
We consider peak learning rates in $\{5e^{-3}, 1e^{-2}, 2e^{-2}\}$ and global batch sizes in $\{4, 8, 16\}$. 
As described in the main paper, 10\% of the total token budget is used for learning-rate warm-up, followed by a decay phase using cosine annealing~\citep{loshchilov-iclr17}. 
All models are trained using the same random seed to control for noise.
Figure~\ref{fig:hp_grid} presents the results of this grid search. 
For each configuration, we report the validation loss of the final checkpoint.

\begin{figure}[h]
    \centering
    \includegraphics[width=0.9\linewidth]{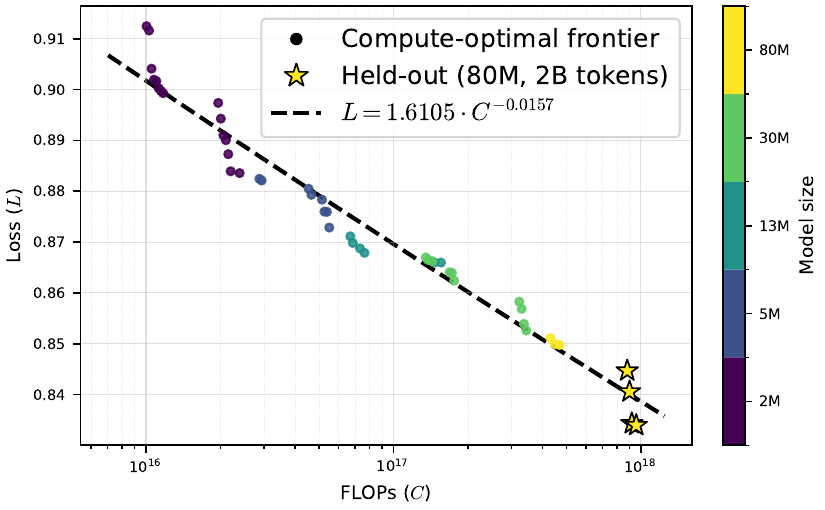}
    \caption{Power-law fit of the validation loss as a function of compute on the Pareto-optimal points of Figure~\ref{fig:scaling} (left).}
    \label{fig:scaling_fit}
\end{figure}

Figure~\ref{fig:scaling_fit} shows the power-law fit of the validation loss as a function of compute discussed in Section~\ref{sub:compute}.

\section{Runtime Evaluation}\label{app:runtime}

\begin{figure}[t]
    \centering
    \includegraphics[width=0.9\linewidth]{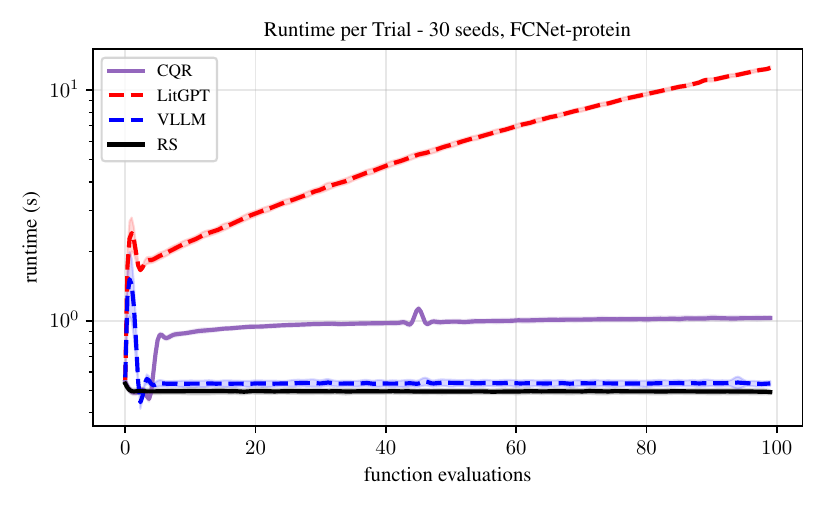}    
    \caption{\textbf{Runtime Comparison on the FC-Net Protein Task.} We report the wall-clock time (log-seconds) across 100 trials for our proposed methods, including native LitGPT and a vLLM-accelerated Hugging Face implementation, as well as Random Search and CQR baselines. Results are aggregated over 30 independent seeds using consistent model checkpoints to ensure comparability.}
    \label{fig:runtimes}
\end{figure}

Figure \ref{fig:runtimes} illustrates the computational efficiency of the evaluated methods when imitating CQR. While the LitGPT-based approach exhibits a monotonic increase in runtime per trial, the remaining methods maintain a relatively constant overhead. Notably, the vLLM-optimized implementation achieves a runtime nearly on par with Random Search and outperforms CQR, demonstrating the scalability of our inference framework for iterative optimization. All experiments were done on Nvidia H100 GPUs.

\section{Additional Comparison for Imitating Optimizers}\label{app:imitation}

This appendix provides extended empirical results for the optimizer imitation experiments described in the main paper. We evaluate generalization across three increasingly challenging settings: unseen tasks drawn from search spaces encountered during training (Section~\ref{app:imitation_seen_search_spaces}), tasks from entirely unseen search spaces (Section~\ref{app:unseen_search_spaces}), and held-out test tasks (Section~\ref{app:imitation_test}) used for final evaluation. In all figures, shaded regions denote standard deviation across 30 seeds, and all methods are compared under identical evaluation budgets of 100 trials.

\subsection{Unseen Tasks within Search Spaces}\label{app:imitation_seen_search_spaces}

\begin{figure}[h]
    \centering
    \begin{subfigure}[h]{0.24\textwidth}
        \includegraphics[width=0.9\textwidth]{figures/evaluation/eval_nas201-ImageNet16-120_comparison.pdf}
        \caption{\tiny{NAS-Bench-201 ImageNet16-120}}   
    \end{subfigure}
    \begin{subfigure}[h]{0.24\textwidth}
        \includegraphics[width=0.9\textwidth]{figures/evaluation/eval_fcnet-protein_comparison.pdf}
        \caption{\tiny{FC-Net Protein}}   
    \end{subfigure}
    \begin{subfigure}[h]{0.24\textwidth}
        \centering
        \includegraphics[width=0.9\textwidth]{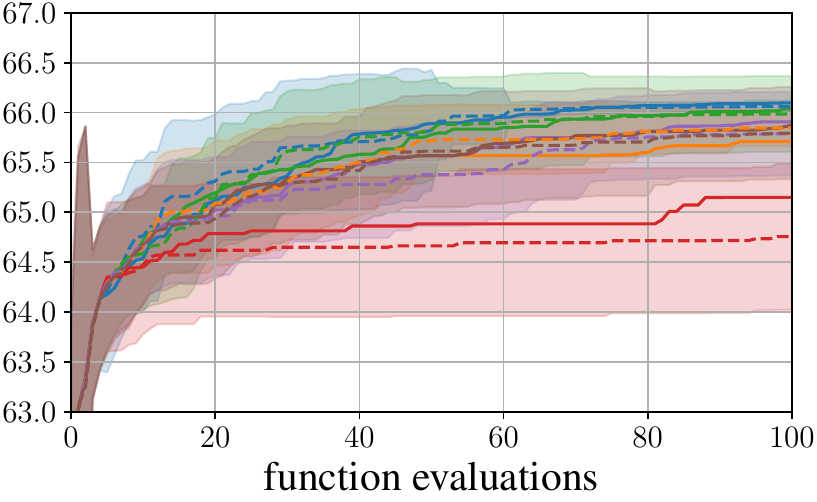}       
        \caption{\tiny{LC-Bench Albert}}   
    \end{subfigure} 
    \begin{subfigure}[h]{0.24\textwidth}
        \centering
        \includegraphics[width=0.9\textwidth]{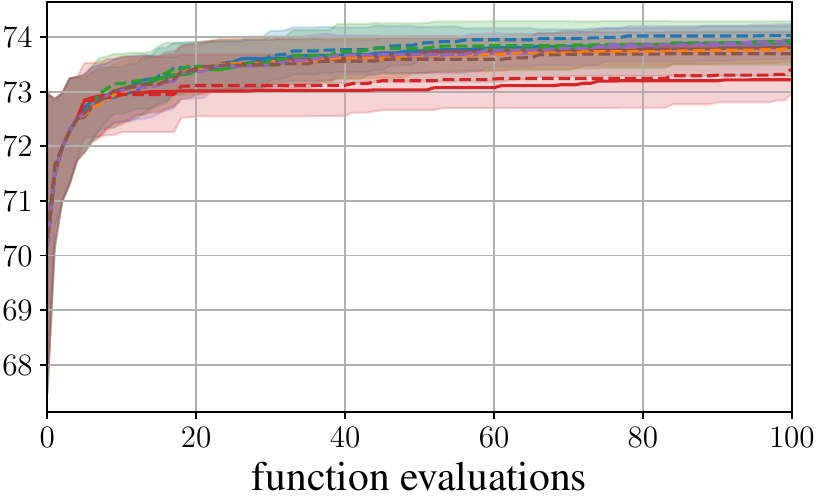}
        \caption{\tiny{LC-Bench Christine}}   
    \end{subfigure} 
    \begin{subfigure}[h]{0.24\textwidth}
        \centering
        \includegraphics[width=0.9\textwidth]{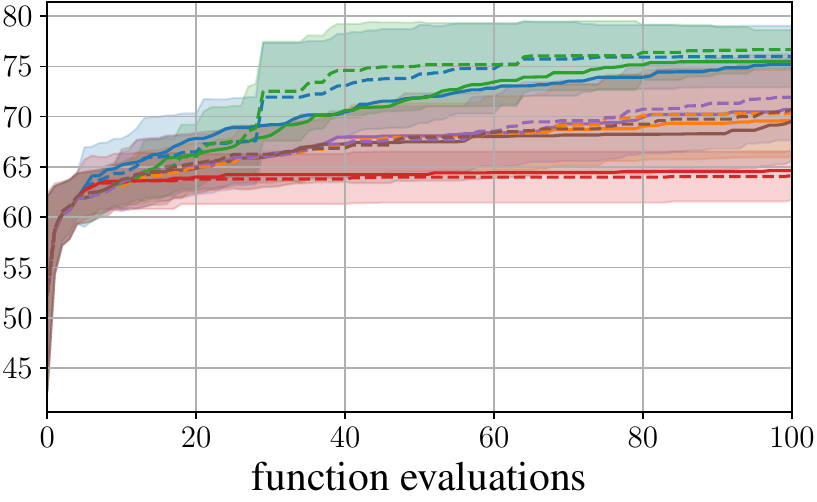}         
        \caption{\tiny{LC-Bench Covertype}}   
    \end{subfigure} 
    \begin{subfigure}[h]{0.24\textwidth}
        \centering
        \includegraphics[width=0.9\textwidth]{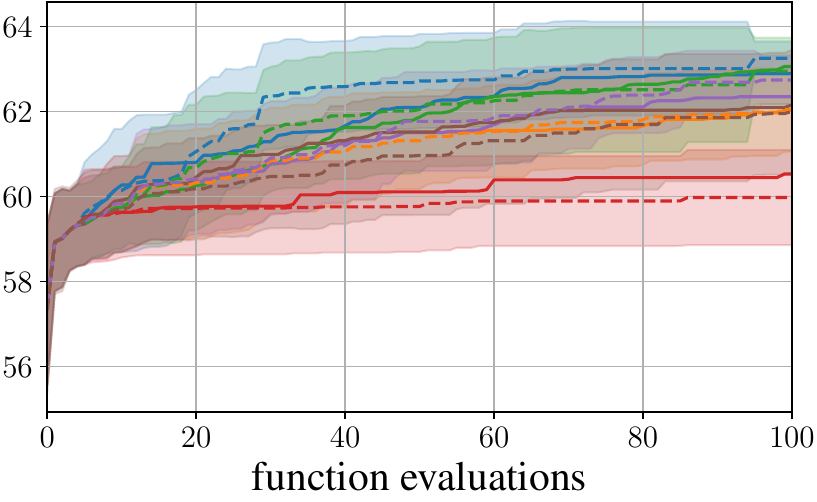}
        \caption{\tiny{LC-Bench Airlines}}   
    \end{subfigure}   
    \begin{subfigure}[h]{0.24\textwidth}
        \centering
        \includegraphics[width=0.9\textwidth]{figures/evaluation/eval_lcbench_Fashion-MNIST_comparison.pdf}
        \caption{\tiny{LC-Bench F-MNIST}}   
    \end{subfigure} 
    \begin{subfigure}[h]{0.24\textwidth}
        \centering
        \includegraphics[width=0.9\textwidth]{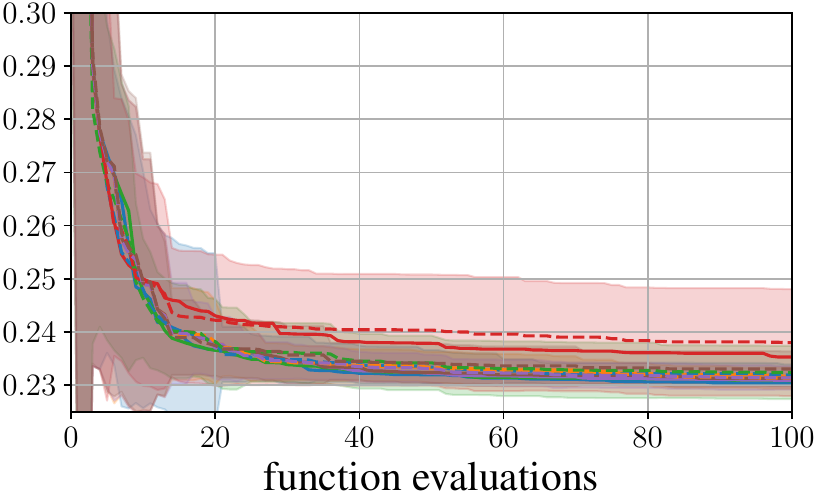}
        \caption{\tiny{PD1 ResNet ImageNet}}   
    \end{subfigure} 
     \begin{subfigure}[h]{0.24\textwidth}
        \centering
        \includegraphics[width=0.9\textwidth]{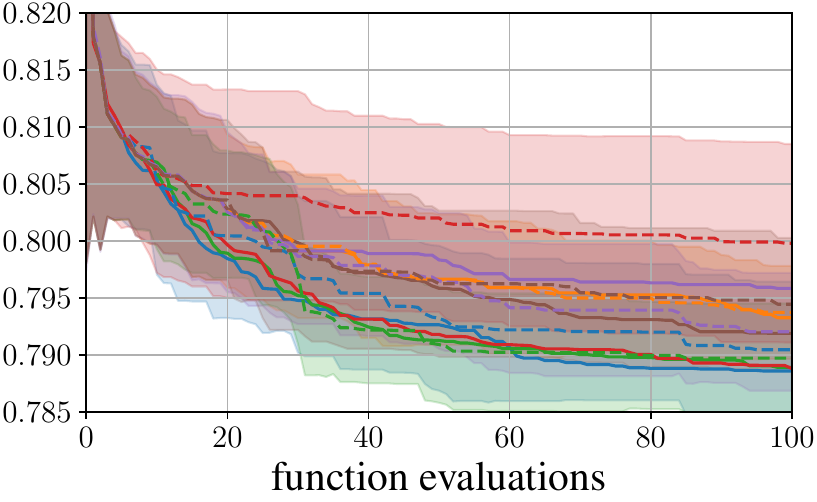}
        \caption{\tiny{PD1 Transformer UniRef50}}   
    \end{subfigure}    
      \begin{subfigure}[h]{0.24\textwidth}
        \centering
        \includegraphics[width=0.9\textwidth]{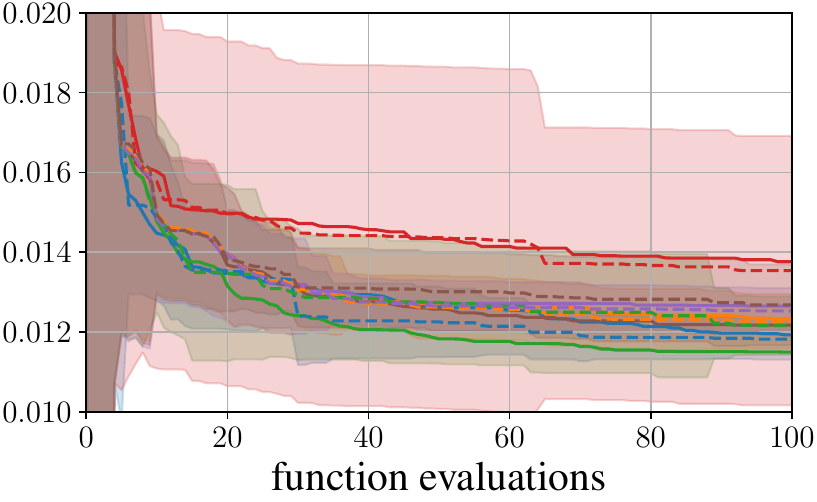}       
        \caption{\tiny{PD1 CNN MNIST}}   
    \end{subfigure}       
    \caption{Generalization to unseen tasks of known search spaces. Each panel shows the objective function value performance as a function of the number of function evaluations. Our optimizer closely tracks the behavior of the target optimizers across tasks, demonstrating robust within-space generalization.}  
    \label{fig:unseen_tasks}
\end{figure}

Figure~\ref{fig:unseen_tasks} evaluates the ability of our model to imitate optimizers on tasks that were not seen during training but belong to search spaces that were. This setting tests whether the learned optimizer captures transferable optimization behavior rather than memorizing task-specific solutions. Results are reported across a diverse suite of benchmarks spanning neural architecture search (NAS-Bench-201), large-scale deep learning pipelines (PD1), and tabular hyperparameter optimization (LC-Bench, FC-Net), covering a range of input dimensionalities and response surface characteristics.

\subsection{Unseen  Search Spaces}\label{app:unseen_search_spaces}

\begin{figure}[h]
    \centering
    \begin{subfigure}[h]{0.24\textwidth}
        \includegraphics[width=0.9\textwidth]{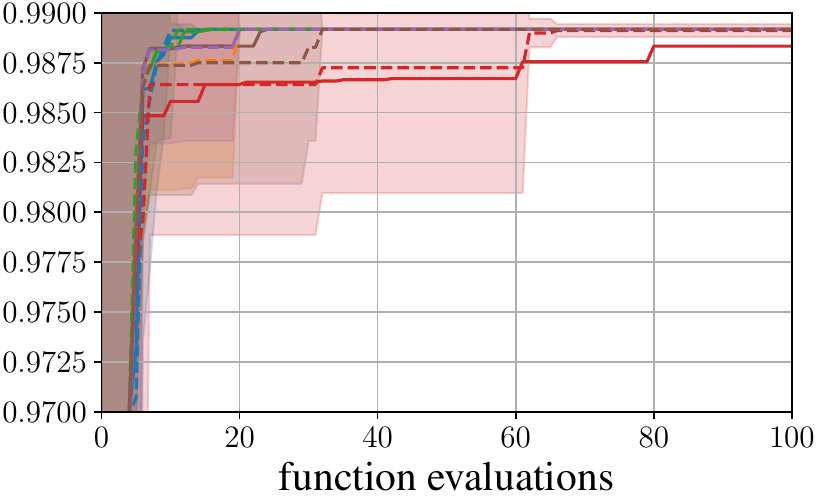}
        \caption{\tiny{HPO-B 7607 146066}}   
    \end{subfigure}
    \begin{subfigure}[h]{0.24\textwidth}
        \centering
        \includegraphics[width=0.9\textwidth]{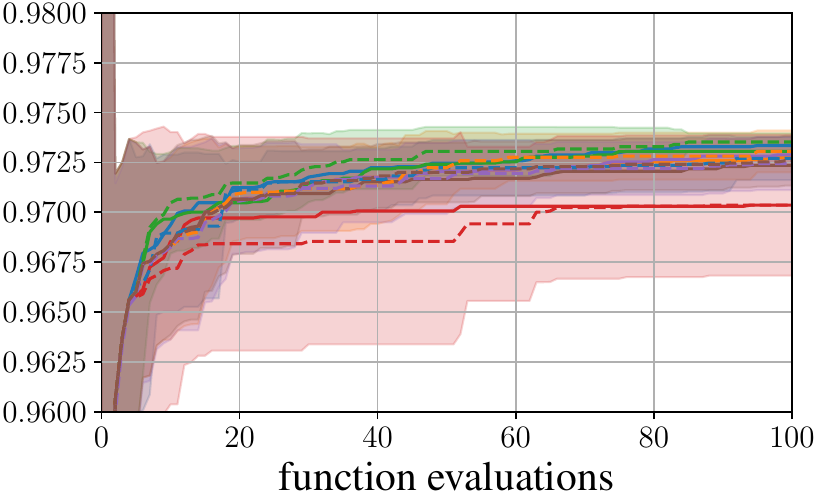}     
        \caption{\tiny{HPO-B 5971 145878}}   
    \end{subfigure} 
    \begin{subfigure}[h]{0.24\textwidth}
        \centering
        \includegraphics[width=0.9\textwidth]{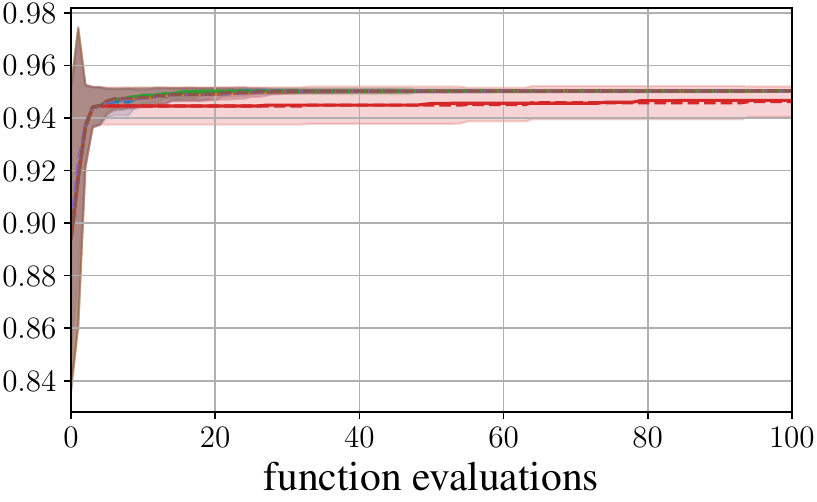}      
        \caption{\tiny{HPO-B 5636 145854}}   
    \end{subfigure} 
    \begin{subfigure}[h]{0.24\textwidth}
        \centering
        \includegraphics[width=0.9\textwidth]{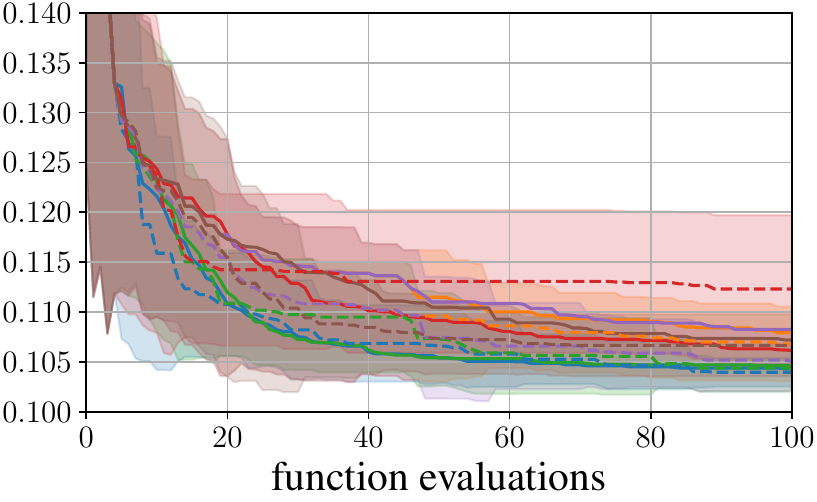}
        \caption{\tiny{TabRepo CatBoost MNIST}}   
    \end{subfigure} 
    \begin{subfigure}[h]{0.24\textwidth}
        \centering
        \includegraphics[width=0.9\textwidth]{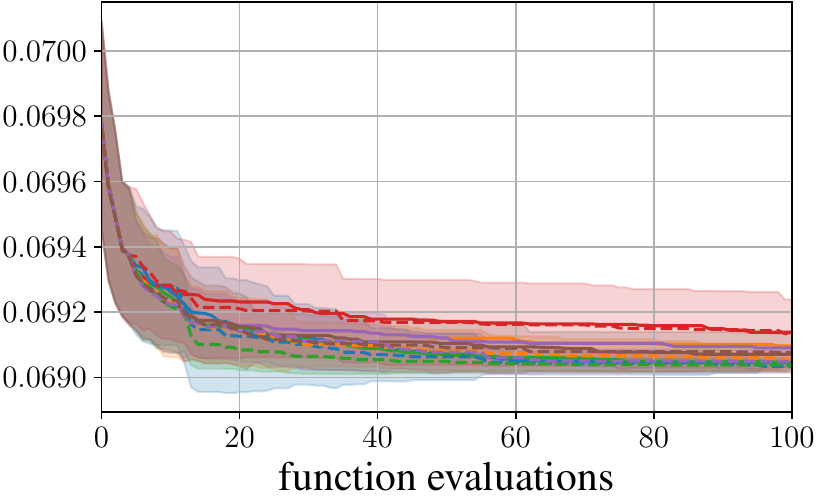}
        \caption{\tiny{TabRepo CatBoost Adult}}   
    \end{subfigure} 
     \begin{subfigure}[h]{0.24\textwidth}
        \centering
        \includegraphics[width=0.9\textwidth]{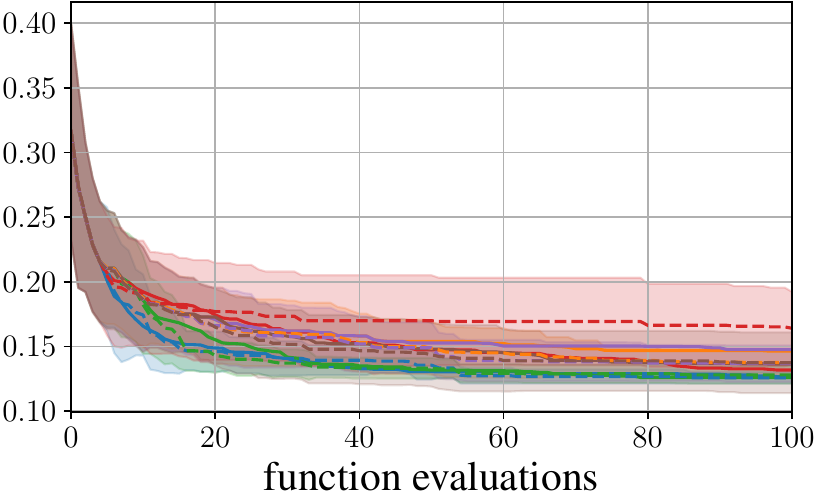}
        \caption{\tiny{TabRepo CatBoost Covertype}}   
    \end{subfigure}    
      \begin{subfigure}[h]{0.24\textwidth}
        \centering
        \includegraphics[width=0.9\textwidth]{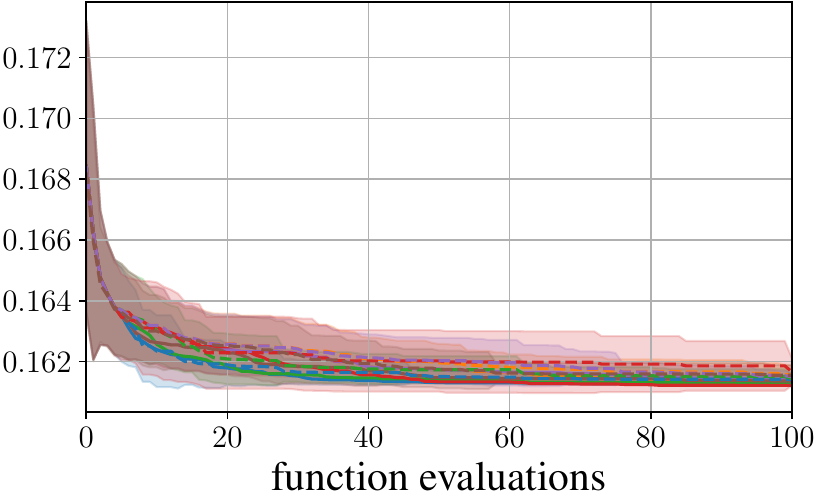}      
        \caption{\tiny{TabRepo CatBoost Higgs}}   
    \end{subfigure}       
   \begin{subfigure}[h]{0.24\textwidth}
        \centering
        \includegraphics[width=0.9\textwidth]{figures/evaluation/eval_tabrepo_CatBoost_Fashion-MNIST_comparison.pdf}                     
        \caption{\tiny{TabRepo CatBoost F-MNIST}}   
    \end{subfigure}       
     \begin{subfigure}[h]{0.24\textwidth}
        \centering
        \includegraphics[width=0.9\textwidth]{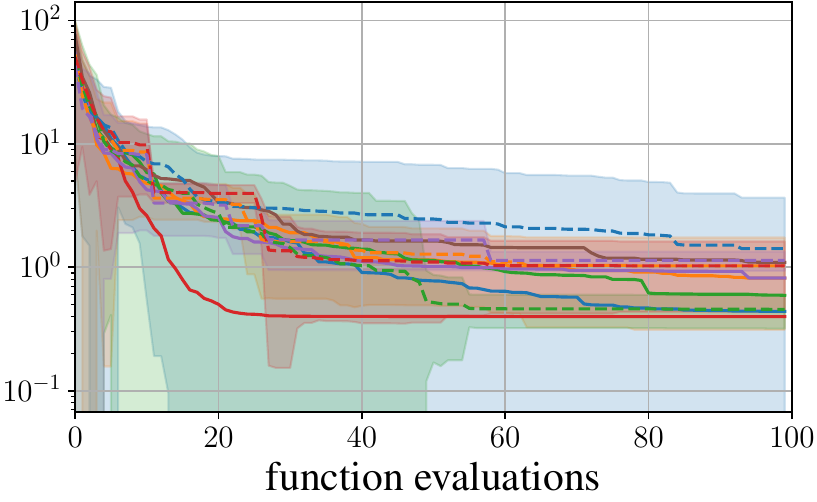}
        \caption{\tiny{Branin}}   
    \end{subfigure}         
    \begin{subfigure}[h]{0.24\textwidth}
        \centering
        \includegraphics[width=0.9\textwidth]{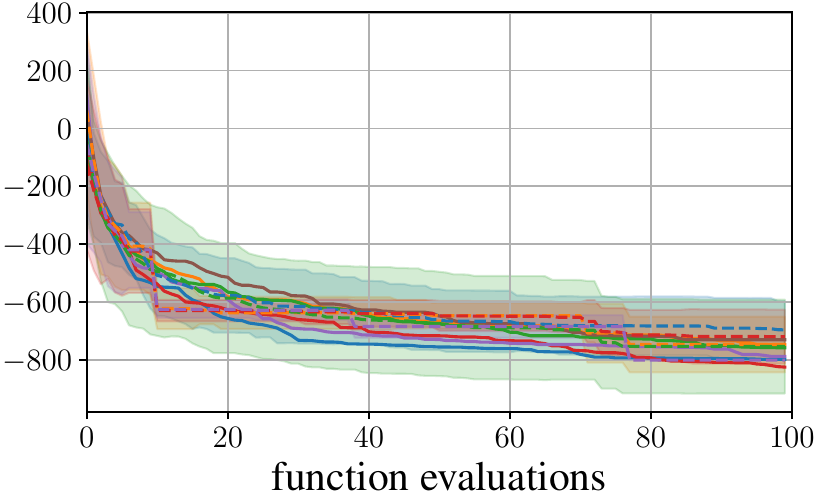}
        \caption{\tiny{Eggholder}}   
    \end{subfigure}          
     \begin{subfigure}[h]{0.24\textwidth}
        \centering
        \includegraphics[width=0.9\textwidth]{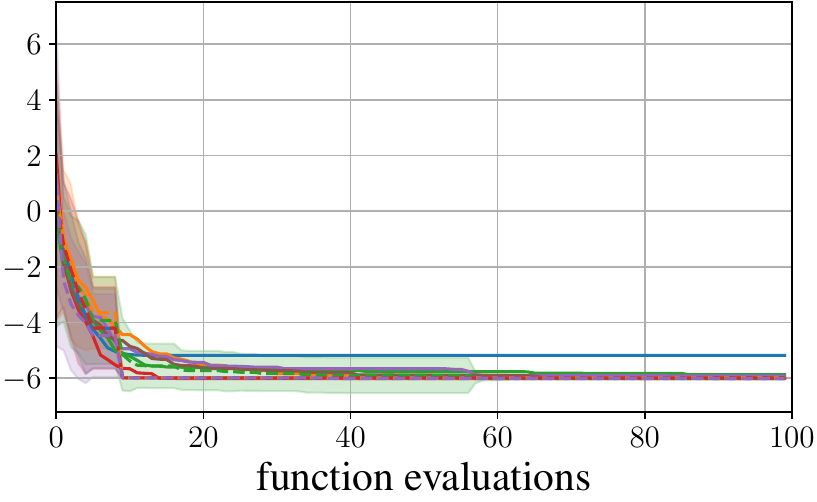}
        \caption{\tiny{Forrester}}   
    \end{subfigure}             
     \begin{subfigure}[h]{0.24\textwidth}
        \centering
        \includegraphics[width=0.9\textwidth]{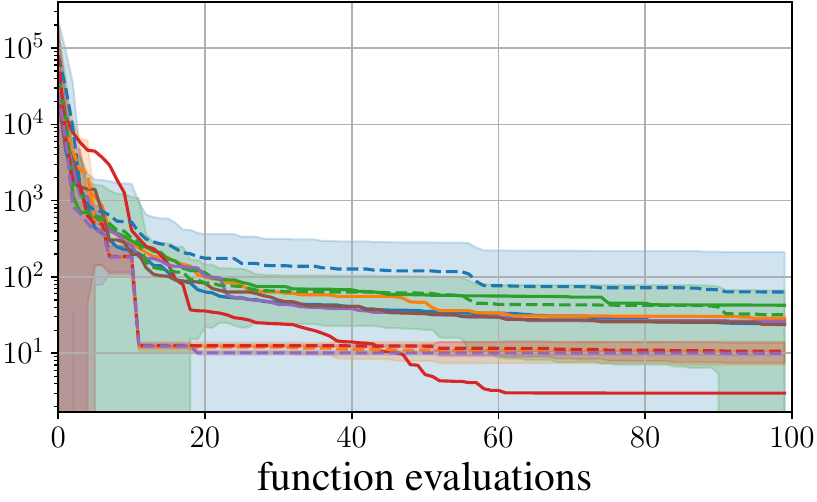}
        \caption{\tiny{Goldstein-Price}}   
    \end{subfigure}             
    \caption{Generalization to unseen search spaces. Results span three HPO-B classification tasks, CatBoost hyperparameter tuning on five TabRepo datasets, and four global optimization functions. Despite having no exposure to these search spaces during training, our optimizer maintains comparable performance relative to the target, indicating that the learned policy captures general exploratory and exploitative strategies rather than space-specific heuristics.}  
    \label{fig:unseen_search_spaces}
\end{figure}

Figure~\ref{fig:unseen_search_spaces} examines the more demanding setting in which the optimizer is evaluated on search spaces that were entirely absent from the training distribution. This tests the capacity of the learned policy to transfer optimization strategies to novel hyperparameter configurations and objective landscapes. Benchmarks include HPO-B, TabRepo CatBoost tasks across multiple datasets, and classical global optimization test functions (Branin, Eggholder, Forrester, Goldstein-Price) that provide interpretable, low-dimensional baselines with known optima.

\subsection{Test Tasks}\label{app:imitation_test}

\begin{figure}[h]
    \centering
    \includegraphics[width=0.24\textwidth]{figures/test/eval_deepar_electricity_comparison.pdf}
    \includegraphics[width=0.24\textwidth]{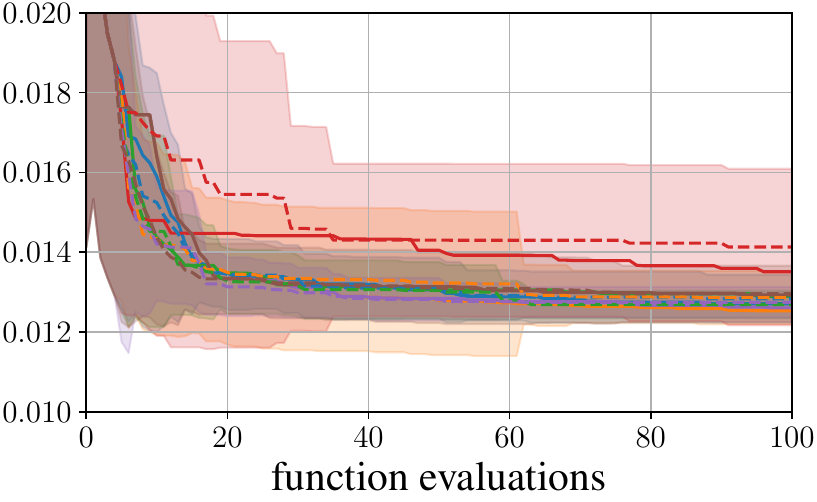}
    \includegraphics[width=0.24\textwidth]{figures/test/eval_deepar_m4-Yearly_comparison.pdf}
    \includegraphics[width=0.24\textwidth]{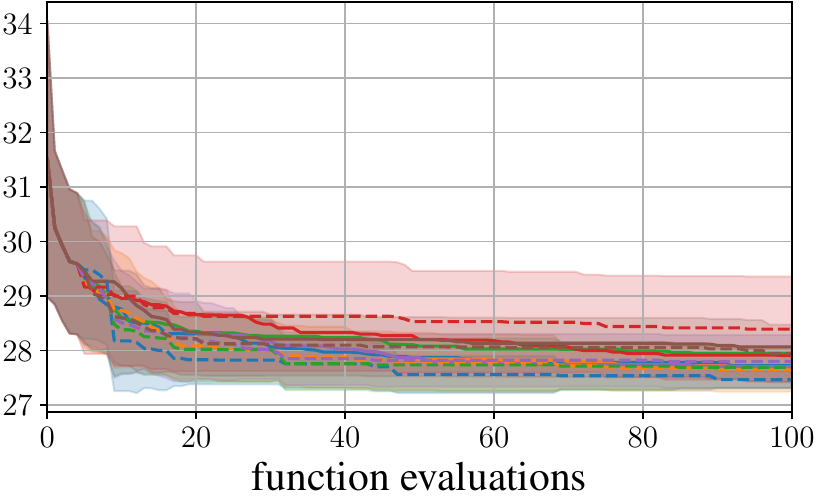}
    \includegraphics[width=0.24\textwidth]{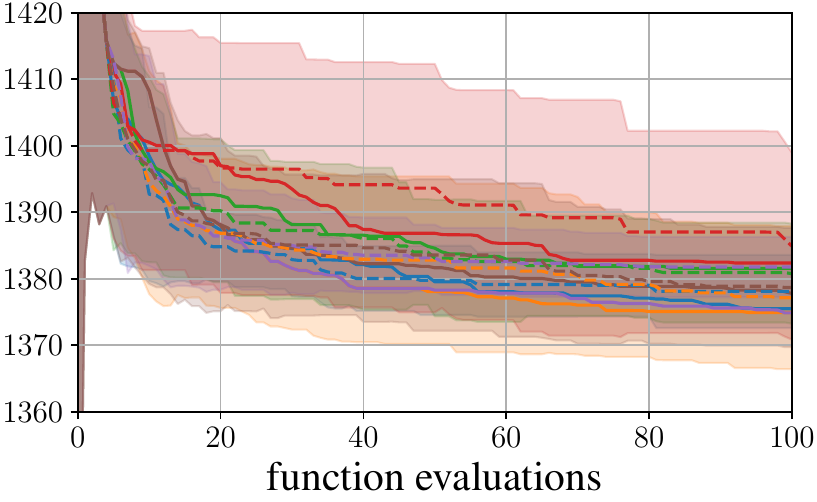}
    \includegraphics[width=0.24\textwidth]{figures/test/eval_deepar_traffic_comparison.pdf}
    \includegraphics[width=0.24\textwidth]{figures/test/eval_deepar_m4-Quarterly_comparison.pdf}
    \includegraphics[width=0.24\textwidth]{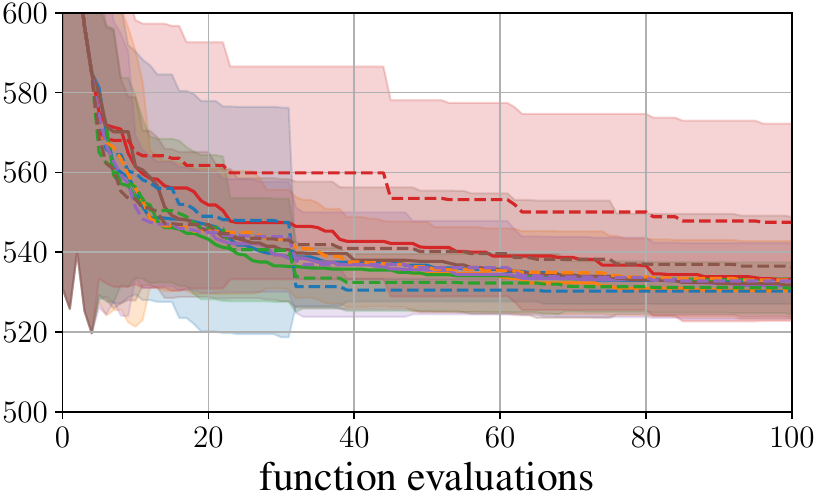}
    \includegraphics[width=0.24\textwidth]{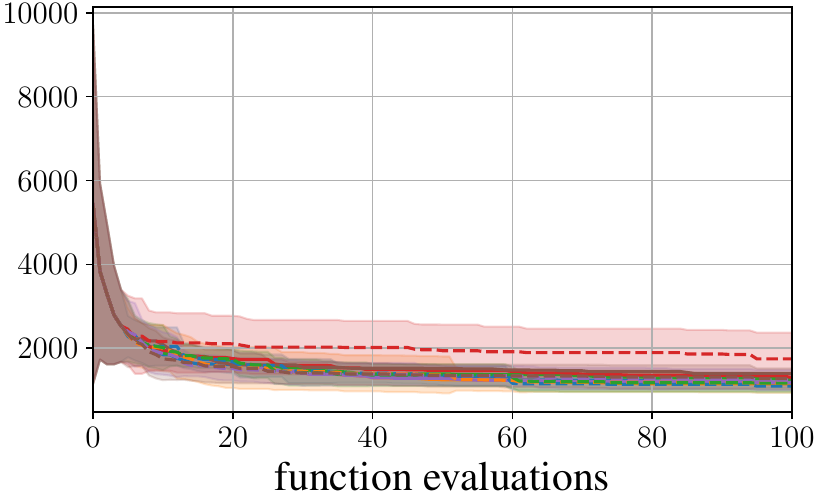}
    \includegraphics[width=0.24\textwidth]{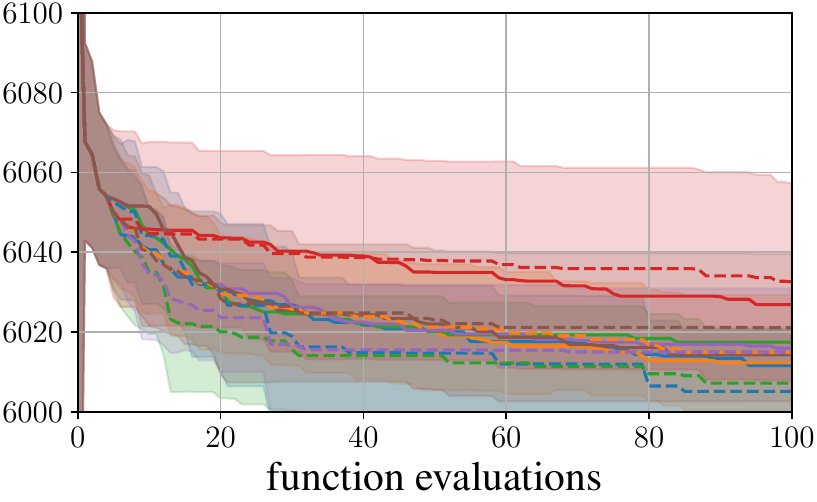}
    \caption{Generalization to held-out test tasks from the DeepAR benchmark. Each panel plots the objective function value against the number of function evaluations for different optimizers. While HEBO shows large variance, our optimizer generally matches or approaches the target optimizer's trajectory closely, demonstrating that the learned policy retains its effectiveness on an unseen domain not represented in the training distribution.}
    \label{fig:test_tasks}
\end{figure}

Figure~\ref{fig:test_tasks} reports results on tasks that were withheld throughout all stages of development and hyperparameter tuning, and thus provide an unbiased estimate of out-of-distribution generalization. The benchmark consists of DeepAR \citep{salinas2020deepar} time series forecasting tasks across ten datasets spanning electricity consumption and car traffic, among others.

\subsection{Imitation Quality Across Scale}\label{app:eval_scale}

\begin{figure}[h]
    \centering
    \includegraphics[width=\linewidth]{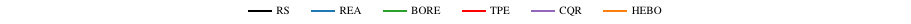}\\[1pt]
    \begin{subfigure}{\linewidth}
        \centering
        \includegraphics[width=\linewidth]{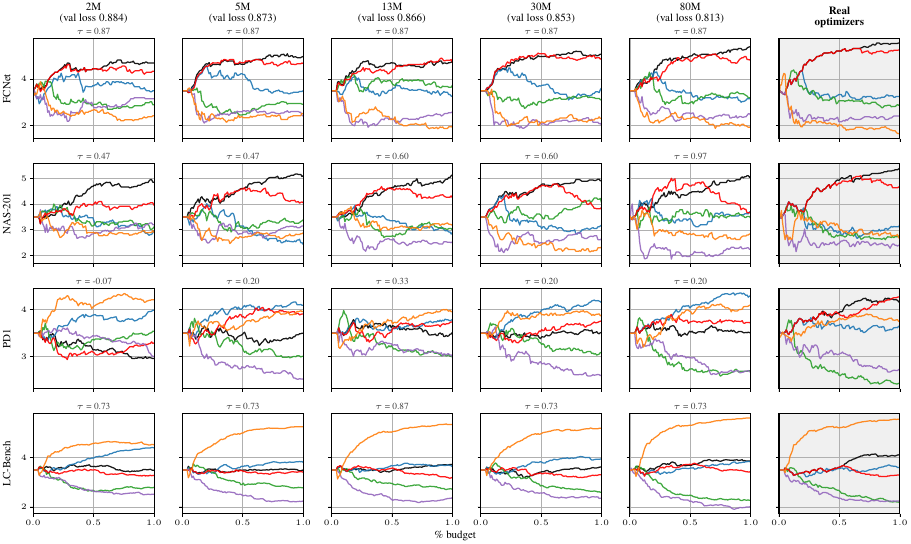}
        \caption{Average ranks.}
        \label{fig:eval:modelsize2B:rank}
    \end{subfigure}\\[1pt]
    \begin{subfigure}{\linewidth}
        \centering
        \includegraphics[width=\linewidth]{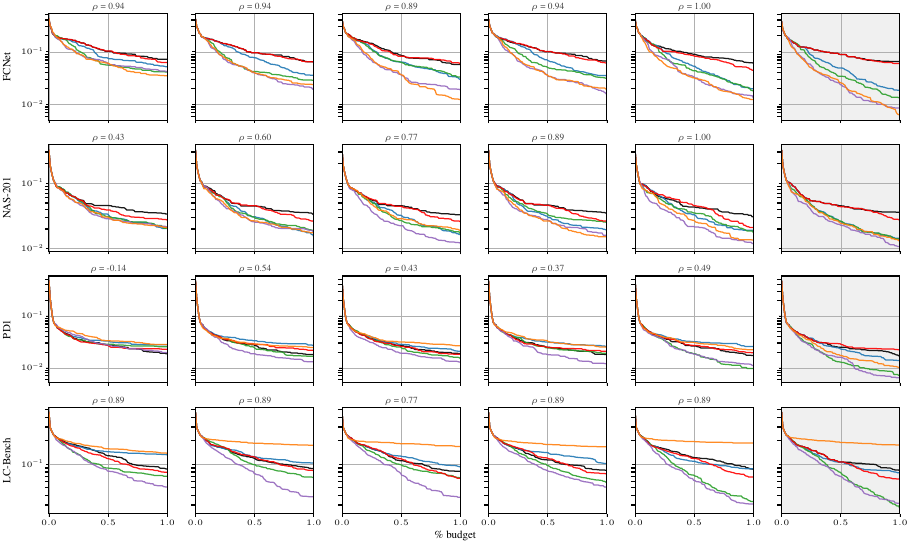}
        \caption{Average normalized regrets.}
        \label{fig:eval:modelsize2B:regret}
    \end{subfigure}
    \caption{\textbf{Imitation quality vs.\ model size at a fixed $2$B-token budget.}
    Average rank~(a) and normalized regret~(b) of the imitated optimizers
    (columns: $2$M/$5$M/$13$M/$30$M/$80$M parameters)
    against the original optimizers (last column), on four validation benchmark families
    ($30$ seeds each). Column headers report the validation loss of each model, from $0.884$ ($2$M) to $0.813$ ($80$M); the number above each panel is the correlation with the original optimizers at the end of the optimization budget.}
    \label{fig:eval:modelsize2B}
\end{figure}

\begin{figure}[h]
    \centering
    \includegraphics[width=\linewidth]{figures/rebuttal_eval/legend_strip.pdf}\\[1pt]
    \begin{subfigure}{\linewidth}
        \centering
        \includegraphics[width=\linewidth]{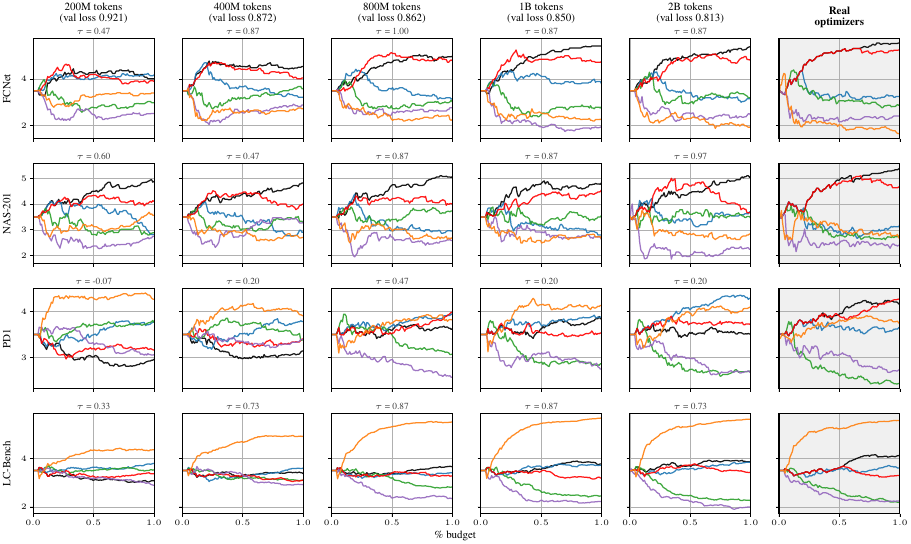}
        \caption{Average ranks.}
        \label{fig:eval:trainbudget:rank}
    \end{subfigure}\\[1pt]
    \begin{subfigure}{\linewidth}
        \centering
        \includegraphics[width=\linewidth]{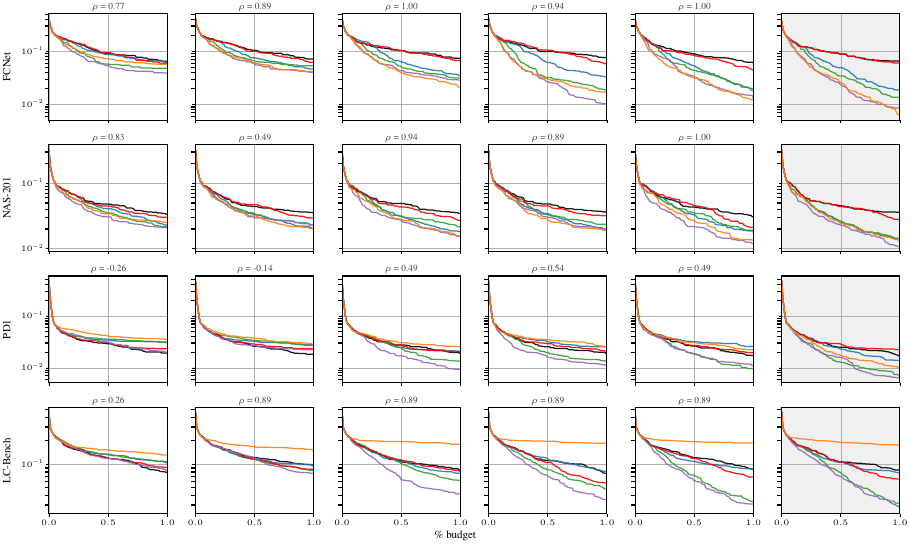}
        \caption{Average normalized regrets.}
        \label{fig:eval:trainbudget:regret}
    \end{subfigure}
    \caption{\textbf{Imitation quality vs.\ training-token budget for the $80$M model.}
    Average rank~(a) and normalized regret~(b) of the imitated optimizers
    (columns: $200$M/$400$M/$800$M/$1$B/$2$B training tokens) against the original
    optimizers (last column), on the same four validation benchmark families
    ($30$ seeds each). Column headers report the validation loss, from $0.921$ ($200$M tokens) to $0.813$ ($2$B tokens); the number above each panel is the correlation with the original optimizers at the end of the optimization budget.}
    \label{fig:eval:trainbudget}
\end{figure}

We evaluate our models across different model sizes and training-token budgets and report aggregate performance statistics on held-out downstream tasks. Concretely, we imitate the six optimizers (RS, REA, BORE, TPE, CQR, HEBO) and run each on the held-out validation tasks of four benchmark families (FC-Net, NAS-Bench-201, PD1, LC-Bench; $30$ seeds each). For every model we report the average rank and normalized regret of the learned optimizers against the original ones (last column, shaded), and we quantify their agreement with the original optimizers through the rank correlation at the end of the optimization budget (Kendall's $\tau$ for ranks, Spearman's $\rho$ for regrets; the number above each panel). We vary two axes independently: (i) model size at a fixed $2$B-token budget ($2$M to $80$M parameters; Figure~\ref{fig:eval:modelsize2B}), and (ii) training-token budget for the $80$M model ($200$M to $2$B tokens; Figure~\ref{fig:eval:trainbudget}); each column header also reports the model's validation loss.

As expected, the performance of the learned optimizers becomes increasingly aligned with that of the original optimization methods as the training-token budget increases. However, for most benchmark families, we observe little to no further improvement beyond a training budget of $1$B tokens: the rank and regret curves and their correlation with the original optimizers largely saturate. This is consistent with the results shown in Figures~\ref{fig:imitation_parameters} and~\ref{fig:imitation_tokens}.

We observe the same pattern along the model-size axis: larger models track the original optimizers more closely, but the gains diminish quickly. At a fixed $2$B-token budget the validation loss changes only mildly (from $0.884$ at $2$M to $0.813$ at $80$M), and, correspondingly, even the smaller models already imitate the original optimizers well. The $80$M model tracks the original optimizers most closely on NAS-Bench-201, whereas PD1 remains difficult at all scales.

Across both axes, downstream alignment tracks the validation loss: a lower validation loss generally corresponds to a higher correlation with the original optimizers. The relationship, however, shows strongly diminishing returns. Once the validation loss is low enough that a model closely imitates the original methods, further reductions yield little additional downstream change. Intuitively, while increasing the compute budget continues to reduce the validation loss, it no longer substantially changes the sampling distribution of our learned optimizers because they already closely imitate the original optimization methods. We expect, however, that as we expand the dataset to include a larger variety of benchmark families and optimization algorithms, the models will continue to benefit from larger model sizes and training-token budgets.

\clearpage

\section{Extensive Qualitative Comparison of the Sampling Distributions}\label{app:sampling_distributions}

\begin{figure}
    \centering
    \begin{subfigure}[h]{\textwidth}
       \centering
       \includegraphics[width=\linewidth]{figures/sampling_distribution/legend_sampling_distribution.pdf}
   \end{subfigure}
    \begin{subfigure}[h]{\textwidth}
       \centering
       \includegraphics[width=\linewidth]{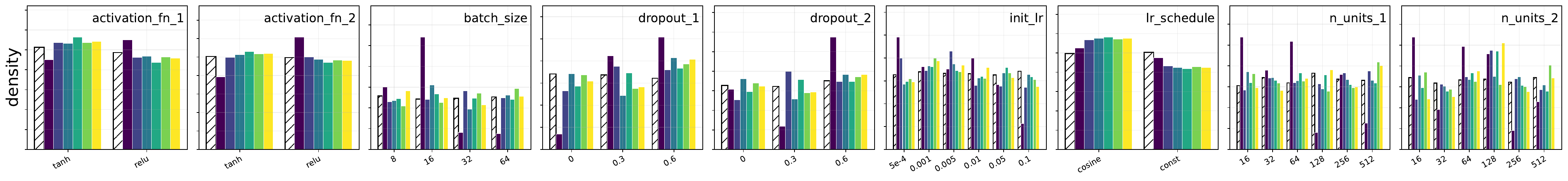}
       \caption{Imitation of Random Search}
   \end{subfigure}
   \begin{subfigure}[h]{\textwidth}
       \centering
        \includegraphics[width=\linewidth]{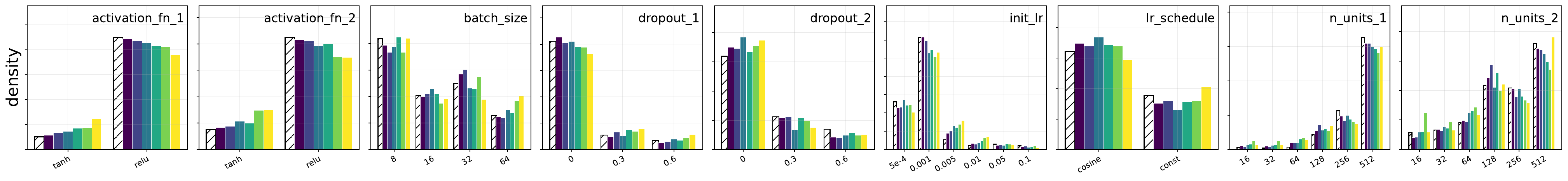}
       \caption{Imitation of CQR}
   \end{subfigure}
   \begin{subfigure}[h]{\textwidth}
       \centering
        \includegraphics[width=\linewidth]{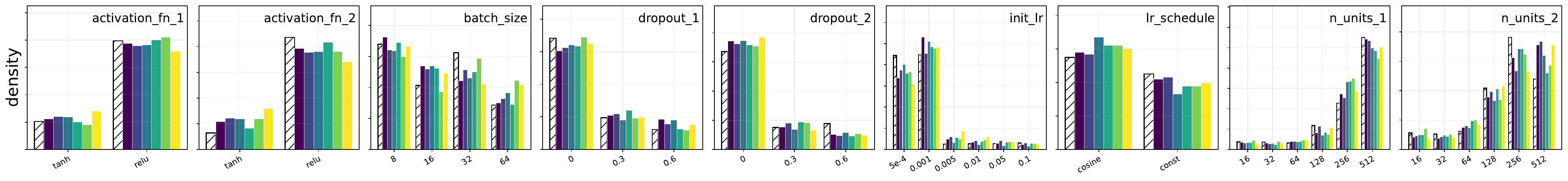}
       \caption{Imitation of BORE}
   \end{subfigure}
   \begin{subfigure}[h]{\textwidth}
       \centering
        \includegraphics[width=\linewidth]{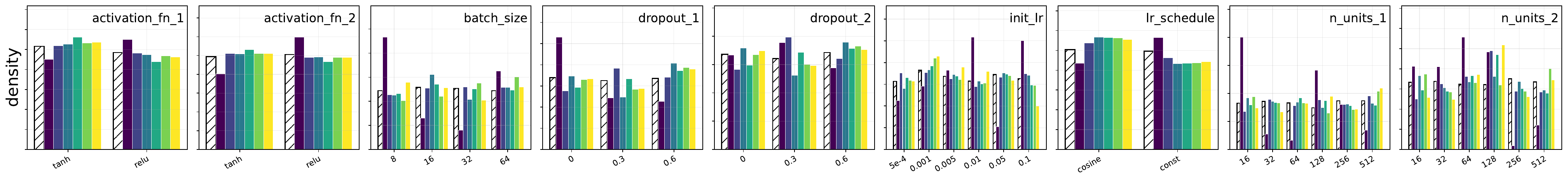}
       \caption{Imitation of TPE}
   \end{subfigure}

   \caption{Comparison of the sampling distributions on FC-Net search space. }\label{fig:sampling_dist:appendix:fcnet}
   
\end{figure}

\begin{figure}
    \centering
    \begin{subfigure}[h]{\textwidth}
       \centering
       \includegraphics[width=\linewidth]{figures/sampling_distribution/legend_sampling_distribution.pdf}
   \end{subfigure}
    \begin{subfigure}[h]{\textwidth}
       \centering
       \includegraphics[width=\linewidth]{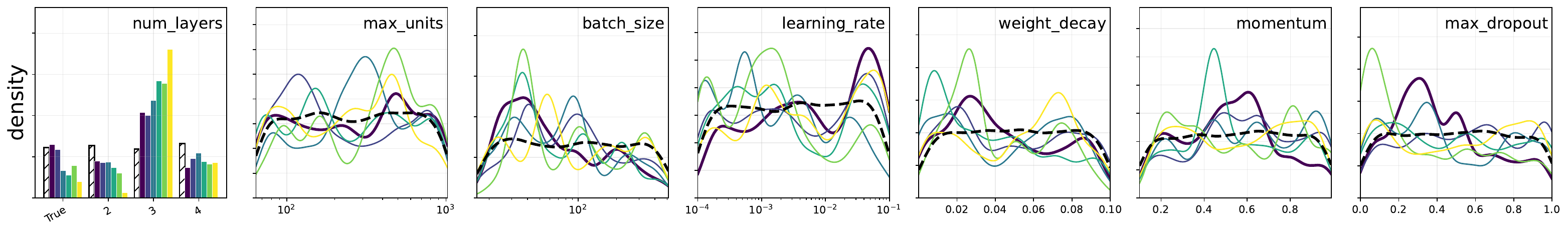}
       \caption{Imitation of Random Search}
   \end{subfigure}
   \begin{subfigure}[h]{\textwidth}
       \centering
        \includegraphics[width=\linewidth]{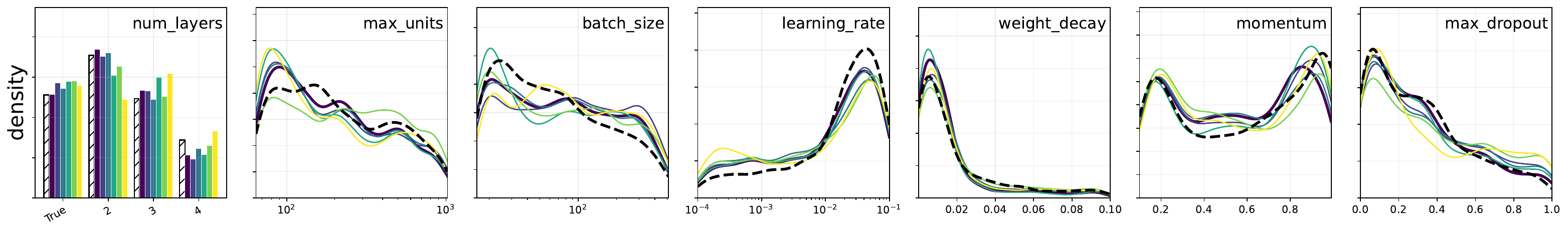}
       \caption{Imitation of CQR}
   \end{subfigure}
   \begin{subfigure}[h]{\textwidth}
       \centering
        \includegraphics[width=\linewidth]{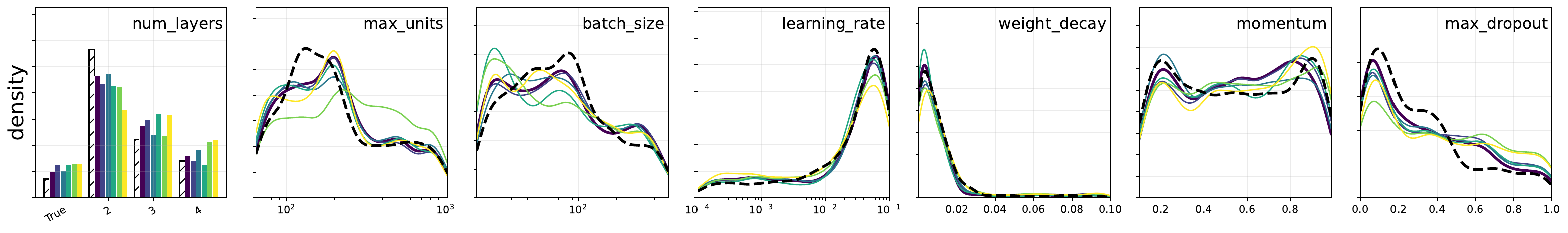}
       \caption{Imitation of BORE}
   \end{subfigure}
   \begin{subfigure}[h]{\textwidth}
       \centering
        \includegraphics[width=\linewidth]{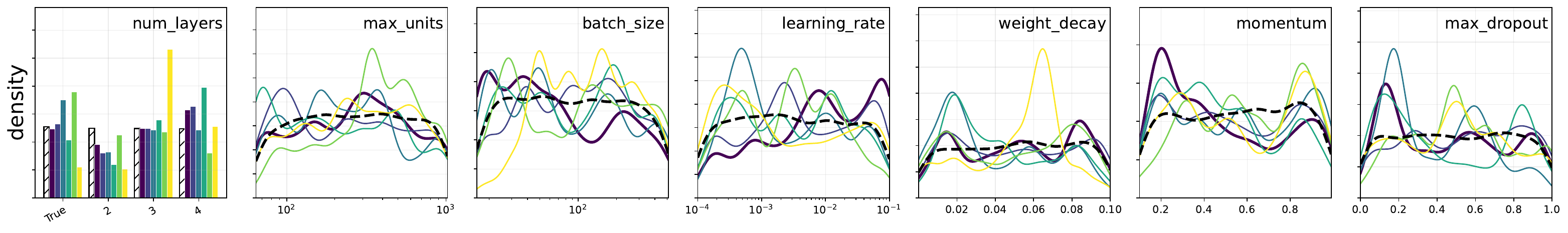}
       \caption{Imitation of TPE}
   \end{subfigure}

   \caption{Comparison of the sampling distributions on LC-Bench (Fashion-MNIST) search space. }\label{fig:sampling_dist:appendix:lcbench}
\end{figure}

\begin{figure}
    \centering
    \begin{subfigure}[h]{\textwidth}
       \centering
       \includegraphics[width=\linewidth]{figures/sampling_distribution/legend_sampling_distribution.pdf}
   \end{subfigure}
    \begin{subfigure}[h]{\textwidth}
       \centering
       \includegraphics[width=\linewidth]{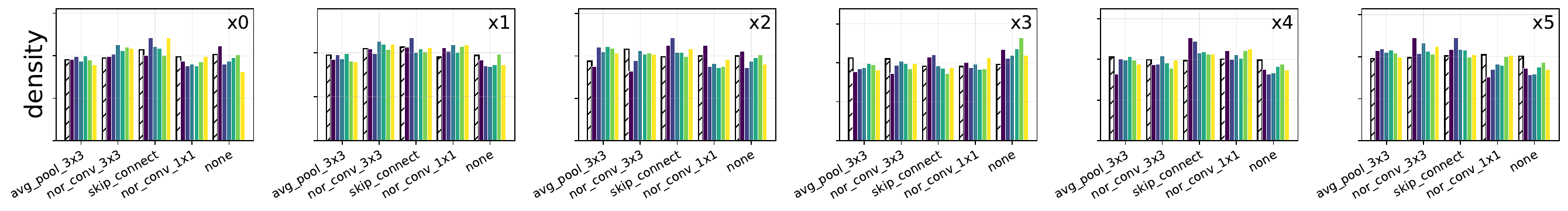}
       \caption{Imitation of Random Search}
   \end{subfigure}
   \begin{subfigure}[h]{\textwidth}
       \centering
        \includegraphics[width=\linewidth]{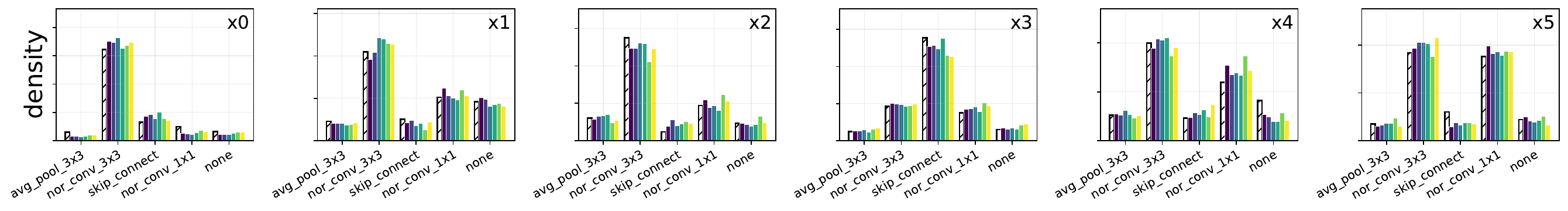}
       \caption{Imitation of CQR}
   \end{subfigure}
   \begin{subfigure}[h]{\textwidth}
       \centering
        \includegraphics[width=\linewidth]{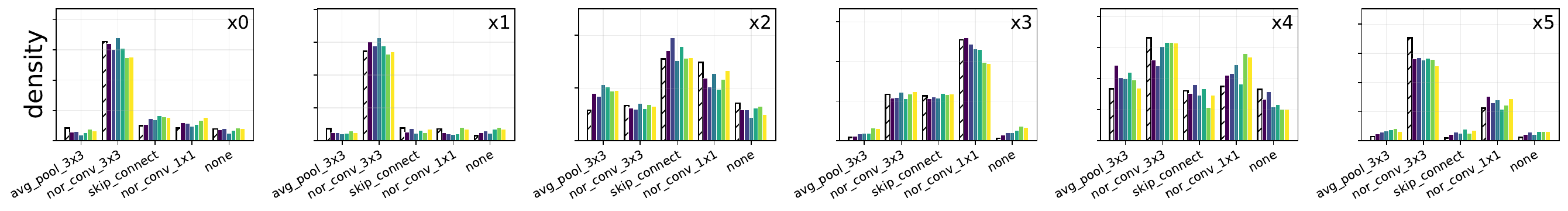}
       \caption{Imitation of BORE}
   \end{subfigure}
   \begin{subfigure}[h]{\textwidth}
       \centering
        \includegraphics[width=\linewidth]{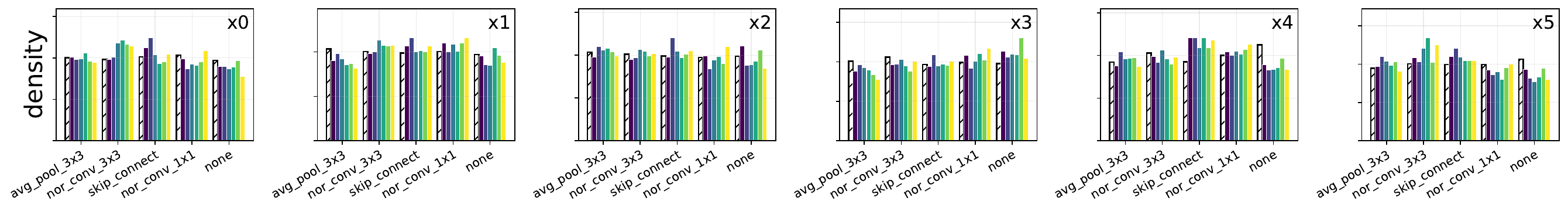}
       \caption{Imitation of TPE}
   \end{subfigure}

   \caption{Comparison of the sampling distributions on NAS-Bench-201 (ImageNet) search space.}\label{fig:sampling_dist:appendix:nas201}
\end{figure}

\begin{figure}
    \centering
    \begin{subfigure}[h]{\textwidth}
       \centering
       \includegraphics[width=\linewidth]{figures/sampling_distribution/legend_sampling_distribution.pdf}
   \end{subfigure}
    \begin{subfigure}[h]{\textwidth}
       \centering
       \includegraphics[width=\linewidth]{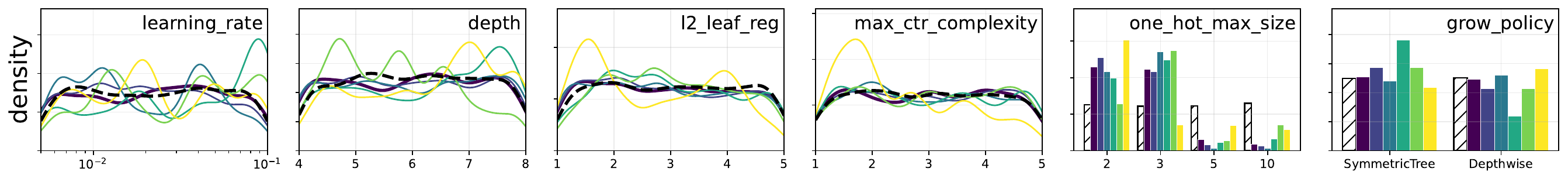}
       \caption{Imitation of Random Search}
   \end{subfigure}
   \begin{subfigure}[h]{\textwidth}
       \centering
        \includegraphics[width=\linewidth]{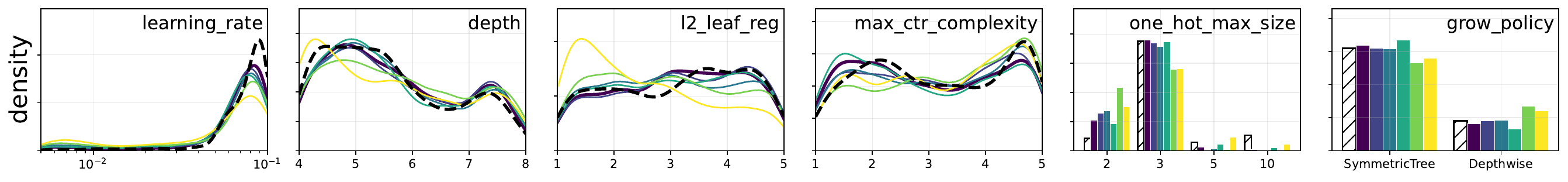}
       \caption{Imitation of CQR}
   \end{subfigure}
   \begin{subfigure}[h]{\textwidth}
       \centering
        \includegraphics[width=\linewidth]{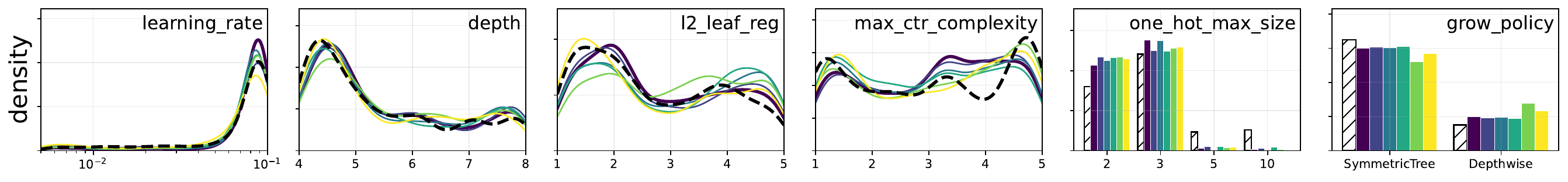}
       \caption{Imitation of BORE}
   \end{subfigure}
   \begin{subfigure}[h]{\textwidth}
       \centering
        \includegraphics[width=\linewidth]{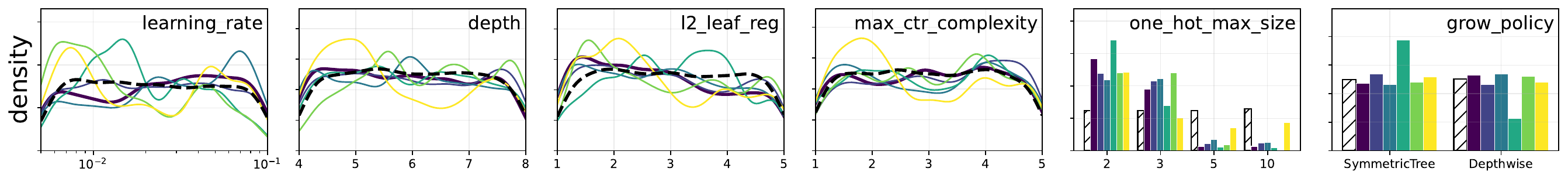}
       \caption{Imitation of TPE}
   \end{subfigure}

   \caption{Comparison of the sampling distributions on TabRepo (CatBoost) search space.}\label{fig:sampling_dist:appendix:tabrepo}
\end{figure}

This section presents a complete comparison of the per-hyperparameter distributions induced by our model and the considered optimizers (Random Search, CQR, BORE, and TPE). For each reference optimizer, we generate a set of
$40$ initial observations using that optimizer, then condition both the optimizer and our model on this same set before sampling. We show results across all the considered search spaces: FC-Net (Figure~\ref{fig:sampling_dist:appendix:fcnet}), LC-Bench (Figure~\ref{fig:sampling_dist:appendix:lcbench}), NAS-Bench-201 (Figure~\ref{fig:sampling_dist:appendix:nas201}), and on TabRepo (Figure~\ref{fig:sampling_dist:appendix:tabrepo}). For our model, we report density as a function of training percentage, with 100\% referring to 2B-token training. 

\clearpage
\section*{NeurIPS Paper Checklist}

\begin{enumerate}

\item {\bf Claims}
    \item[] Question: Do the main claims made in the abstract and introduction accurately reflect the paper's contributions and scope?
    \item[] Answer: \answerYes{} 
    \item[] Justification: Section~\ref{sec:experiments} empirically demonstrates that the dataset proposed in the paper enables the training of foundation models for hyperparameter optimization that can imitate state-of-the-art methods.

\item {\bf Limitations}
    \item[] Question: Does the paper discuss the limitations of the work performed by the authors?
    \item[] Answer: \answerYes{} 
    \item[] Justification: We discuss limitations of the paper in Section~\ref{sec:limitations}.

\item {\bf Theory assumptions and proofs}
    \item[] Question: For each theoretical result, does the paper provide the full set of assumptions and a complete (and correct) proof?
    \item[] Answer: \answerNA{}{} 
    \item[] Justification: The paper does not include theoretical contributions.

    \item {\bf Experimental result reproducibility}
    \item[] Question: Does the paper fully disclose all the information needed to reproduce the main experimental results of the paper to the extent that it affects the main claims and/or conclusions of the paper (regardless of whether the code and data are provided or not)?
    \item[] Answer: \answerYes{} 
    \item[] Justification: We provide details on dataset generation and preprocessing in Section~\ref{sec:dataset}, and details on the training process and model architecture in Section~\ref{sec:model} and Section~\ref{sec:experiments}.

\item {\bf Open access to data and code}
    \item[] Question: Does the paper provide open access to the data and code, with sufficient instructions to faithfully reproduce the main experimental results, as described in supplemental material?
    \item[] Answer: \answerYes{} 
    \item[] Justification: We make the dataset, the code for dataset generation, training, and evaluation, as well as the model checkpoints, publicly available.

\item {\bf Experimental setting/details}
    \item[] Question: Does the paper specify all the training and test details (e.g., data splits, hyperparameters, how they were chosen, type of optimizer) necessary to understand the results?
    \item[] Answer: \answerYes{} 
    \item[] Justification: Details regarding the dataset are presented in Section~\ref{sec:dataset}, and details about the experiments are provided in Section~\ref{sec:experiments}.

\item {\bf Experiment statistical significance}
    \item[] Question: Does the paper report error bars suitably and correctly defined or other appropriate information about the statistical significance of the experiments?
    \item[] Answer: \answerYes{} 
    \item[] Justification: We report the standard deviation over 30 independent repetitions for each optimization method in Section~\ref{sec:experiments}.

\item {\bf Experiments compute resources}
    \item[] Question: For each experiment, does the paper provide sufficient information on the computer resources (type of compute workers, memory, time of execution) needed to reproduce the experiments?
    \item[] Answer: \answerYes{} 
    \item[] Justification: We provide details about our compute environment in Section~\ref{sec:experiments}. Training all models described in that section required an estimated total of 1,872 GPU hours on H100 GPUs. For dataset generation, we ran the optimizers on CPUs, resulting in an estimated total of 50,595 CPU hours.
    
\item {\bf Code of ethics}
    \item[] Question: Does the research conducted in the paper conform, in every respect, with the NeurIPS Code of Ethics \url{https://neurips.cc/public/EthicsGuidelines}?
    \item[] Answer: \answerYes{} 
    \item[] Justification: 

\item {\bf Broader impacts}
    \item[] Question: Does the paper discuss both potential positive societal impacts and negative societal impacts of the work performed?
    \item[] Answer: \answerNA{} 
    \item[] Justification: Models trained on the dataset presented in this work imitate existing state-of-the-art black-box optimization methods; therefore, we do not anticipate any societal impact.
    
\item {\bf Safeguards}
    \item[] Question: Does the paper describe safeguards that have been put in place for responsible release of data or models that have a high risk for misuse (e.g., pre-trained language models, image generators, or scraped datasets)?
    \item[] Answer: \answerNA{} 
    \item[] Justification: Our models generate black-box optimization trajectories and therefore do not require any safeguards.

\item {\bf Licenses for existing assets}
    \item[] Question: Are the creators or original owners of assets (e.g., code, data, models), used in the paper, properly credited and are the license and terms of use explicitly mentioned and properly respected?
    \item[] Answer: \answerYes{} 
    \item[] Justification: We reference all the original work for the benchmarks included in our dataset.

\item {\bf New assets}
    \item[] Question: Are new assets introduced in the paper well documented and is the documentation provided alongside the assets?
    \item[] Answer: \answerYes{} 
    \item[] Justification: We provide details of our dataset in Section~\ref{sec:dataset}.

\item {\bf Crowdsourcing and research with human subjects}
    \item[] Question: For crowdsourcing experiments and research with human subjects, does the paper include the full text of instructions given to participants and screenshots, if applicable, as well as details about compensation (if any)? 
    \item[] Answer: \answerNA{} 
    \item[] Justification: No human subjects were involved in creating the dataset.

\item {\bf Institutional review board (IRB) approvals or equivalent for research with human subjects}
    \item[] Question: Does the paper describe potential risks incurred by study participants, whether such risks were disclosed to the subjects, and whether Institutional Review Board (IRB) approvals (or an equivalent approval/review based on the requirements of your country or institution) were obtained?
    \item[] Answer: \answerNA{} 
    \item[] Justification: No human subjects were involved in this dataset.

\item {\bf Declaration of LLM usage}
    \item[] Question: Does the paper describe the usage of LLMs if it is an important, original, or non-standard component of the core methods in this research? Note that if the LLM is used only for writing, editing, or formatting purposes and does \emph{not} impact the core methodology, scientific rigor, or originality of the research, declaration is not required.
    \item[] Answer: \answerNA{} 
    \item[] Justification: No LLMs were involved in the creation of the dataset.

\end{enumerate}

\end{document}